%% file: iclr2027_conference_v4.tex
\documentclass{article} 
\usepackage{iclr2027_conference,times}
\input{math_commands.tex}

\usepackage{xcolor}
\usepackage[normalem]{ulem}   
\usepackage{hyperref}
\usepackage{url}
\usepackage{comment}
\usepackage{booktabs}
\usepackage{graphicx}
\usepackage{array}
\usepackage{multirow}

\newcommand{\del}{\bgroup\markoverwith{\textcolor{red}{\rule[0.5ex]{2pt}{0.4pt}}}\ULon}   

\newcommand{\Narrow}{\mathrm{Narrow}}
\newcommand{\Headroom}{\mathrm{Headroom}}

\newcommand{\topten}{\mathrm{top10}}

\title{An important Bottleneck in LLM Optimizers: Principle-Level Local Optima}

\author{%
\begin{minipage}{\textwidth}\raggedright\bfseries
Nian Ran$^{1,2,\ddagger}$ \quad Tianhe Jia$^{3,4,5,\ddagger,*}$ \quad Zhengyang Mu$^{5}$ \quad
Yue Wang$^{4,7, \dagger}$ \quad  Sebastian Brulin$^{6}$ \quad Markus Olhofer$^{6}$ \quad
Mauricio A.~\'Alvarez$^{1}$ \quad Yang Liu$^{3,\dagger}$ \quad Richard Allmendinger$^{1,\dagger}$
\end{minipage}  \\[6pt]
\normalfont\small $^{1}$The University of Manchester \quad $^{3}$University of Chinese Academy of Sciences \\
\normalfont\small $^{2}$UKRI AI Centre for Doctoral Training in Decision Making for Complex Systems \\
\normalfont\small $^{4}$Zhongguancun Academy \quad $^{5}$NVIDIA \quad $^{6}$Honda Research Institute Europe \quad $^{7}$XinZhuAI \\
\normalfont\small $^{\ddagger}$Equal contribution. \quad $^{*}$This work was conducted during an internship at NVIDIA. \\
\normalfont\small $^{\dagger}$Corresponding authors: yuewang@bza.edu.cn, liuyang22@ucas.ac.cn, \\
\normalfont\small richard.allmendinger@manchester.ac.uk%
}

\iclrfinalcopy
\begin{document}
\maketitle
\ificlrfinal\fancyhead[L]{An important Bottleneck in LLM Optimizers: Principle-Level Local Optima}\fi

\begin{abstract}
    Feedback-driven LLM optimizers are increasingly used for evolutionary search, Bayesian optimization, heuristic design, and scientific discovery. Yet rapid initial gains often give way to prolonged plateaus despite substantial remaining evaluation budgets. We investigate \emph{principle-level search collapse}, where an optimizer's search becomes confined to a narrow set of design principles, even though it may continue to generate distinct candidates and achieve small improvements. We assemble a diagnostic benchmark spanning molecular optimization, network sentinel-node selection, and GPU-kernel optimization, and introduce a decision-tree protocol combining principle narrowing, prolonged performance plateaus, and gaps to attainable reference solutions. Across 161 assessable runs on the three problem domains, 78.3\% receive support for collapse, including 55.3\% with strong support. Candidate-level variation can nevertheless conceal restricted search at the principle level. We then test \emph{diversity injection}, a common response to search stagnation, such as higher t, multiple LLMs, and web search, and show that it provides limited and inconsistent benefits: among 65 molecular runs, still 73.8\% and 53.8\% receive support and strong support respectively. We finally test reference-principle interventions and find that they improve some runs, but their benefits vary across models and tasks: access to a better principle does not guarantee its effective use. Our benchmark and diagnostic protocol provide a basis for evaluating future optimizers on principle-level collapse. Our findings highlight the need for online collapse detection and better mechanisms for generating and maintaining alternative principles during optimization.
\end{abstract}

\section{Introduction}
\label{sec:intro}

Large language models are increasingly used as iterative optimizers: they propose candidates, receive evaluator feedback, and use that feedback to guide the next proposals. This pattern appears in evolutionary and Bayesian search, heuristic design, code optimization and open-ended scientific discovery \citep{ran2025exllm, agarwal2025searching, ye2024reevo, novikov2025alphaevolve, agarwal2026autodiscovery, qu2026coral}. However, the ability to improve from feedback raises a deeper question: when an LLM optimizer stops improving, has it exhausted the search space, or has it become trapped by the \emph{principles} that made its early improvements possible? Recent work has begun to report versions of this behavior: LLM optimizers often exploit early successes, converge quickly, and then plateau in ways that are difficult to escape \citep{redko2026prior, wiedemann2026kernelfoundry, li2026apex, ran2025exllm}, but what exactly has collapsed, and at which level, remains undiagnosed.

We found that this collapse often happens at the level of design principles rather than at the level of surface candidates. We define a \emph{principle} as a design assumption behind a family of candidates, such as ``modifying a known molecular scaffold", ``selecting nodes with a particular network role", or ``split the KV range into parallel blocks and merge partial results". This failure can be hard to see because candidate-level diversity can remain healthy: molecules may remain diverse under Tanimoto-based measures, code may not repeat verbatim, and node sets may continue to change. Yet these different candidates can still instantiate a narrow set of underlying principles, and these principles stop changing, causing the search to become stuck in a local optimum. We call this failure \textbf{principle-level search collapse}; Section~\ref{sec:definitions} makes the two levels precise and Section~\ref{sec:diagnostic-rule} states how we diagnose it.

The usual response to a plateau is \emph{diversity injection}: we define \emph{diversity injection} as methods that try to broaden the candidate pool through more random search or external source, such as higher temperature, external search, multiple LLMs, or other sources of variation \citep{wang2026towards, ran2025mcce, qu2026coral}; in our results, these methods can increase candidate diversity but do not change which principle dominates the retained solutions. Because they are still exploring the upper bound of the existing prior, their benefits are limited, especially for open-ended scientific discovery problems where good answers cannot be found in existing sources. Methods that revise principles in response to anomalies face a different problem: a poorly performing search direction can produce feedback consistent with predictions, so stagnation need not trigger revision \citep{pu2026principle}. Furthermore, we show that even supplying a better principle is a non-trivial operation because it needs to consider the current population and generalizability across different LLMs. 

\paragraph{Contributions:}
(i)~\textbf{A diagnostic framework.} We characterize principle-level search collapse and introduce a decision-tree protocol that combines principle narrowing, prolonged performance plateaus, and gaps to attainable reference solutions. This framework distinguishes evidence of harmful restriction from benign convergence without treating candidate diversity alone as evidence of continued exploration (\S\ref{sec:framework}).
(ii)~\textbf{A cross-domain diagnostic benchmark.} We assemble a reusable benchmark spanning molecular optimization, network sentinel-node selection, and GPU-kernel optimization, covering multiple optimizer designs and LLM backbones. The results reveal that principle-level restriction can persist beneath continued candidate variation, providing a basis for evaluating whether future optimizers alleviate this failure (\S\ref{sec:protocol}). We provide open-source code for the benchmark and analysis.\footnote{https://github.com/HuskyNian/LLM-Principle-Collapse}
(iii)~\textbf{Evidence on how restricted search persists.} We examine how weak first attempts are associated with less subsequent investment in kernel search, while PiEvo's prediction-error-based updates can leave poorly performing but predictable directions unrevised (\S\ref{sec:why}).
(iv)~\textbf{An evaluation of current remedies.} We systematically test diversity injection and reference-principle interventions. Diversity injection offers limited and inconsistent benefits, while access to a better principle does not guarantee its effective use. These findings motivate online collapse detection and better mechanisms for generating and maintaining principles (\S\ref{sec:fix-diversity}, \S\ref{sec:fix-hint}).

\section{Related Work}
\label{sec:related}

\paragraph{Exploration, Prior Dependence, and Diversity Collapse.}
Exploration collapse in self-evolving agents \citep{li2026apex}, semantic diversity collapse in multi-agent ideation \citep{chen2026diversity}, prior-driven rather than feedback-driven behaviour in code optimization \citep{redko2026prior}, and mode collapse in GPU-kernel evolution \citep{wiedemann2026kernelfoundry} all report that LLM search narrows. These works point to a recurring narrowing behavior in LLM search. Although KernelFoundry and K-Search \citep{cao2026k} show that preserving routes can be useful, our focus is different. We treat narrowing as a failure mode to be diagnosed rather than only as a behavior to be mitigated architecturally. In particular, we ask which level has collapsed: the surface candidates, the principles being proposed, or the principles that survive selection and condition future search. A run can continue generating diverse candidates while surviving principles have already narrowed. We also separate benign convergence from local optima by comparing the narrowed trajectory to a third-party reference solution. This points to a concrete research direction: optimizers should detect principle-level locking online and should have mechanisms for proposing and maintaining alternative principles during the run.

\paragraph{LLMs as feedback-driven optimizers.}
LLM-in-the-loop search appears as evolutionary optimization with LLM operators \citep{ran2025exllm, ran2025mcce, novikov2025alphaevolve, ye2024reevo}, LLM-assisted Bayesian and hierarchical optimization \citep{agarwal2025searching, schwanke2026improving}, and open-ended discovery driven by surprise \citep{agarwal2026autodiscovery}. These works propose methods; they do not diagnose trajectory-level failure. We study a failure mode of the structure they share, and we cover dominant families: LLM-as-optimizer in evolutionary loop and Bayesian Optimization, and with explicit principle space modeling. Recently, there are principle-aware methods try to address the this object, but is incomplete. They still depend on the coverage and revision of the principle space: a predefined or early principle set that misses a high-ceiling direction can itself become the search bottleneck, and an anomaly-based revision rule may not expand it when poor performance remains consistent with the model's predictions \citep{pu2025piflow,pu2026principle}. Our PiEvo experiments in \S\ref{sec:fix-principle-explicit} evaluate this boundary empirically.

\paragraph{Quality-diversity and behavioural descriptors.}
Our descriptor is a behaviour descriptor in the sense of quality-diversity (QD) search \citep{mouret2015illuminating, pugh2016quality}, a concept established in the evolutionary computation literature, and our candidate-versus-principle distinction is the genotype-versus-behaviour distinction of \citet{mouret2012encouraging}. More broadly, this concentration has a classical analogue in evolutionary computation: Holland’s schema theorem describes how above-average schemata can be amplified under selection \citep{holland1992adaptation}. The self-reinforcement of a low-ceiling principle is the deception problem that motivated novelty search \citep{lehman2011abandoning}. Descriptors need not be hand-specified: AURORA learns them \citep{cully2019autonomous}, QDAIF and ELM let a language model define or apply them \citep{bradley2024qdaif, lehman2022evolution}. A QD archive is built to stop selection from erasing already discovered behavioural niches: once a cell is filled, a better candidate from another cell cannot overwrite it. That protection applies only to the descriptor used to define the grid, and only to niches the search has already reached. Mainstream LLM optimizers typically apply selection directly to surviving candidates and keep no archive at the principle level, so concentration at that level is the expected outcome.

\section{Measuring Principle-Level Collapse}
\label{sec:framework}

This section makes the failure measurable. Section~\ref{sec:definitions} defines principles and candidates. Section~\ref{sec:diagnostic-rule} states the three conditions under which we diagnose principle-level collapse and how they combine into a verdict. Section~\ref{sec:instantiation} describes how we observe the principles an optimizer keeps on each task, and Table~\ref{tab:instantiation} gives the task-specific values.

\subsection{Principles and candidates}
\label{sec:definitions}

A \emph{candidate} is one concrete solution the optimizer emits and the evaluator scores: a kernel source file, a molecule, a set of sentinel nodes. A \emph{principle} is the design idea a family of candidates shares. On GPU kernels, ``pad the keys and values into one batched matrix multiply'' and ``split the KV range into parallel blocks and merge the partial results'' are two principles; the dozens of kernels that vary tile sizes under either are candidates of the same principle. On molecules an example principle is the scaffold kept while substituents change; on sentinel-node selection it is the pool of nodes a run keeps drawing from. We model the loop as sampling candidates from a mixture over principles,
\begin{equation}
x_{t+1} \sim q(x \mid H_t) = \sum_{p \in \mathcal{P}} q(x \mid p, H_t)\, P(p \mid H_t),
\label{eq:mixture}
\end{equation}
where $H_t$ is the history available at step $t$. A candidate-level local optimum means the candidates stop changing; principle-level collapse means the weights $P(p\mid H_t)$ have concentrated on a few principles and stopped moving while candidates within them can still change.

\subsection{Diagnosing principle-level collapse}
\label{sec:diagnostic-rule}
We diagnose principle-level collapse using three complementary conditions:
\textbf{(1)A substantial reference gap.} The run remains well below an attainable reference solution, indicating scope for improvement rather than convergence to a known optimum.
\textbf{(2)Principle narrowing.} Search concentrates on a restricted set of principles, measured through task-specific structural descriptors or, when available, an explicit model of the principles used during search.
\textbf{(3)A long performance plateau.} Best-so-far performance makes little subsequent progress despite substantial remaining evaluation budget.

\begin{figure}[t]
\centering
\includegraphics[width=\linewidth]{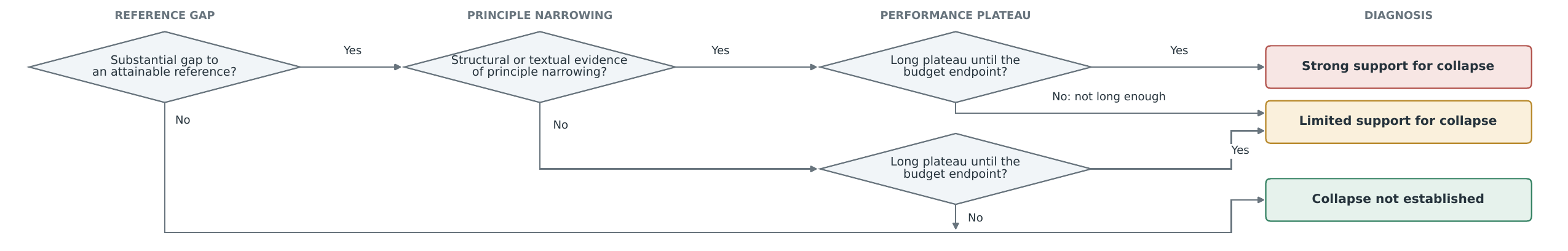}
\caption{Overview of the collapse diagnosis. Narrowing may be supported by structural metrics or an eligible audit of principle texts. A long plateau leaves at least $30\%$ of planned budget. The lower branch summarizes limited narrowing evidence. Text-audit eligibility and exceptions are specified in the full protocol (Appendix~\ref{app:decision-counts}, Figure~\ref{fig:collapse-decision-tree-full}).}
\label{fig:collapse-decision-tree}
\end{figure}

No single condition establishes collapse. Let $F_T=\max_{t\le T}f(x_t)$ and $\Delta_T=f(x_{\mathrm{ref}})-F_T$. We denote a substantial reference gap by \refstepcounter{equation}
\label{eq:headroom}
$\Headroom=[\Delta_T\ge\gamma]$~(\theequation) and write $\Narrow$ for structural evidence of principle narrowing. Narrowing is measured on the diagnostic set $\mathcal A(b)$, the candidates the optimizer keeps at budget $b$ (Table~\ref{tab:instantiation}). A post-hoc descriptor assigns each valid candidate one structural label, $\phi:\mathcal X^{\mathrm{valid}}\to\mathcal P$. Its empirical distribution in the diagnostic set $\mathcal A(b)$ is
\begin{equation}
\hat\pi_b(p)=\frac{|\{x\in\mathcal A(b):\phi(x)=p\}|}{|\mathcal A(b)|}.
\label{eq:label-frequencies}
\end{equation}
Effective count $N_{\mathrm{eff}}=\exp(-\sum_p\hat\pi_b(p)\ln\hat\pi_b(p))$ measures effective scaffolds or routes. This metric accounts for both the number of unique scaffolds and their frequency: for example, if two scaffolds occur among 50 candidates but one appears 49 times and the other only once, $N_{\mathrm{eff}}$ remains close to 1 rather than 2, reflecting the strong concentration on a single scaffold. $\Narrow$ holds when the final $N_{\mathrm{eff}}$ is at most the task cutoff in Table~\ref{tab:instantiation}. $\mathrm{Plateau}$ holds when the run's best result stops improving while a substantial share of the budget is still unspent (cutoffs in Table~\ref{tab:instantiation}). Strong support requires all three conditions:
\begin{equation}
\mathrm{Strong}=\Headroom\wedge {\mathrm{Plateau}}\wedge
\Narrow\;.
\label{eq:verdict}
\end{equation}
If $\Headroom$ fails, collapse is not established: the run may have converged to a competitive solution. If $\Headroom$ holds but only one of $\Narrow$ and $\mathrm{Plateau}$ does, the run receives limited support.
For an optimizer that keeps explicit principle texts (PiEvo here), an audit of those texts can also establish $\Narrow$ (Appendix~\ref{app:diagnostic-details}). Incomplete evidence yields limited support or collapse not established, instead of a claim that search remains diverse. Figure~\ref{fig:collapse-decision-tree} summarizes the rule. Because the latent principle distribution $P(p\mid H_t)$ is not directly observable, we use these structural measurements as observable proxies for principle diversity and concentration.

\subsection{Instantiation on the three tasks}
\label{sec:instantiation}

\begin{table}[h]
\centering
\caption{Instantiation of the framework on the three tasks. Task-specific structural descriptors and references complement the graded decision tree in Figure~\ref{fig:collapse-decision-tree}.}
\label{tab:instantiation}
\scriptsize
\resizebox{\linewidth}{!}{
\begin{tabular}{@{}lp{4.6cm}p{4.2cm}p{4.2cm}@{}}
\toprule
 & \textbf{GPU kernel} & \textbf{Molecular optimization} & \textbf{Sentinel-node selection} \\
\midrule
Candidate $x$ & one PyTorch/Triton source file & SMILES string & node subset of size $r$ (4 or 5) \\ \midrule
Utility $f$ & $-\ln$(geometric-mean latency) & scalarized five-objective score & $-\ln$(approximation error) \\\midrule
Descriptor $\phi$ & executed route, blind LLM annotation against a per-problem rubric (6--9 leaf routes + families) & Bemis--Murcko scaffold & node incidence in selected sets \\\midrule
Diagnostic set $\mathcal A$ & fastest 10 valid kernels & top 50 distinct molecules & top 50 distinct node sets \\\midrule
Candidate-level diversity & 5-gram Jaccard distance between source files & fingerprint diversity, uniqueness & set uniqueness \\\midrule
Reference $x_{\mathrm{ref}}$ & FlashInfer library kernel, same GPU & best known molecule, $F=4.6985$ & exhaustive optimum (Dolphin); published optimized set (residence hall) \\\midrule
Headroom threshold $\gamma$ & $\ln 2$ (best latency $\ge 2\times$ expert) & top-1 below $4.3$ & $\ln 2$ (error $\ge 2\times$ reference) \\\midrule
Narrowing cutoff (final $N_{\mathrm{eff}}$) & $\le 2$ routes & $\le 3$ scaffolds & $\le 4\times$ the distinct-set lower bound ($31.4$ / $31.9$ nodes) \\\midrule
Plateau tolerance (later gain allowed) & $\le 10\%$ of the remaining gap to the expert & $\le 0.01$ in Top-1 fitness & $\le 1\%$ relative error reduction \\\midrule
Negative control & RMSNorm & --- & Dolphin (exhaustive reference) \\\midrule
Intervention & L1/L2/L3 $\times$ \{onset, start\} & L2 $\times$ \{40\%, start\} & L2 $\times$ \{40\%\} \\
\bottomrule
\end{tabular}}
\end{table}

Table~\ref{tab:instantiation} instantiates the framework across tasks. We measure search concentration through task-specific structural descriptors: molecular scaffolds, node incidence in sentinel sets, and executed kernel routes. For molecules and networks, we reconstruct a diagnostic set $\mathcal A_{50}(b)$ of top 50 distinct candidates from the archive at evaluation budget $b$. Kernels use the top 10 candidates because of smaller budget and population size. Networks instead normalize node-occurrence counts across top 50 unordered sets, since each candidate contains several nodes instead of a single label; applying the same entropy-based measure to node frequencies gives effective nodes, not a literal number of principles, so the network cutoff is stated relative to the lower bound that 50 distinct sets impose. References are the best molecule reported across 100 runs by ExLLM, the authors' published residence-hall set, and the expert library kernel timed on the same GPU. PiEvo keeps explicit principle texts, so its narrowing is additionally audited on its principle texts. We additionally report molecular scaffold richness, dominant share and Morgan-fingerprint diversity; network normalized top-five coverage and pair/triple reuse; and kernel dominant-route share and source-code 5-gram Jaccard diversity. Appendix~\ref{app:population-metrics} gives the formulas, bounds and plotting conventions. We use task-specific working thresholds, including at least $30\%$ remaining budget for a long plateau. Detailed thresholds and textual-audit rules are in Appendix~\ref{app:diagnostic-details}; the full decision tree and classifications are in Appendix~\ref{app:decision-counts}. Threshold sensitivity is discussed in Appendix~\ref{app:sensitivity}.

\section{Results}
\label{sec:protocol}
\paragraph{Tasks.} 
\textbf{GPU kernel optimization} uses four operators from DeepSeek-V3 and Qwen3 inference on FlashInfer-Bench \citep{lin2026sol} (GQA paged decode, MLA paged decode, MLA paged prefill with causal mask, FP8 block-scale MoE) plus RMSNorm as a negative control. A candidate is one self-contained PyTorch/Triton file; the official harness compiles it, checks correctness on every workload and reports geometric-mean latency on a data-center GPU with locked clocks. The library is not installed in the evaluation container, so candidates cannot reach it by delegation. We also adopt a five-objective \textbf{Molecular Optimization} from prior LLM optimizers \citep{ran2025exllm, ran2025mcce, wang2025efficient}: maximize QED and JNK3 activity while minimizing SA, DRD2 and GSK3$\beta$; candidates are SMILES strings. Reference molecules have scores $>$4.6 and the best one achieves 4.699 from ExLLM. \textbf{Sentinel-node selection} follows \citet{maclaren2025observing}: choose a small node subset whose mean state tracks the mean state of a networked dynamical system, on the Dolphin social network ($N=62$, $r=4$, double-well dynamics) and a residence-hall network ($N=217$, $r=5$, gene-regulatory dynamics). The Dolphin instance is small enough to enumerate, so its reference is the exhaustive optimum ($4.041\times10^{-3}$); for the residence hall, with $3.8\times10^{9}$ subsets, the reference is the authors' published optimized set ($2.3818\times10^{-7}$). Except for Dolphin and RMSNorm, we study these tasks as open-ended scientific optimization problems without known global optima. We therefore use independently obtained reference solutions to assess the remaining scope for improvement. For molecular optimization and Residence Hall, the references are the best solutions reported across 100 runs in the original studies; for the remaining kernel tasks, they are expert implementations reported in the corresponding papers. These references are not guaranteed to be globally optimal. They demonstrate performance achieved with finite computational or human effort under a fixed evaluation budget. We use the reference gaps to contextualize search stagnation and principle interventions. Evaluation budgets are 200 for each kernel task, 5,000 for molecules, 3,100 for Dolphin, and 10,850 for residence hall following the corresponding studies.

\paragraph{Optimizers.}
We select optimizers to cover LLM-as-optimizers in evolutionary search, reflective optimization, and Bayesian optimization. ExLLM \citep{ran2025exllm} combines evolutionary search with a compact, evolving experience mechanism and reports state-of-the-art results across multiple domains. ReEvo \citep{ye2024reevo} represents reflective evolutionary optimization, while MCCE \citep{ran2025mcce} combines multi-LLM collaboration with RL-based adaptation of a trainable optimizer. PiEvo \citep{pu2026principle}, introduced at ICML 2026, is also a LLM optimizer performing Bayesian optimization over an evolving principle space and revising principles in response to detected anomalies. For ExLLM, we use GPT-5.4, Claude Sonnet~5, Qwen3.5-35B-A3B, and DeepSeek V4 Flash: two proprietary and two open-weight backbones, to assess whether the findings persist across different LLMs. Together, these settings allow us to test whether principle-level collapse is broadly observed across different LLM-optimizer algorithms, LLM backbones, and problem domains. We additionally evaluate GPT-4o on molecules and kernels to compare older and newer backbones. This comparison is motivated by the substantially higher molecular uniqueness reported for GPT-4o in ExLLM; the molecular findings are discussed in Appendix~\ref{app:backbone-controls}, with kernel results in Appendix~\ref{app:kernel-results}.

\paragraph{Intervention protocol.}
All tasks include an \textbf{L2 intervention}, which provides a specific design principle underlying reference solutions without revealing the candidates themselves. For kernel optimization, we additionally test two intervention strengths. \textbf{L1} provides only a general warning that the search may have settled prematurely on one decomposition and encourages exploring a fundamentally different one, without naming a specific alternative. \textbf{L3} supplements L2 with an implementation outline, including grid organization, tiling, and data layout, but provides no code.

\subsection{Diagnosing Collapse and Its Severity}
\label{sec:collapse}

We ask three questions in the result section: (1) how common and severe is principle-level collapse, (2) can diversity injection resolve it, and (3) does supplying a better principle directly improve performance? 
\begin{figure}[t]
\centering

\includegraphics[width=.33\linewidth]{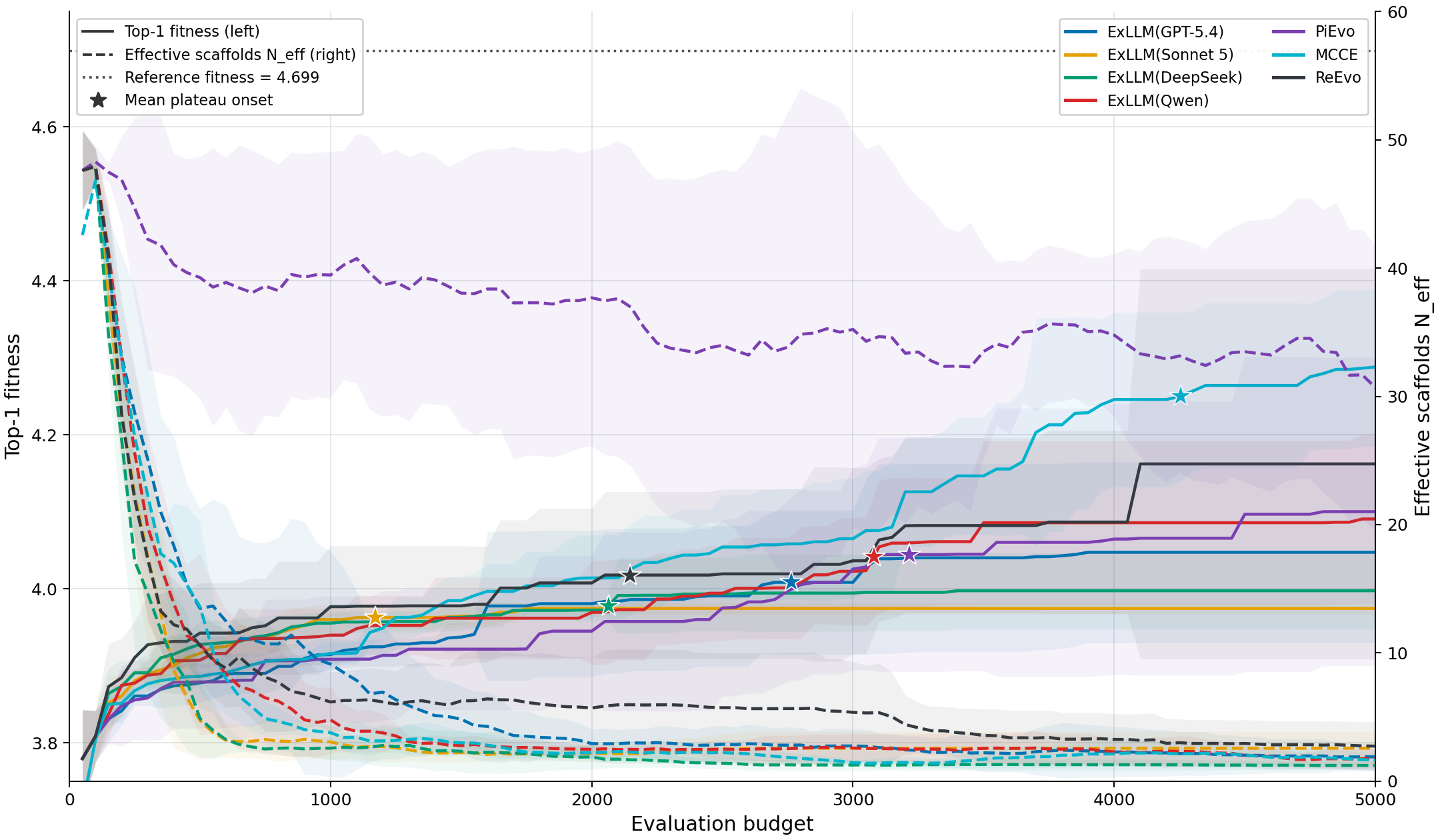}\hfill
\includegraphics[width=.33\linewidth]{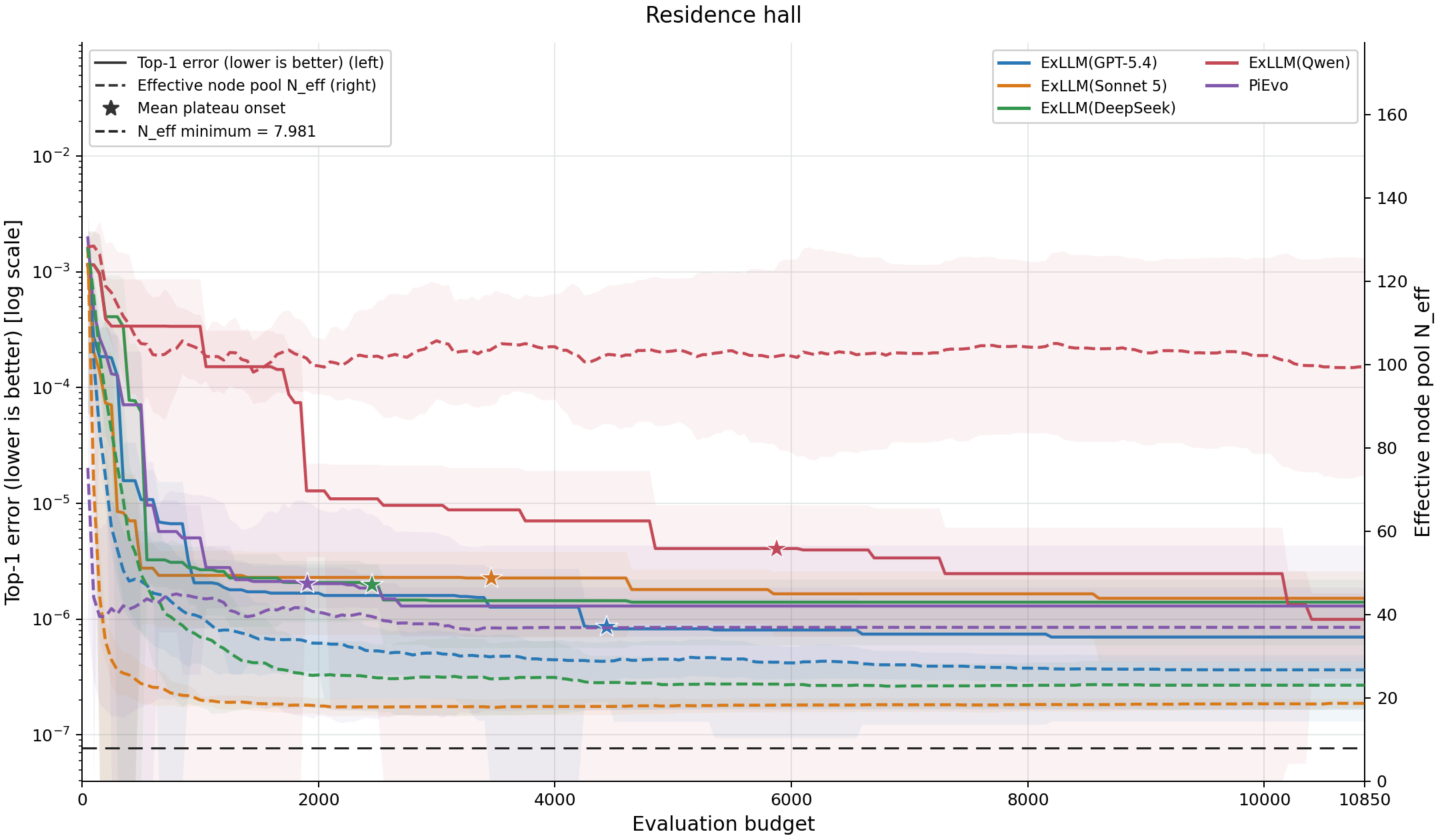}\hfill
\includegraphics[width=.33\linewidth]{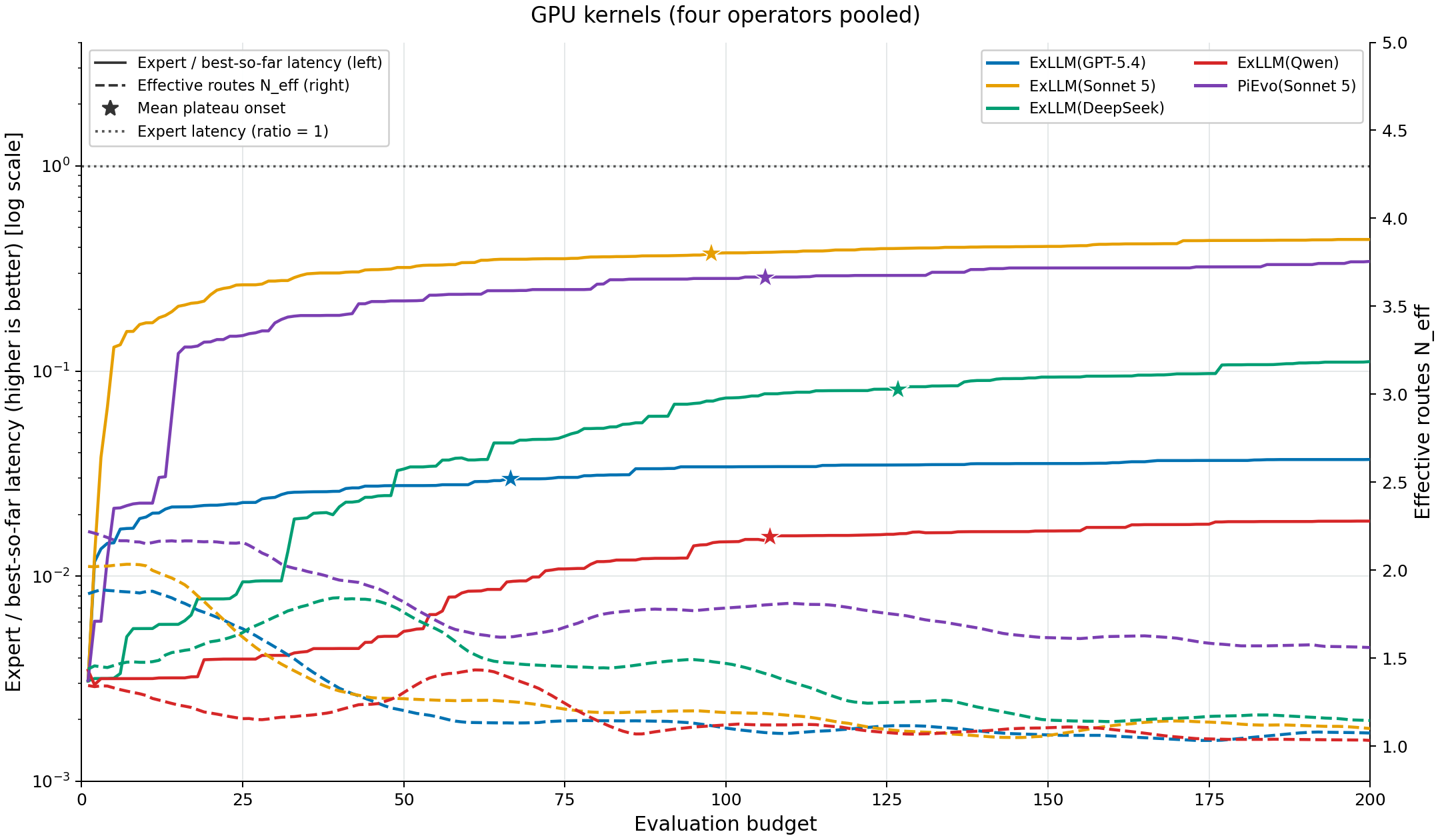}
\caption{Best performance and effective count: molecules (left), residence hall (middle) and GPU kernels (right; the four operators pooled). Solid/dashed lines use the left/right axes. Stars mark mean plateau onset. The molecular reference is $4.699$; the dashed line in the network is the lower bound of $N_{\mathrm{eff}}$; the kernel panel shows expert latency divided by best-so-far latency (expert $=1$) as the geometric mean across runs, and $N_{\mathrm{eff}}$ in the diagnostic set of the ten fastest kernels; per-operator curves are in Figure~\ref{fig:kernel-budget}.}
\label{fig:teaser}
\end{figure}

\paragraph{Principle-level collapse}
We report diagnostic counts as strong support / limited support / not established (S/L/N). Including GPT-4o, the molecular comparison covers five ExLLM backbones, PiEvo, MCCE and ReEvo, giving $23/13/4$ across 40 runs. Residence hall gives $13/6/6$ and Dolphin $0/0/25$, each across 25 runs. On residence hall, the four ExLLM backbones have mean best errors $2.94$--$6.36$ times the published reference. Sonnet, DeepSeek and GPT-5.4 reduce their effective node pools from $72$--$119$ to $19$--$27$. Their saved experiences contain fixed pools or anchors, such as \emph{keep 143 fixed} (Appendix~\ref{app:experience}). Qwen retains a broad effective node pool of about $99$ despite its performance gap. Its S/L/N count is $0/4/1$: four runs provide limited evidence of harmful stagnation. The molecular comparison additionally includes MCCE and ReEvo. For the four primary ExLLM backbones, all 20 molecular runs finish below $4.3$; their mean plateau onsets fall between $23\%$ and $62\%$ of budget. Their mean effective scaffold counts fall from $39$--$43$ to $1.24$--$2.58$. MCCE and ReEvo also concentrate strongly, ending at $1.66$ and $2.74$ effective scaffolds on average. Their performance qualifications differ: MCCE's mean plateau begins at $85\%$ of budget, and one ReEvo seed reaches $4.579$. Limited support reflects either a late plateau or incomplete evidence of narrowing. 

Kernel search shows the same restriction at the level of executed decompositions (Figure~\ref{fig:teaser}, right, and Figure~\ref{fig:kernel-budget}). We measure route frequencies among the top 10 kernels found so far. Across the four main ExLLM backbones and four different kernels, 64/80 runs finish with at most two effective routes while remaining at least twice as slow as the expert. Of these, 48 also plateau with at least $30\%$ of their evaluation attempts remaining: 5/20 on GQA decode, 9/20 on MLA decode, 19/20 on MLA prefill and 15/20 on FP8 MoE. Here a plateau allows subsequent gains of at most $10\%$ of the remaining gap to the expert; task-specific definitions and sensitivity tests are in Appendix~\ref{app:kernel-results}. The other 16 concentrated, large-gap runs plateau too late to meet this joint criterion. Concentration is widespread, but its performance cost and duration vary by task and backbone.

Together, these results identify principle-level collapse as a recurring limitation across tasks, LLM backbones and optimizer designs, warranting more attention to this bottleneck.

\begin{figure}[t]
\centering
\includegraphics[width=\linewidth]{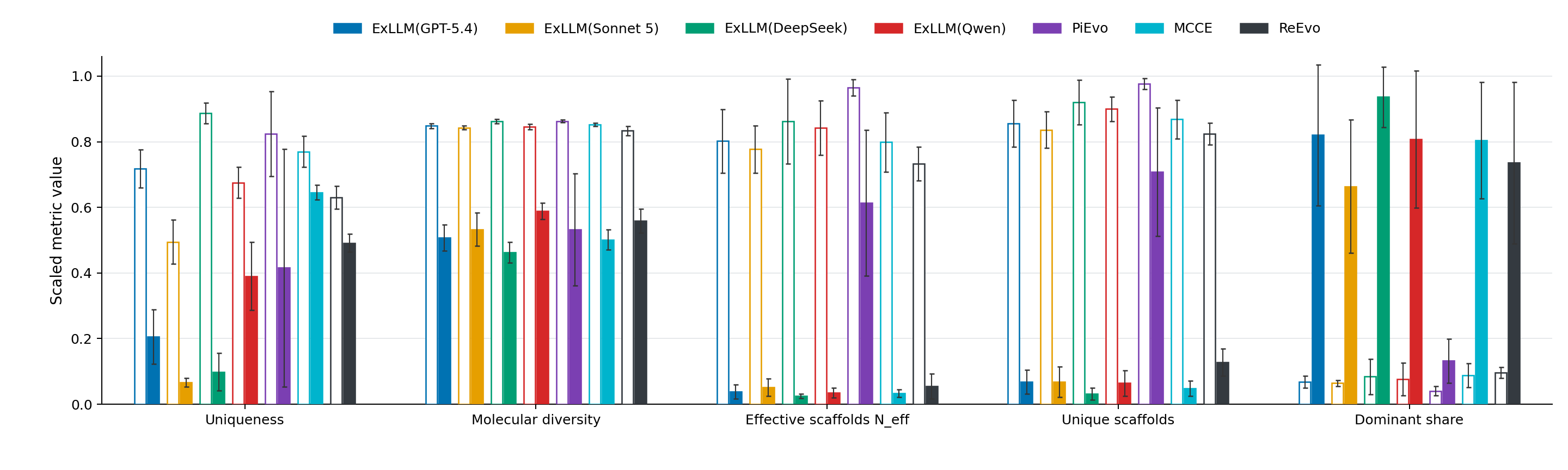}
\caption{Molecular early/late diagnostics. Hollow/filled bars are early/final values. Uniqueness uses the first/last 1,000 generated candidates. Other metrics use reconstructed 50-candidate sets. Effective and unique scaffold counts are divided by 50.}
\label{fig:verdict}
\end{figure}

\paragraph{Candidate variation may conceal principle restriction.}
Figure~\ref{fig:verdict} shows marked declines in uniqueness, scaffold richness and effective scaffold count, alongside increasing dominant-scaffold share. Mean molecular fingerprint diversity remains $0.46$--$0.59$ across the four main ExLLM backbones, even as dominant share reaches $0.66$--$0.94$. Candidate diversity also decreases, but its still-moderate values can obscure how few scaffolds account for most selected molecules. Network diagnostics measure normalized coverage of the five most-used nodes and reuse of node pairs or triples; higher values indicate concentration. All are computed within seeds, not from overlap between seeds. Supplementary trajectories and network bars show that repeated structures and node combinations become more prevalent, although the extent differs across backbones (Appendix~\ref{app:collapse-details}, Figures~\ref{fig:mol-extra} and~\ref{fig:dyn-collapse}).
For kernels, the final effective route count averages only $1.03$--$1.14$ across the four backbones, with dominant-route shares of $0.96$--$0.99$. Yet mean code diversity, measured by pairwise token 5-gram Jaccard distance, remains $0.55$--$0.72$ for GPT-5.4, DeepSeek and Qwen. Sonnet's code diversity falls more sharply, to $0.31$. Thus changing source code need not change the computational decomposition. Appendix~\ref{app:kernel-results} shows the early/late comparisons 

\paragraph{First-attempt quality and subsequent development.}
\label{sec:why}
A route that first appears with a weak implementation tends to receive little further development. For every route that enters a kernel run after the run's first route, we compare its first kernel with the run's best latency at that moment and count the kernels the run later spends on it (Figure~\ref{fig:mechanism}). Routes whose first kernel already beats the incumbent are developed for dozens of further kernels; routes that start behind receive a handful, and those that start far behind are often never tried again. Restricting to routes that appear with ample budget left does not change this ordering, so late arrival does not explain it. The expert's route is no exception: it almost always enters slower than the incumbent, and whether the run then keeps investing in it decides whether it ever reaches expert speed. These are associations between first-attempt quality and later investment, not a measured causal effect of selection (Appendix~\ref{app:mech}).

\paragraph{Concentration is not always failure.}
Dolphin provides a reference with a different status: exhaustive enumeration establishes its optimum. PiEvo's five best errors are within $0$--$4.4\%$ of it, including one exact optimum; the ExLLM mean errors are also within $5\%$. Restriction here need not imply a costly missed direction. RMSNorm plays the analogous role for kernel search, although beating an expert implementation does not prove global optimality.
All 25 ExLLM RMSNorm runs, including the GPT-4o control, finish at $0.437$--$0.480$ times the expert latency; 24 also have at most two effective routes. None meets the large-gap collapse criterion. These two controls distinguish benign convergence toward competitive solutions from harmful narrowing that leaves substantial attainable gains unexplored (Appendix~\ref{app:collapse-details}, Figure~\ref{fig:negative-controls}).

\paragraph{PiEvo: coverage without continued development.}
\label{sec:fix-principle-explicit}
PiEvo's final molecular top-50 retains $30.69$ effective scaffolds on average. Yet broader structures can implement the same strategy. In seed 42, search plateaus at budget 1,295 with Top-1 $3.898$, leaving a $0.800$ reference gap and performing worse than the other methods' mean endpoints. A macrocycle-oriented strategy accounts for $90.4\%$ of the last 1,000 generated attempts; candidate inspection links explicit macrocycle instructions to molecules containing 8--15-member rings. Thus broad scaffold coverage can coexist with restricted strategy use and substantial performance headroom. Across molecular seeds, mean principle-generation/update counts fall from $3.8$ in the first 1,000 generated attempts (after the ten forced principle-generation warm-up rounds) to $1.0$ in the last 1,000 attempts. The late windows also average only one anomaly-triggered round. Combining structural evidence with the eligible textual audit gives PiEvo S/L/N counts of $2/3/0$ for molecules and $4/1/0$ for residence hall.

PiEvo's anomaly detector measures GP prediction error. A poorly performing direction can therefore produce feedback consistent with the model's predictions and persist without triggering revision. This is a limitation of using surprise to guide principle updates. The saved-record audit also identifies rounds with too few scored candidates to satisfy the trigger condition. Together, these results show that PiEvo can preserve structural diversity without reliably sustaining the development of alternative principles. Maintaining several directions is insufficient if they remain unproductive and no better direction is developed. Appendix~\ref{app:pievo-details} reports the text, candidate and detector evidence, including the additional $\sigma=0.9$ PiEvo experiment for comparison.

On kernels, PiEvo(Sonnet) reduces its mean effective route count from $2.44$ to $1.51$. Three of its 12 runs meet the joint collapse criterion; seven have concentrated routes but a latency ratio below two, one has a large gap but plateaus late, and one remains unconcentrated. PiEvo(Gemini) has fewer than 20 valid kernels in 10/12 runs, so those runs cannot support a reliable concentration verdict. Explicit principle modeling therefore neither guarantees escape nor implies failure in every run; valid-candidate coverage must also be reported.

\subsection{Can Diversity Injection Resolve Collapse?}
\label{sec:fixes}
\label{sec:fix-diversity}
\begin{figure}[t]
\centering
\includegraphics[width=\linewidth]{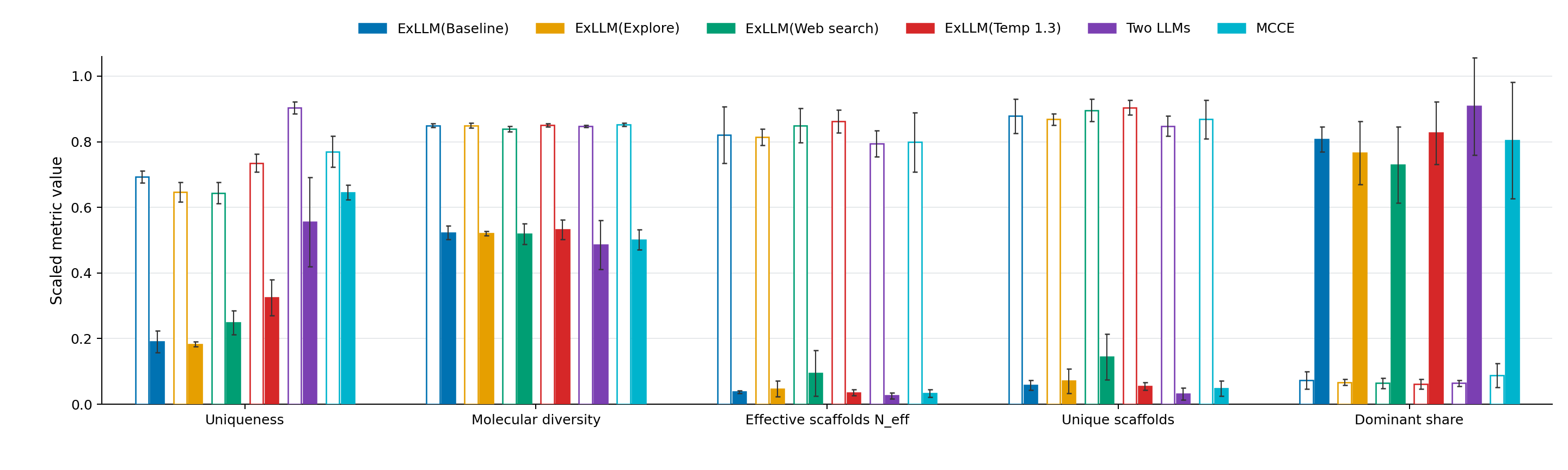}
\caption{Diversity-injection diagnostics, with the definitions and scaling of Figure~\ref{fig:verdict}. ExLLM variants average backbones within each seed. Two LLMs and MCCE are independent five-seed comparisons. Broader generation does not reliably preserve a broad selected scaffold distribution.}
\label{fig:diversity-bars}
\end{figure}

A natural response to prolonged stagnation is diversity injection: broadening candidate generation in the hope of finding a more productive direction. We test whether such changes prevent or reduce principle-level collapse. On molecules, we test higher temperature ($1.3$), parent-independent proposals (crossover:mutation:explore $=7:2:1$), external literature/database search, and joint proposing by two LLMs. MCCE adds a comparison combining multi-LLM collaboration and reinforcement-based adaptation. Figure~\ref{fig:diversity-bars} averages the four ExLLM backbones within each seed to summarize the common concentration trend; temperature uses three because Sonnet does not support this setting. Per-backbone plots and five-seed tables are in Appendix~\ref{app:diversity-details}. Across 65 runs, the S/L/N counts are $35/13/17$. Among the 17 not-established runs, nine reach Top-1 $\ge4.3$; eight remain below $4.3$ but lack both structural concentration and a long plateau. The nine higher-scoring runs are scattered across methods and backbones, with no method--backbone combination reaching this threshold in all five seeds. Only four of the 65 runs exceed $4.6$. These gains are occasional rather than a reliable resolution of collapse; detailed counts are in Appendix~\ref{app:decision-counts}.

Mean Top-1 rises from $4.028$ for pooled ExLLM to $4.048$ with explore and $4.108$ with web search. Temperature reaches $4.086$ on its three-backbone subset. Two-LLM proposing and MCCE reach $4.165$ and $4.288$, with global uniqueness $0.636$ and $0.656$, but retain concentrated final scaffold distributions. MCCE delays its plateau: its S/L/N count is $0/3/2$, reflecting two endpoints above the large-gap cutoff and three late plateaus rather than consistently successful search. Overall, diversity injection introduces additional search variation and can yield occasional substantial gains, but its benefits are limited and inconsistent across seeds and backbones. This supports our hypothesis that broadening candidate generation alone does not reliably overcome principle-level collapse. These open-ended tasks have no known globally optimal answer to retrieve. Diversity injection broadens sampling from model priors and external knowledge, but does not itself ensure that the optimizer recognizes an unproductive principle or develops an alternative. The results are consistent with this limitation. Because the variants operate from the start, they test prevention or reduction of collapse, not recovery from an already diagnosed collapse. We next test whether supplying a specific reference-informed direction helps.

\subsection{Reference Principles Help, but Their Use Is Nontrivial}
\label{sec:fix-hint}

\paragraph{A better direction can improve an established trajectory.}
Using ExLLM with the four main backbones, we first supply L2 at $40\%$ of the molecular and network budgets; kernel interventions branch at detected narrowing onset. Hints name a search direction without giving a molecule, node set or kernel implementation (Appendix~\ref{app:hints}). On residence hall, mean best error decreases by $52.7\%$ for Qwen, $32.6\%$ for DeepSeek and $27.0\%$ for GPT-5.4; Sonnet does not improve on average. Dolphin improves little near its optimum. On molecules, delayed L2 raises mean Top-1 from $3.975$ to $4.485$ for Sonnet and from $3.998$ to $4.699$ for DeepSeek, but not for GPT-5.4 or Qwen. Success means improvement beyond the previous plateau. Full comparisons of metrics are in Figure \ref{fig:intervention-bars} and numbers are in Appendix~\ref{app:intervention-results}. For kernels, onset-L2 improves 13/41 matched runs, worsens nine and leaves 19 within a $\pm5\%$ band of paired gain (baseline latency divided by hinted latency). The geometric-mean gain is $1.010$, close to neutral. These are the four main backbones, restricted to trajectories for which the onset detector produced a branch; they are not a random subset of all baseline runs. Figure~\ref{fig:kernel-hints} separates adoption of the reference route from performance, and Appendix~\ref{app:kernel-results} reports all hint conditions.

\paragraph{Earlier injection does not remove model dependence.}
A delayed hint may arrive after earlier candidates and experience have restricted search. Providing L2 from the first molecular generation tests whether this timing alone explains poor outcomes. DeepSeek again reaches mean Top-1 $4.699$, and Sonnet improves further to $4.617$. GPT-5.4 improves to $4.264$ but remains below the reference region; Qwen falls from $4.091$ to $4.020$. Thus failures cannot be explained solely by delayed access: a hint from the start also does not ensure effective use. Kernel L2 from the start likewise gives mixed outcomes: 14/46 usable matched runs improve, 21 worsen and 11 stay within the gain band; geometric-mean gain is $0.833$. Thus late access cannot be the sole explanation for ineffective hints. This is not an isolated causal effect of timing: onset arms exist only for detector-triggered baselines, whereas start arms cover all four non-control tasks.

\paragraph{More specific instructions need not help more.}
Kernels additionally compare a general redirection warning (L1), a named reference strategy (L2), and a strategy plus implementation outline (L3) as explained above.
From the start, L1 has the largest pooled geometric-mean gain ($1.169$), compared with $0.833$ for L2 and $1.067$ for L3. L1 improves 28/48 matched runs but worsens 12; it is not a universal remedy. The ordering also depends on the statistic: at onset L3 improves more runs than L1 (16 versus 12 of 41), yet its geometric-mean gain is lower ($0.986$ versus $1.110$). Naming a route and supplying an implementation outline can increase adoption without producing a fast implementation. The route-conditioned results in Appendix~\ref{app:kernel-results} expose this remaining implementation gap.
These experiments separate receiving a promising direction from developing a competitive candidate. Effective principle generation and maintenance must support both.

\section{Conclusion}
\label{sec:agenda}
Across molecular, network and GPU-kernel optimization, our results identify principle-level collapse as a serious and easily overlooked bottleneck in feedback-driven LLM optimization. Distinct candidates and occasional small gains can coexist with restricted design principles, prolonged plateaus and substantial gaps to attainable references. Diversity injection offers limited and inconsistent benefits, while explicit principle modeling and access to better principles do not guarantee effective exploration. The challenge is not simply to generate more candidates, but to recognize when current principles no longer support meaningful progress. We call for research on online collapse detection and better mechanisms for generating, developing and maintaining alternative principles in scientific discovery tasks. We encourage future optimizers to use our benchmark to demonstrate reduced evidence of collapse alongside improved performance and smaller reference gaps. Progress should also be evaluated across LLM backbones and task domains to establish whether these mechanisms generalize beyond individual settings.

\section{AI Use Statement}
In this work, we did not use generative AI tools for any tasks requiring mandatory disclosure. In particular, generative AI was not used to develop theoretical or conceptual frameworks, formulate mathematical claims, propose or refine research hypotheses, design the research methodology or experiments, implement the proposed methods, analyze data, or interpret results.

We used generative AI tools for several tasks for which disclosure is recommended. These included assisting with the organization and structuring of the manuscript, including drafting and refining the paper outline; editing the manuscript to improve grammar, clarity, and readability; and assisting with non-core software development tasks. In particular, generative AI was used to help write auxiliary helper functions for data processing and visualization, review code for potential errors and completeness, and improve code organization and readability. These uses did not involve the design or implementation of the proposed methods or the scientific interpretation of experimental results.

All AI-assisted text and code were reviewed by the authors. AI-assisted code was manually inspected and tested for correctness before being used in the reported experiments or analysis.

We take full responsibility for the final content of this work, including all text, claims, code, figures, and other artifacts produced with the aid of generative AI.

\ificlrfinal
\section*{Acknowledgements}
This work is supported by the Engineering and Physical Sciences Research Council [EP/Y030826/1]. The supervisory team of Nian Ran is: Richard Allmendinger, Mauricio  Alvarez, Markus Olhofer, Sebastian Brulin. This work is supported by the Zhongguancun Academy (Grant No.~C20250203). We thank NVIDIA for access to the GPU resources and model APIs used in the GPU-kernel experiments. 

\section*{Author contributions}
Nian Ran conceived the study, designed and ran the molecular-optimization and sentinel-node experiments, and led the writing. Tianhe Jia designed and ran the GPU-kernel experiments, discussed with the main concepts, and co-wrote the paper. Zhengyang Mu supervised the GPU-kernel work at NVIDIA. Yue Wang and Yang Liu supervised Tianhe Jia at the Zhongguancun Academy and the University of Chinese Academy of Sciences. Richard Allmendinger and Mauricio A.~\'Alvarez supervised Nian Ran at The University of Manchester. Markus Olhofer and Sebastian Brulin co-supervised Nian Ran at Honda Research Institute Europe. Yang Liu and Richard Allmendinger are the corresponding authors.
\fi

\bibliography{iclr2027_conference}
\bibliographystyle{iclr2027_conference}
\appendix
\section*{Appendix}

\section{Limitations}
\label{sec:limitations}

The latent principle distribution is not directly computable or observable, so our diagnosis relies on proxies that introduce measurement uncertainty. Each task requires its own principle descriptors, metrics and thresholds, which limits direct comparison across domains. Applying the protocol to a new problem therefore may require identifying suitable descriptors and validating how well they capture meaningful differences in search direction, as well as calibrating the diagnostic thresholds.

\section*{Reproducibility Statement}
Our code is publicly available at \url{https://github.com/HuskyNian/LLM-Principle-Collapse}. Experimental settings and diagnostic procedures are detailed in the main text and appendices to facilitate reproduction of our results.

\section{Diagnostic Thresholds and Textual Audit}
\label{app:diagnostic-details}
The reference-gap indicator in Eq.~\ref{eq:headroom} uses task-specific boundaries: molecular Top-1 below $4.3$ (strictly), and network error at least twice the reference. These are working diagnostic thresholds, not universal definitions of failure. A feasible reference need not be globally optimal; it establishes attainable improvement.

Structural concentration, denoted $\Narrow$, uses the final selected-50 set:
\begin{equation}
\Narrow=
\begin{cases}
[N_{\mathrm{eff}}\le3],&\text{molecules},\\
[N_{\mathrm{eff}}\le4N_{\mathrm{eff,min}}],&\text{networks}.
\end{cases}
\label{eq:narrow}
\end{equation}
The distinct-set lower bounds are approximately $7.837$ for Dolphin and $7.981$ for residence hall, giving cutoffs $31.35$ and $31.92$. Node concentration is a structural proxy, not a literal count of principles. Kernel diagnostics instead use at most two effective routes among the fastest ten valid kernels; their current export and audit limitations are described in Appendix~\ref{app:kernel-results}.

Let $L=[(B-b_{\mathrm{plat}})/B\ge0.30]$, where $B$ is planned budget. The retrospective plateau begins when all subsequent Top-1 gains total at most $0.01$ on molecules, or the subsequent best-error reduction is at most $1\%$ on networks. 

If structural concentration is absent, textual auditing is permitted only when an explicit principle-space model covers the algorithm's actual search directions. PiEvo qualifies in this comparison; ExLLM, MCCE and ReEvo experience summaries do not represent all directions and are not audited as if they did. Coverage here does not mean that the model contains every possible useful principle. Over the last 1,000 generated attempts, including duplicates and invalid candidates, we merge wording variants and test:
\begin{itemize}
\item $T_1$: one semantic strategy accounts for at least $50\%$ of generated attempts, weighted by actual use rather than number of texts.
\item $T_2$: at most five minor, same-direction revisions, with no major direction change.
\item $T_3$: no new principle direction is proposed, including directions that are never adopted.
\end{itemize}
For eligible runs, the textual branch of Eq.~\ref{eq:verdict} is
\begin{equation}
\text{Principle narrowing in text}=T_1\wedge T_2\wedge T_3.
\label{eq:textual-restriction}
\end{equation}
This combines concentrated strategy use with little revision and no newly proposed direction. For other optimizers, the textual branch cannot establish strong support; it is inapplicable, not evidence that their principles remain diverse.
Figure~\ref{fig:collapse-decision-tree} summarizes the diagnosis in four steps: reference gap, principle narrowing, plateau duration and support level. Structural and textual evidence are two ways of assessing the same narrowing condition, not separate failure types. The simplified figure omits text-audit eligibility and exceptions; the full protocol in Appendix~\ref{app:decision-counts}, Figure~\ref{fig:collapse-decision-tree-full}, defines the exact classifications. In particular, not every failed or unresolved textual audit qualifies for limited support merely because a plateau is long.
With a large gap, concentration but a shorter plateau gives limited support. Without structural concentration, a long plateau also gives limited support if no covering principle model exists, or if the eligible audit satisfies $\neg T_1\wedge T_2\wedge T_3$. Other cases are classified as collapse not established. Missing audit evidence never passes a textual gate. Limited support based only on performance does not establish principle concentration; not established is not proof of continuing exploration. Full branch counts and audit qualifications appear in Appendix~\ref{app:decision-counts}.

\begin{figure}[t]
\centering
\includegraphics[width=\linewidth]{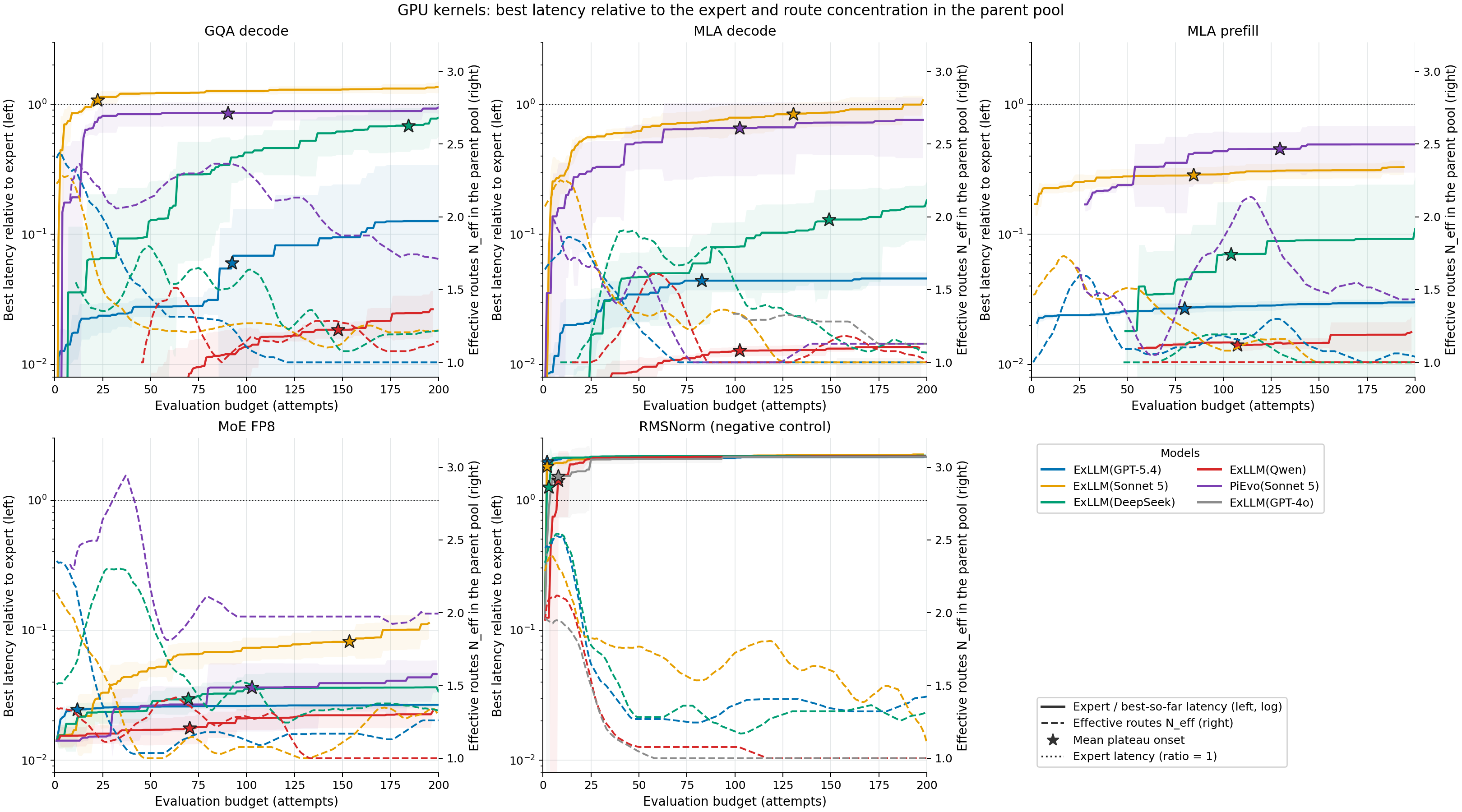}
\caption{Kernel performance and route concentration. Solid lines show expert latency divided by best-so-far latency (left, log scale; higher is better); dashed lines show effective route count in the top 10 (right). Stars mark mean plateau onset. RMSNorm is a negative control.}
\label{fig:kernel-budget}
\end{figure}
\section{Decision-Tree Counts and Text Audit}
\label{app:decision-counts}
\begin{figure}[htbp]
\centering
\includegraphics[width=\linewidth]{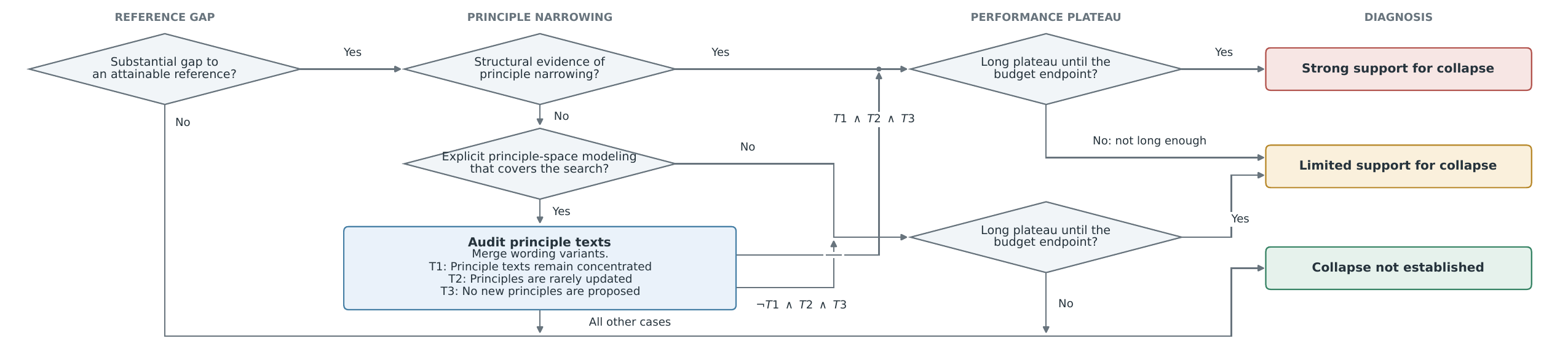}
\caption{Full diagnostic protocol. Structural evidence, eligibility for textual auditing, and the text audit are all parts of assessing principle narrowing. Text auditing requires explicit principle-space modeling covering the actual search directions. In the final 1,000 generated attempts, $T_1$ requires dominant strategy share $\ge50\%$, $T_2$ allows at most five minor revisions without a major change, and $T_3$ requires no new proposed direction.}
\label{fig:collapse-decision-tree-full}
\end{figure}

\begin{table}[h]
\centering
\scriptsize
\caption{Molecules: current non-intervention diagnostic counts.}
\begin{tabular}{@{}llrrr@{}}
\toprule
Method & Condition & S & L & N \\
\midrule
PiEvo ($\sigma=0.9$) & Sigma 0.9 & 1 & 1 & 3 \\
PiEvo & Original & 2 & 3 & 0 \\
ExLLM(Sonnet 5) & Explore (7:2:1) & 5 & 0 & 0 \\
ExLLM(Sonnet 5) & Original & 3 & 2 & 0 \\
ExLLM(Sonnet 5) & Web search & 3 & 2 & 0 \\
ExLLM(Sonnet 5 Thinking) & Original & 5 & 0 & 0 \\
ExLLM(GPT-4o) & Original & 2 & 3 & 0 \\
ExLLM(DeepSeek) & Explore (7:2:1) & 4 & 0 & 1 \\
ExLLM(DeepSeek) & Original & 5 & 0 & 0 \\
ExLLM(DeepSeek) & Temperature 1.3 & 5 & 0 & 0 \\
ExLLM(DeepSeek) & Web search & 3 & 0 & 2 \\
ExLLM(Qwen) & Explore (7:2:1) & 1 & 2 & 2 \\
ExLLM(Qwen) & Original & 4 & 1 & 0 \\
ExLLM(Qwen) & Temperature 1.3 & 0 & 3 & 2 \\
ExLLM(Qwen) & Web search & 1 & 2 & 2 \\
ExLLM(GPT-5.4) & Explore (7:2:1) & 4 & 1 & 0 \\
ExLLM(GPT-5.4) & Original & 4 & 0 & 1 \\
ExLLM(GPT-5.4) & Temperature 1.3 & 3 & 0 & 2 \\
ExLLM(GPT-5.4) & Web search & 2 & 0 & 3 \\
MCCE & MCCE & 0 & 3 & 2 \\
ReEvo & Original & 3 & 1 & 1 \\
ExLLM(GPT-4o + Gemini 2.5 Flash) & Two LLMs & 4 & 0 & 1 \\
\bottomrule
\end{tabular}
\end{table}
\begin{table}[htbp]
\centering\small
\caption{Non-intervention diagnostic counts on both networks. S/L/N denote strong support, limited support, and collapse not established. All rows use the original optimizer setting.}
\label{tab:network-diagnostic-counts}
\begin{tabular}{@{}lrrr@{\hspace{1.5em}}rrr@{}}
\toprule
 & \multicolumn{3}{c}{Dolphin} & \multicolumn{3}{c}{Residence hall} \\
\cmidrule(lr){2-4}\cmidrule(lr){5-7}
Method & S & L & N & S & L & N \\
\midrule
PiEvo & 0 & 0 & 5 & 4 & 1 & 0 \\
ExLLM(Sonnet 5) & 0 & 0 & 5 & 3 & 0 & 2 \\
ExLLM(DeepSeek) & 0 & 0 & 5 & 5 & 0 & 0 \\
ExLLM(Qwen) & 0 & 0 & 5 & 0 & 4 & 1 \\
ExLLM(GPT-5.4) & 0 & 0 & 5 & 1 & 1 & 3 \\
\bottomrule
\end{tabular}
\end{table}

\paragraph{Why a run is not strong.}
Among the 40 main molecular runs, 13 receive limited support and four are not established. MCCE contributes 0/3/2 and ReEvo 3/1/1 to the combined S/L/N counts. Borderline cases include Sonnet seed 44 ($N_{\mathrm{eff}}=3.010$) and PiEvo seed 43 (29.5\% budget remaining). Such boundaries motivate sensitivity analysis rather than distinct mechanistic claims. On residence hall, five main runs fail the twofold-gap gate, one Qwen run lacks both structural concentration and a long plateau, and six runs receive limited support. All 25 main Dolphin runs fail the large-gap gate; this does not mean every run reaches the exhaustive optimum.
\paragraph{PiEvo audit.}
Final-1,000-attempt windows are reconstructed from IDS proposal batches and checkpoint generation counters, including retries. Strategy labels merge wording variants using rule-assisted semantic coding with retained original texts and explicit review of revisions; this is not an independent double-annotator validation. Unknown labels are not automatically merged. In original molecular seed 45, five reviewed minor revisions satisfy $T_2$, but only 4.7\% of budget remains at plateau onset, giving limited support. Original molecular counts are 2/3/0; the $\sigma=0.9$ sensitivity run gives 1/1/3.
For residence-hall PiEvo, seeds 42 and 43 pass the structural route. Seeds 44 and 46 pass the textual route with dominant shares 59.6\% and 62.4\%, no late revisions or new directions, and long plateaus. Seed 45 has dominant share 40.0\%, no revisions or new directions, and a long plateau: limited support, not established concentration. These statistics use actual generated attempts rather than distinct principle strings.

\section{Detailed Performance and Concentration}
\label{app:tables}
\label{app:collapse-details}
The tables report final-checkpoint performance and selected-50 concentration statistics. Each ordinary method row summarizes five seeds. Molecular performance and concentration are separated for readability. ReEvo and MCCE are presently molecular comparisons only; missing network or kernel coverage is not assumed.
\begin{table}[h]
\centering
\caption{Molecular baseline performance. HV and diversity use the reconstructed 50-candidate set. Remaining budget is planned budget after the retrospective plateau onset, not measured wasted evaluations.}
\label{tab:mol}
\label{tab:mol-full}
\label{tab:overall}
\scriptsize
\resizebox{\linewidth}{!}{%
\begin{tabular}{@{}lccccccc@{}}
\toprule
Method / condition & Top-1 & Top-10 & HV (50) & $U$ & $D$ & $b_{\mathrm{plat}}$ & Remaining (\%) \\
\midrule
ExLLM(GPT-5.4) & $4.047\pm0.118$ & $4.027\pm0.110$ & $0.551\pm0.199$ & $0.326\pm0.067$ & $0.507\pm0.040$ & $2762.800\pm829.500$ & $44.744\pm16.590$ \\
ExLLM(Sonnet 5) & $3.975\pm0.008$ & $3.962\pm0.012$ & $0.374\pm0.022$ & $0.122\pm0.011$ & $0.533\pm0.050$ & $1172.400\pm459.202$ & $76.552\pm9.184$ \\
ExLLM(DeepSeek) & $3.998\pm0.049$ & $3.986\pm0.043$ & $0.419\pm0.093$ & $0.312\pm0.072$ & $0.463\pm0.032$ & $2064.000\pm802.579$ & $58.720\pm16.052$ \\
ExLLM(Qwen) & $4.091\pm0.109$ & $4.047\pm0.077$ & $0.659\pm0.220$ & $0.338\pm0.094$ & $0.589\pm0.025$ & $3079.200\pm1415.091$ & $38.416\pm28.302$ \\
PiEvo & $4.100\pm0.200$ & $4.057\pm0.189$ & $0.610\pm0.363$ & $0.559\pm0.303$ & $0.532\pm0.170$ & $3214.800\pm1509.945$ & $35.704\pm30.199$ \\
MCCE & $4.288\pm0.103$ & $4.220\pm0.059$ & $0.860\pm0.080$ & $0.656\pm0.020$ & $0.501\pm0.030$ & $4255.600\pm534.429$ & $14.888\pm10.689$ \\
ReEvo & $4.162\pm0.253$ & $4.116\pm0.201$ & $0.684\pm0.301$ & $0.533\pm0.039$ & $0.559\pm0.036$ & $2146.200\pm1502.429$ & $57.076\pm30.049$ \\
\bottomrule
\end{tabular}}
\end{table}
\begin{table}[h]
\centering
\caption{Molecular concentration. Early/late denote first-offspring/final reconstructed sets; uniqueness instead uses first/last 1,000 generated candidates. These are top-50 measurements, not the earlier top-10 window-based verdicts.}
\label{tab:scaf-full}
\label{tab:recent-top10}
\scriptsize
\resizebox{\linewidth}{!}{%
\begin{tabular}{@{}lcccccccccc@{}}
\toprule
Method / condition & $U_e$ & $U_l$ & $D_e$ & $D_l$ & $N_{\mathrm{eff},e}$ & $N_{\mathrm{eff},l}$ & $S_e$ & $S_l$ & $m_e$ & $m_l$ \\
\midrule
ExLLM(GPT-5.4) & $0.717\pm0.058$ & $0.206\pm0.082$ & $0.849\pm0.008$ & $0.507\pm0.040$ & $40.117\pm4.846$ & $1.908\pm1.116$ & $42.800\pm3.564$ & $3.400\pm1.817$ & $0.068\pm0.018$ & $0.820\pm0.215$ \\
ExLLM(Sonnet 5) & $0.495\pm0.067$ & $0.067\pm0.013$ & $0.842\pm0.006$ & $0.533\pm0.050$ & $38.842\pm3.583$ & $2.577\pm1.312$ & $41.800\pm2.775$ & $3.400\pm2.302$ & $0.064\pm0.009$ & $0.664\pm0.204$ \\
ExLLM(DeepSeek) & $0.886\pm0.032$ & $0.098\pm0.058$ & $0.862\pm0.007$ & $0.463\pm0.032$ & $43.106\pm6.482$ & $1.241\pm0.332$ & $46.000\pm3.391$ & $1.600\pm0.894$ & $0.084\pm0.054$ & $0.936\pm0.092$ \\
ExLLM(Qwen) & $0.675\pm0.047$ & $0.390\pm0.103$ & $0.845\pm0.008$ & $0.589\pm0.025$ & $42.094\pm4.131$ & $1.755\pm0.730$ & $45.000\pm1.871$ & $3.200\pm1.924$ & $0.076\pm0.050$ & $0.808\pm0.209$ \\
PiEvo & $0.824\pm0.129$ & $0.415\pm0.363$ & $0.863\pm0.005$ & $0.532\pm0.170$ & $48.276\pm1.257$ & $30.685\pm11.115$ & $48.800\pm0.837$ & $35.400\pm9.813$ & $0.040\pm0.014$ & $0.132\pm0.067$ \\
MCCE & $0.770\pm0.047$ & $0.646\pm0.022$ & $0.852\pm0.004$ & $0.501\pm0.030$ & $39.926\pm4.504$ & $1.659\pm0.547$ & $43.400\pm2.966$ & $2.400\pm1.140$ & $0.088\pm0.036$ & $0.804\pm0.177$ \\
ReEvo & $0.631\pm0.035$ & $0.491\pm0.028$ & $0.834\pm0.014$ & $0.559\pm0.036$ & $36.663\pm2.549$ & $2.742\pm1.885$ & $41.200\pm1.643$ & $6.400\pm2.074$ & $0.096\pm0.017$ & $0.736\pm0.246$ \\
\bottomrule
\end{tabular}}
\end{table}
\begin{table}[h]
\centering\scriptsize
\caption{Network baseline performance and effective node pool. Effective nodes summarize node usage, not semantic principles; lower errors are better.}
\label{tab:network-performance}
\label{tab:dolphin-updated}
\label{tab:residence_hall-updated}
\label{tab:dyn}
\label{tab:dyn-full}
\resizebox{\linewidth}{!}{%
\begin{tabular}{@{}lccccccc@{}}
\toprule
Method / condition & Top-1 error & Top-10 error & $U$ & $N_{\mathrm{eff},e}$ & $N_{\mathrm{eff},l}$ & $b_{\mathrm{plat}}$ & Remaining (\%) \\
\midrule
\multicolumn{8}{l}{\textbf{Dolphin}} \\
ExLLM(GPT-5.4) & $4.071\times10^{-3}\pm2.406\times10^{-5}$ & $4.147\times10^{-3}\pm2.617\times10^{-5}$ & $0.228\pm0.028$ & $45.780\pm2.351$ & $13.971\pm1.419$ & $852.400\pm276.991$ & $72.503\pm8.935$ \\
ExLLM(Sonnet 5) & $4.115\times10^{-3}\pm7.582\times10^{-5}$ & $4.189\times10^{-3}\pm8.557\times10^{-5}$ & $0.420\pm0.035$ & $44.787\pm4.002$ & $13.990\pm0.630$ & $929.000\pm570.675$ & $70.032\pm18.409$ \\
ExLLM(DeepSeek) & $4.216\times10^{-3}\pm3.604\times10^{-5}$ & $4.320\times10^{-3}\pm3.408\times10^{-5}$ & $0.271\pm0.061$ & $48.251\pm2.670$ & $14.214\pm1.602$ & $657.400\pm309.967$ & $78.794\pm9.999$ \\
ExLLM(Qwen) & $4.234\times10^{-3}\pm1.073\times10^{-4}$ & $4.373\times10^{-3}\pm1.501\times10^{-4}$ & $0.482\pm0.115$ & $50.305\pm1.465$ & $21.705\pm7.702$ & $1291.200\pm334.970$ & $58.348\pm10.805$ \\
PiEvo & $4.098\times10^{-3}\pm7.122\times10^{-5}$ & $4.176\times10^{-3}\pm1.034\times10^{-4}$ & $0.099\pm0.052$ & $33.441\pm12.692$ & $14.108\pm1.878$ & $962.600\pm453.699$ & $68.948\pm14.635$ \\
\midrule
\multicolumn{8}{l}{\textbf{Residence hall}} \\
ExLLM(GPT-5.4) & $7.005\times10^{-7}\pm5.689\times10^{-7}$ & $1.352\times10^{-6}\pm6.547\times10^{-7}$ & $0.378\pm0.109$ & $101.948\pm6.277$ & $26.752\pm3.587$ & $4440.800\pm2574.126$ & $59.071\pm23.725$ \\
ExLLM(Sonnet 5) & $1.514\times10^{-6}\pm1.084\times10^{-6}$ & $4.434\times10^{-6}\pm2.171\times10^{-6}$ & $0.303\pm0.109$ & $72.039\pm9.282$ & $18.709\pm1.172$ & $3462.400\pm3369.235$ & $68.088\pm31.053$ \\
ExLLM(DeepSeek) & $1.402\times10^{-6}\pm8.028\times10^{-7}$ & $3.846\times10^{-6}\pm1.887\times10^{-6}$ & $0.255\pm0.049$ & $118.957\pm5.068$ & $23.065\pm5.774$ & $2454.600\pm1514.687$ & $77.377\pm13.960$ \\
ExLLM(Qwen) & $9.961\times10^{-7}\pm6.875\times10^{-7}$ & $4.993\times10^{-5}\pm6.598\times10^{-5}$ & $0.731\pm0.087$ & $128.440\pm3.310$ & $99.446\pm26.186$ & $5875.800\pm3029.916$ & $45.845\pm27.925$ \\
PiEvo & $1.299\times10^{-6}\pm5.886\times10^{-7}$ & $4.436\times10^{-6}\pm1.781\times10^{-6}$ & $0.315\pm0.064$ & $75.339\pm39.760$ & $36.953\pm19.704$ & $1906.400\pm986.742$ & $82.429\pm9.094$ \\
\bottomrule
\end{tabular}}
\end{table}

\begin{table}[h]
\centering\scriptsize
\caption{Network node concentration. $C_5$ is normalized top-five coverage; $R_2,R_3$ are maximum pair/triple reuse. Higher values indicate concentration; $e/l$ denote early/late.}
\label{tab:network-concentration}
\label{tab:dolphin-concentration}
\label{tab:residence_hall-concentration}
\resizebox{\linewidth}{!}{%
\begin{tabular}{@{}lcccccccc@{}}
\toprule
Method / condition & $U_e$ & $U_l$ & $C_{5,e}$ & $C_{5,l}$ & $R_{2,e}$ & $R_{2,l}$ & $R_{3,e}$ & $R_{3,l}$ \\
\midrule
\multicolumn{9}{l}{\textbf{Dolphin}} \\
ExLLM(GPT-5.4) & $0.482\pm0.021$ & $0.169\pm0.041$ & $0.270\pm0.021$ & $0.742\pm0.080$ & $0.088\pm0.011$ & $0.324\pm0.134$ & $0.052\pm0.011$ & $0.128\pm0.056$ \\
ExLLM(Sonnet 5) & $0.652\pm0.132$ & $0.282\pm0.047$ & $0.294\pm0.024$ & $0.745\pm0.041$ & $0.084\pm0.009$ & $0.372\pm0.070$ & $0.056\pm0.009$ & $0.156\pm0.038$ \\
ExLLM(DeepSeek) & $0.546\pm0.079$ & $0.154\pm0.027$ & $0.244\pm0.017$ & $0.703\pm0.114$ & $0.072\pm0.011$ & $0.360\pm0.235$ & $0.044\pm0.009$ & $0.108\pm0.048$ \\
ExLLM(Qwen) & $0.658\pm0.040$ & $0.389\pm0.172$ & $0.248\pm0.019$ & $0.617\pm0.178$ & $0.064\pm0.009$ & $0.360\pm0.258$ & $0.036\pm0.009$ & $0.132\pm0.097$ \\
PiEvo & $0.447\pm0.129$ & $0.039\pm0.029$ & $0.404\pm0.126$ & $0.693\pm0.050$ & $0.216\pm0.109$ & $0.328\pm0.076$ & $0.080\pm0.037$ & $0.124\pm0.022$ \\
\midrule
\multicolumn{9}{l}{\textbf{Residence hall}} \\
ExLLM(GPT-5.4) & $0.715\pm0.073$ & $0.255\pm0.129$ & $0.152\pm0.011$ & $0.443\pm0.069$ & $0.080\pm0.020$ & $0.236\pm0.067$ & $0.068\pm0.011$ & $0.104\pm0.009$ \\
ExLLM(Sonnet 5) & $0.749\pm0.030$ & $0.171\pm0.098$ & $0.285\pm0.033$ & $0.604\pm0.019$ & $0.148\pm0.033$ & $0.428\pm0.088$ & $0.072\pm0.018$ & $0.196\pm0.033$ \\
ExLLM(DeepSeek) & $0.665\pm0.043$ & $0.149\pm0.044$ & $0.114\pm0.013$ & $0.504\pm0.089$ & $0.072\pm0.011$ & $0.324\pm0.071$ & $0.064\pm0.009$ & $0.160\pm0.024$ \\
ExLLM(Qwen) & $0.903\pm0.065$ & $0.624\pm0.138$ & $0.110\pm0.010$ & $0.178\pm0.096$ & $0.044\pm0.009$ & $0.080\pm0.032$ & $0.032\pm0.011$ & $0.064\pm0.017$ \\
PiEvo & $0.599\pm0.084$ & $0.217\pm0.088$ & $0.270\pm0.200$ & $0.390\pm0.146$ & $0.208\pm0.210$ & $0.216\pm0.124$ & $0.140\pm0.097$ & $0.108\pm0.046$ \\
\bottomrule
\end{tabular}}
\end{table}

\begin{figure}[h]
\centering
\includegraphics[width=.495\linewidth]{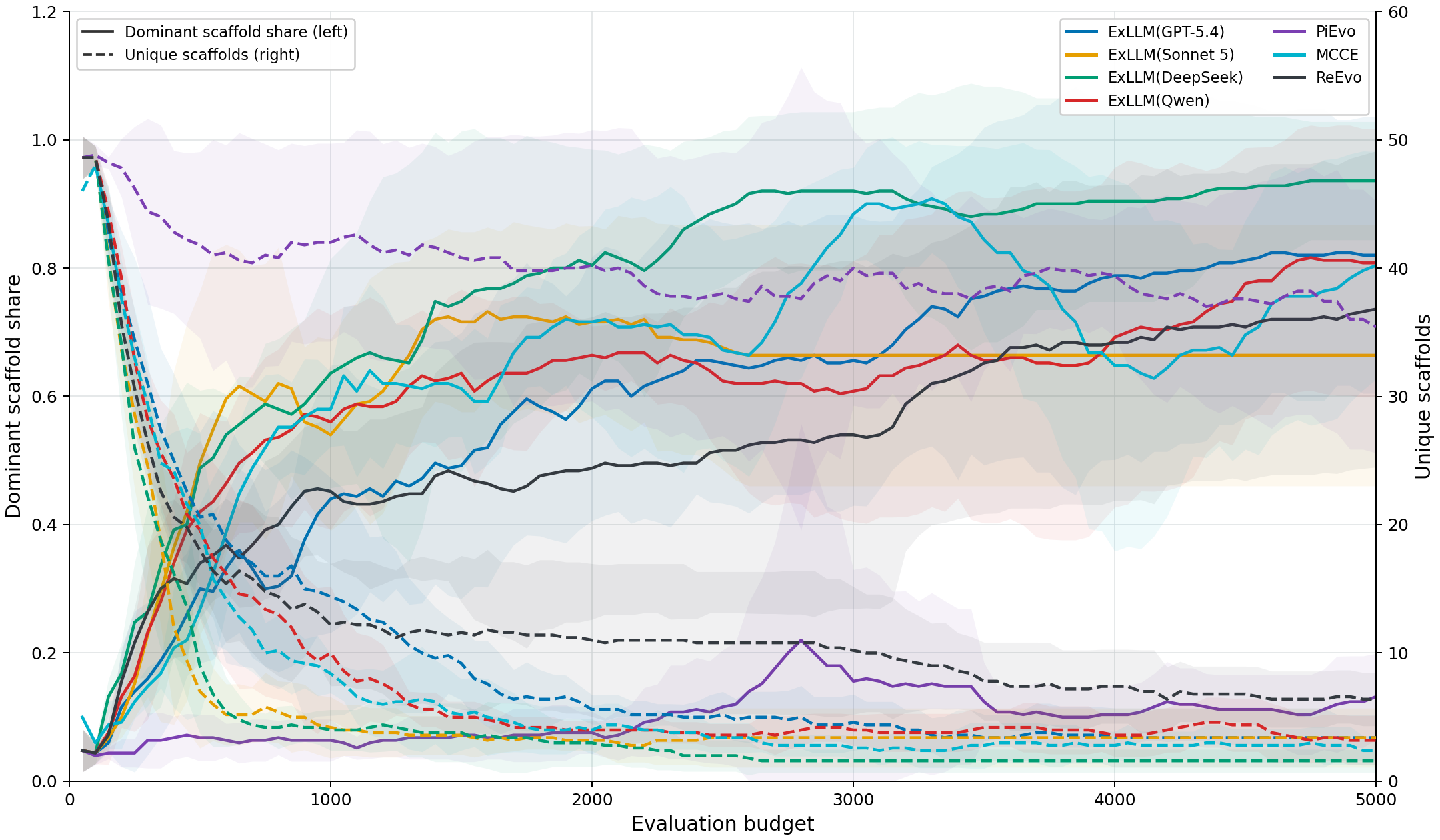}\hfill
\includegraphics[width=.495\linewidth]{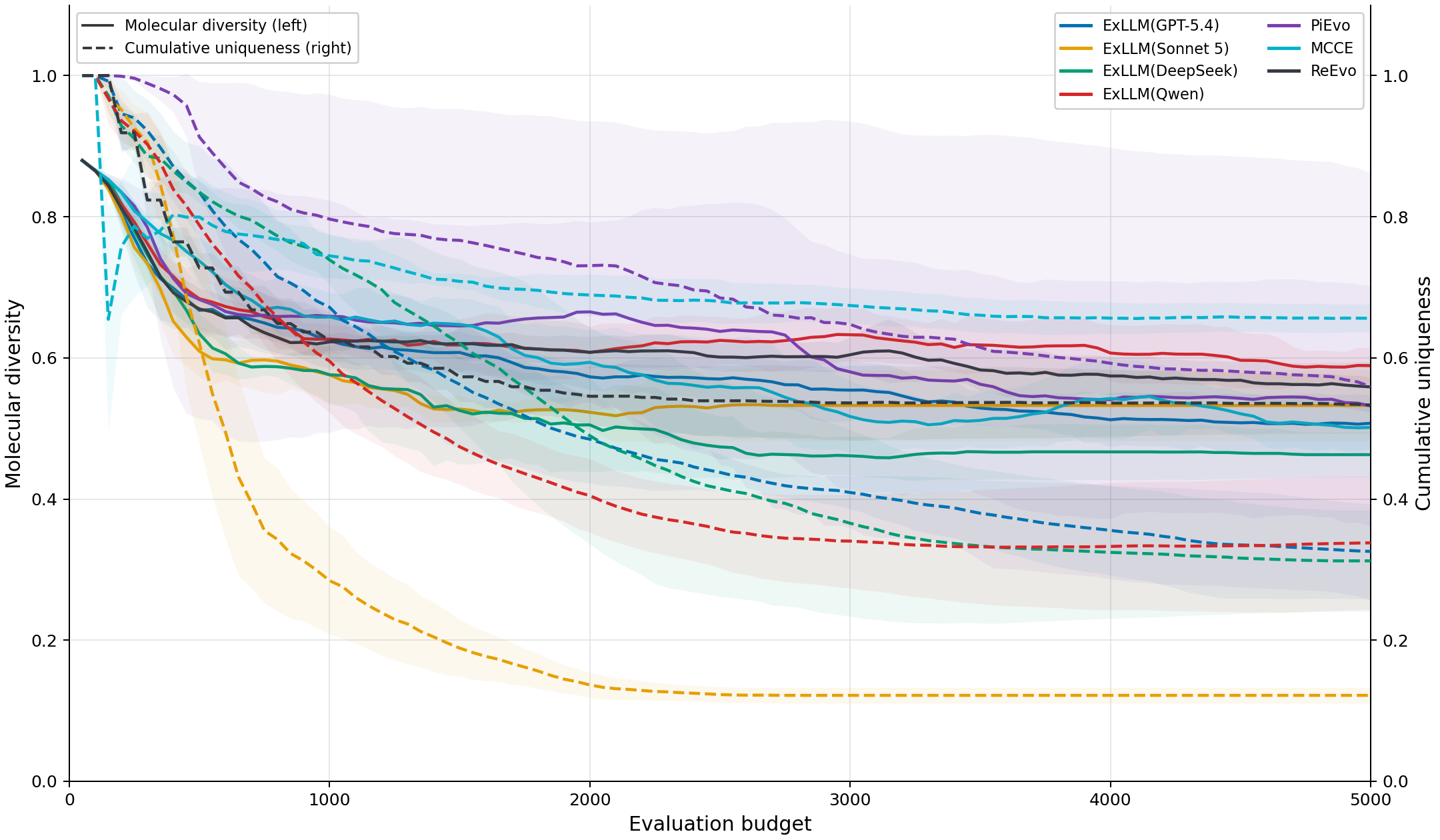}
\caption{Additional molecular trajectories: dominant scaffold share and unique scaffold count (left), fingerprint diversity and cumulative uniqueness (right). Final values are carried forward to budget 5,000.}
\label{fig:mol-extra}
\end{figure}

\begin{figure}[h]
\centering
\includegraphics[width=.495\linewidth]{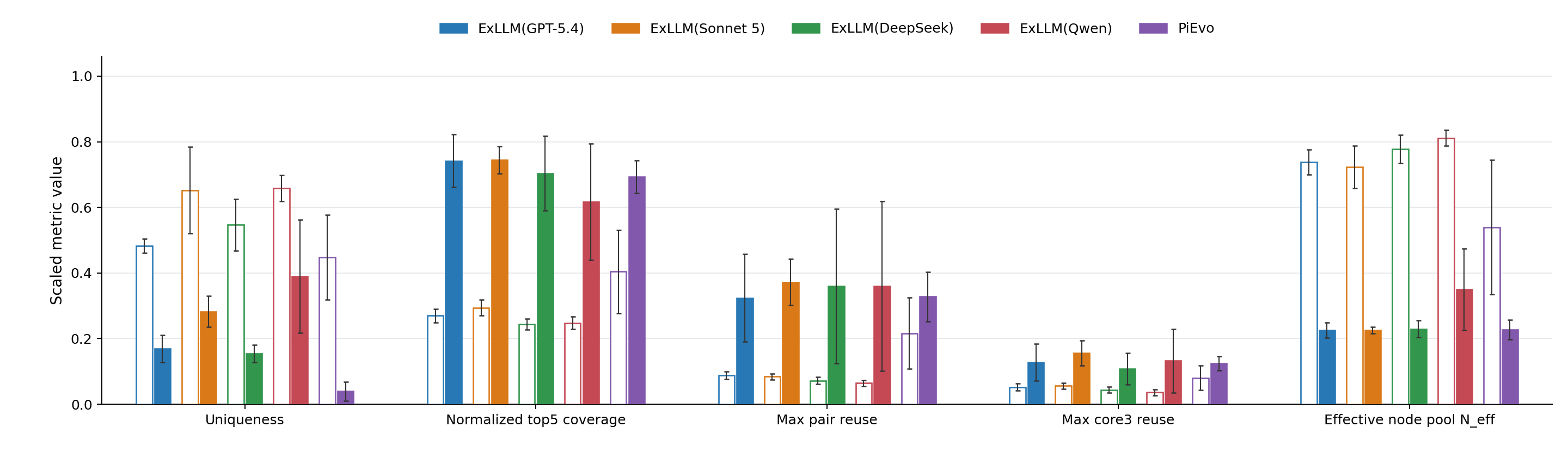}\hfill
\includegraphics[width=.495\linewidth]{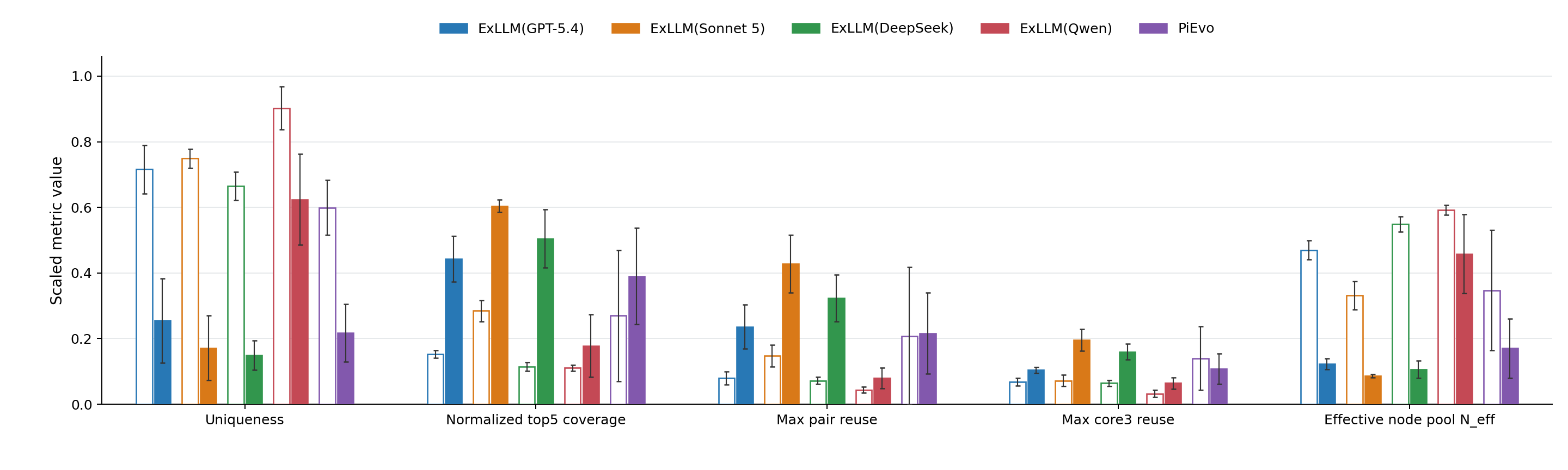}\\
\includegraphics[width=.495\linewidth]{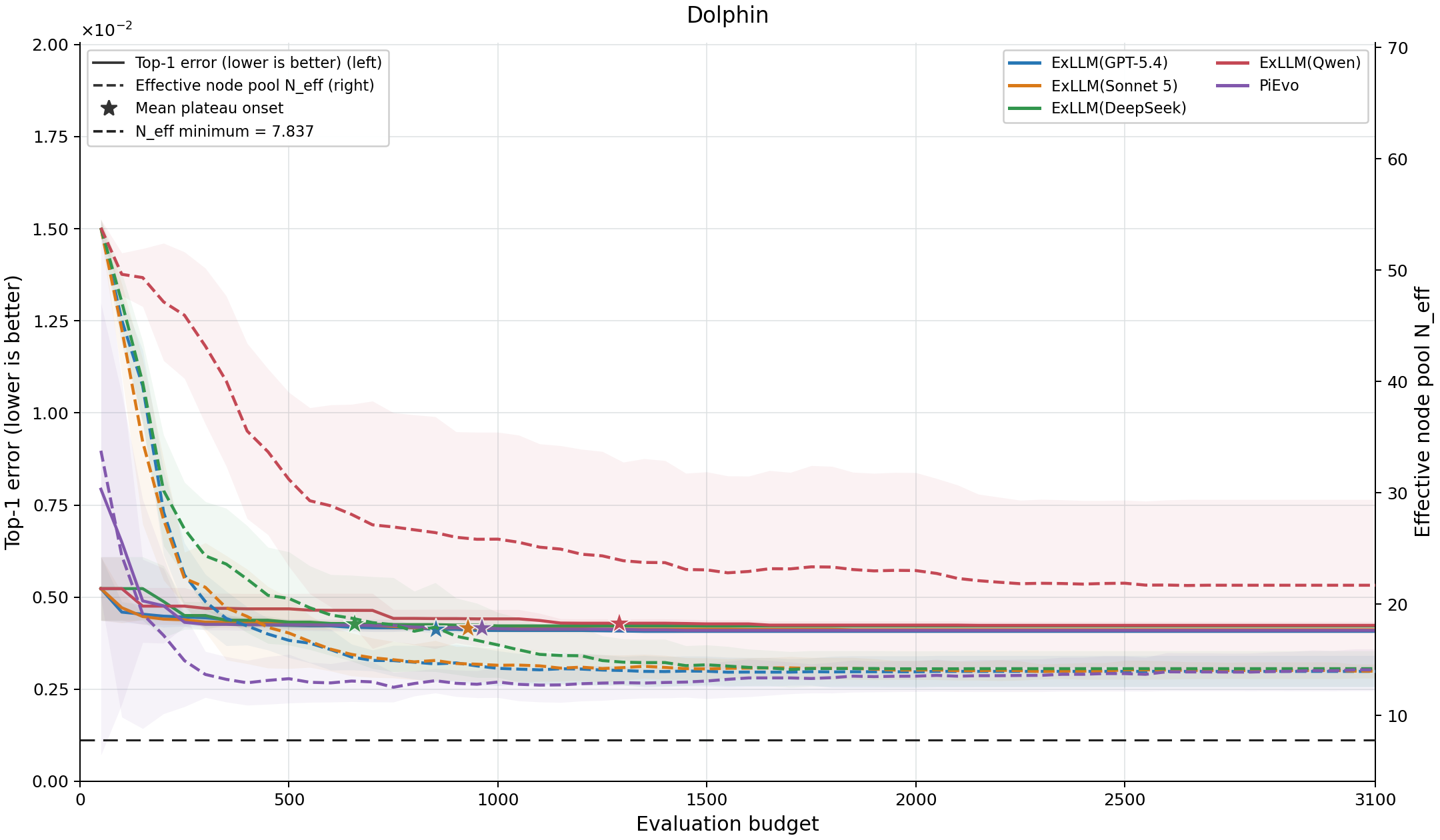}\hfill
\includegraphics[width=.495\linewidth]{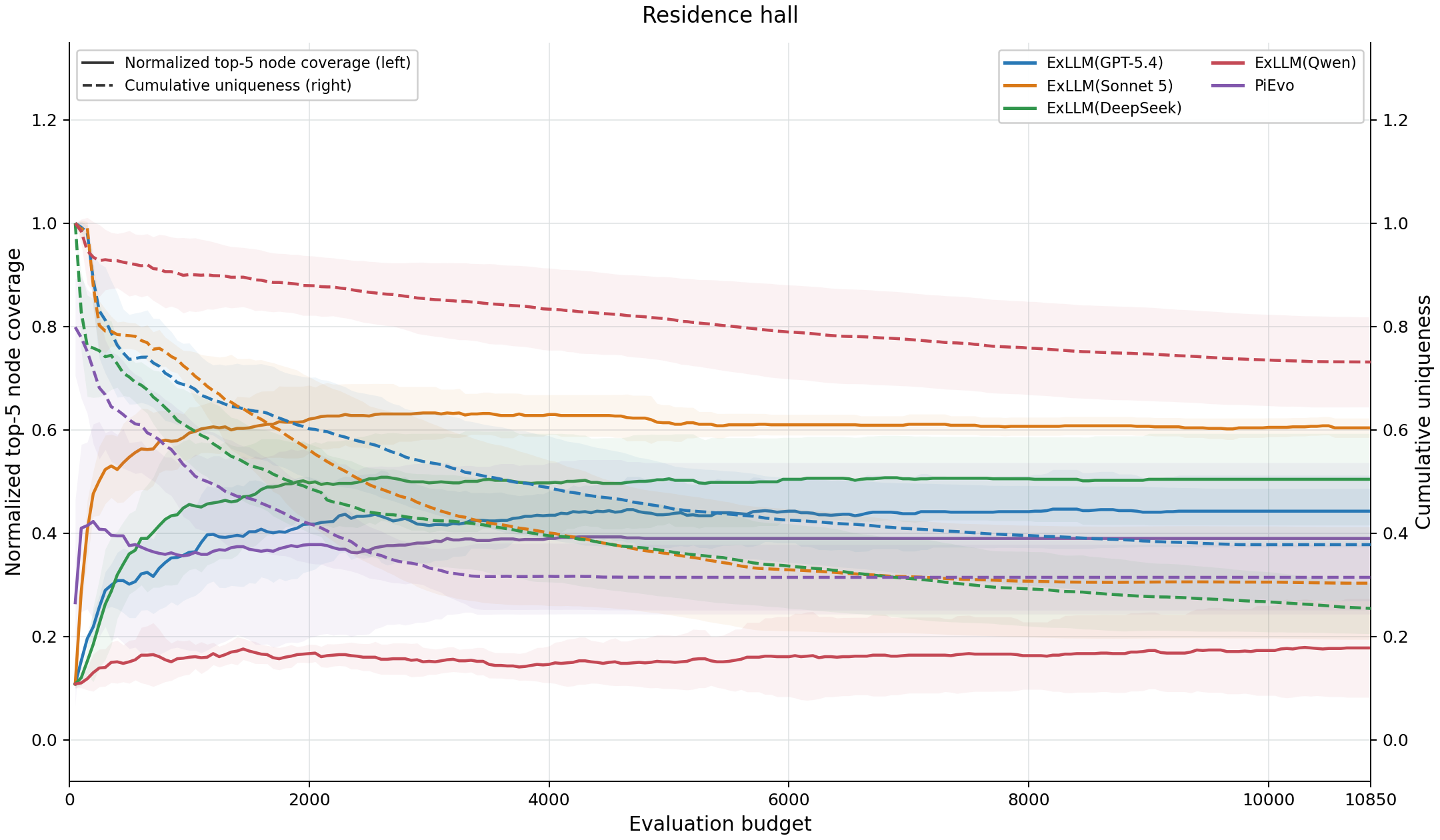}\\
\includegraphics[width=.495\linewidth]{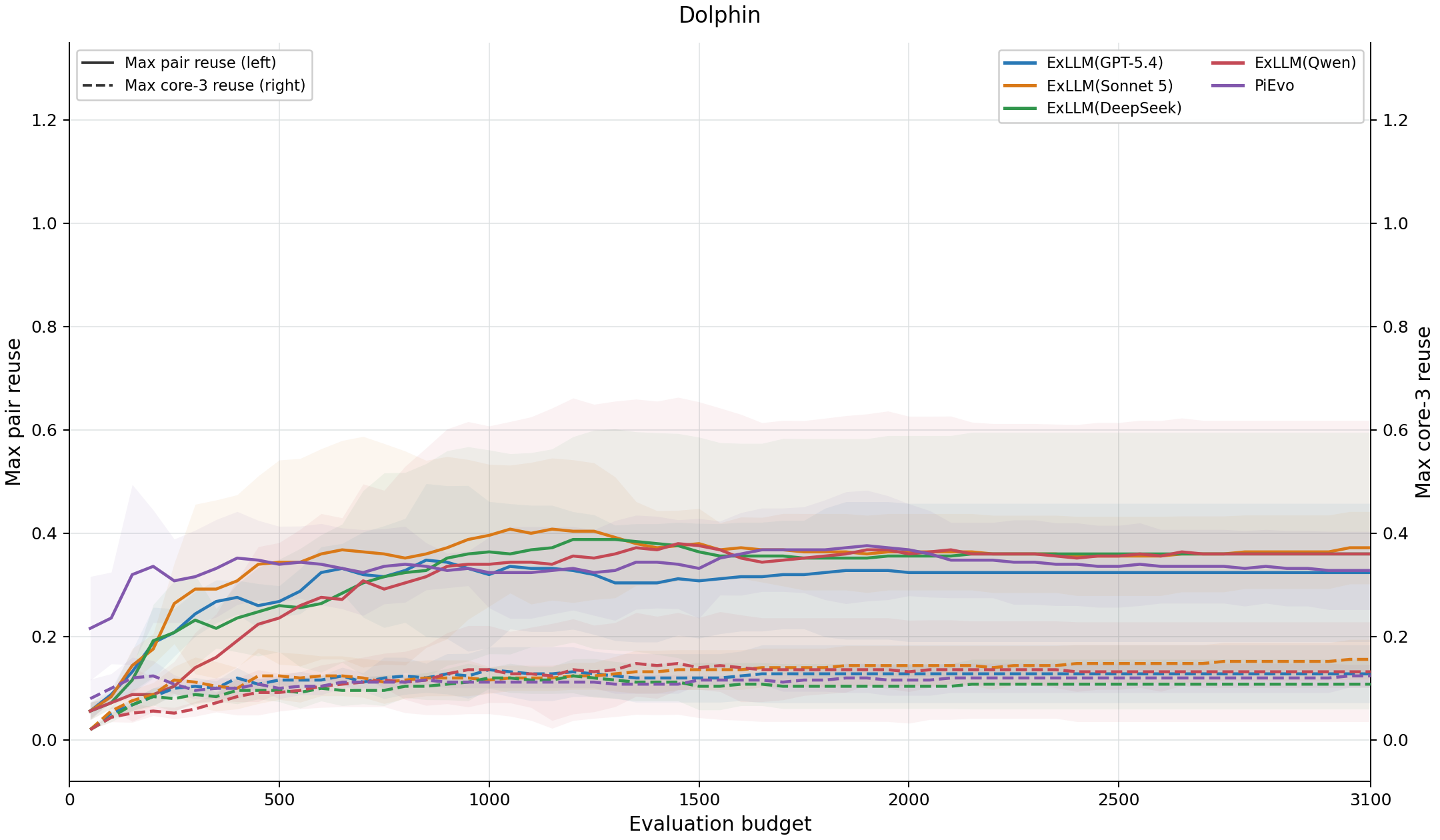}\hfill
\includegraphics[width=.495\linewidth]{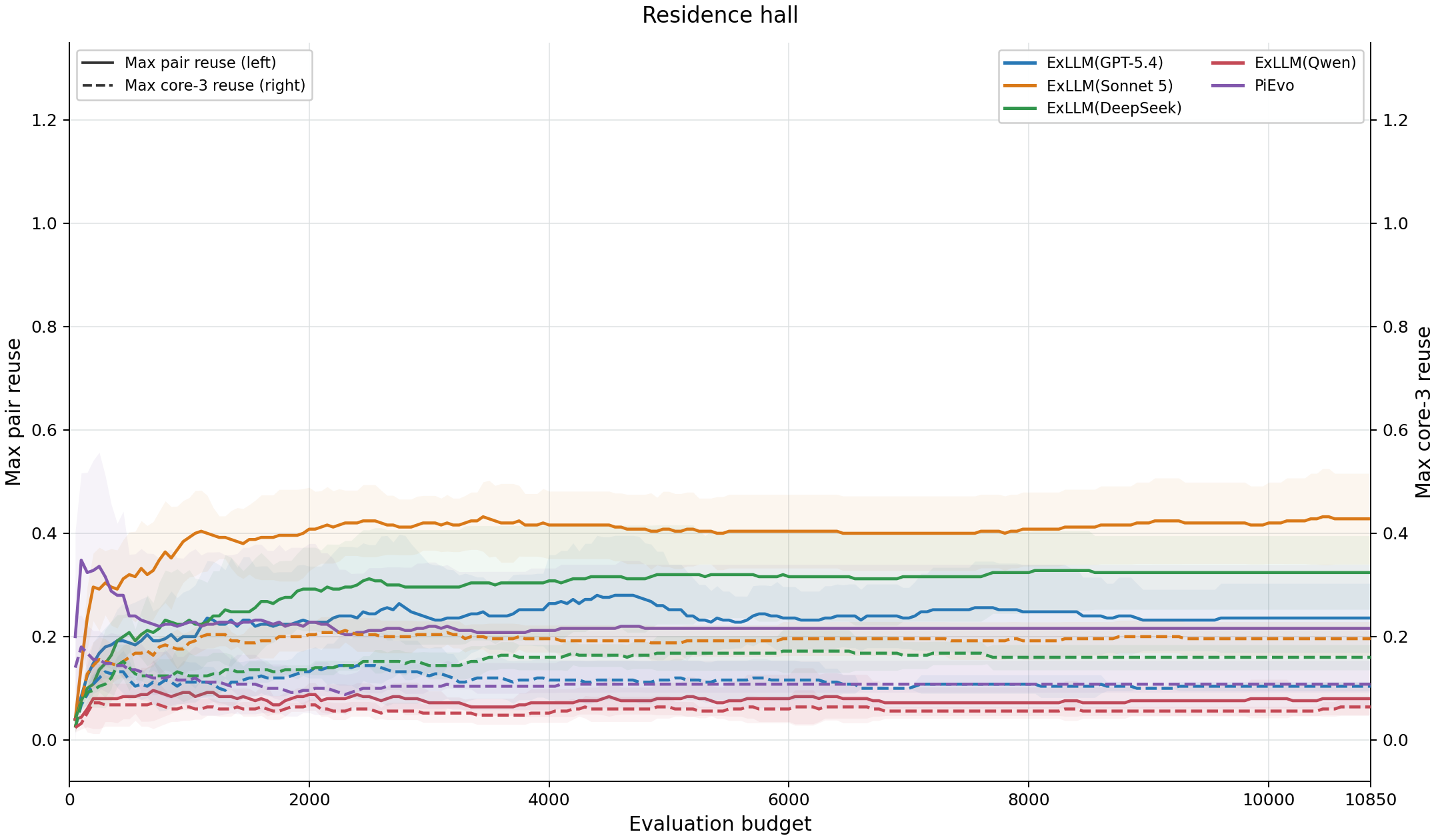}
\caption{Network early/late bars and supplementary trajectories. Dolphin's concentration is interpreted against its exhaustive optimum, not as failure on its own. Pair/triple reuse measures repeated node combinations; effective-node count measures spread of node usage.}
\label{fig:dyn-collapse}
\end{figure}

\begin{figure}[h]
\centering
\includegraphics[width=.495\linewidth]{results_v4/dolphin_top1_effective_nodes.png}\hfill
\includegraphics[width=.495\linewidth]{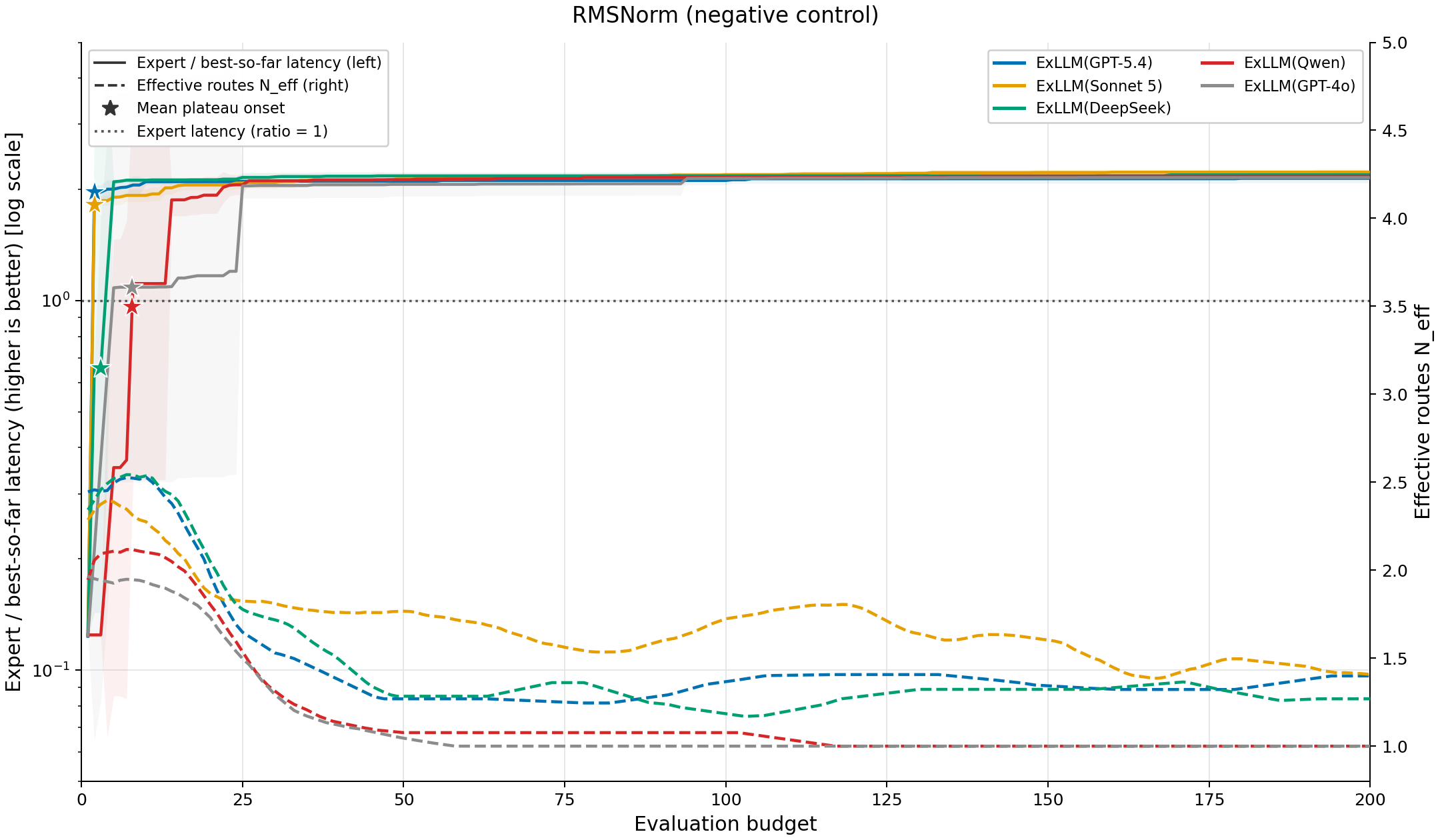}
\caption{Negative controls in the layout of Figure~\ref{fig:teaser}: Dolphin (left; the reference is the exhaustive optimum) and RMSNorm (right; the reference is the expert kernel, ratio $=1$). Solid/dashed lines use the left/right axes; stars mark mean plateau onset. Both tasks plateau early with concentrated pools, but near or beyond the reference, so the Headroom condition fails and collapse is not established. The RMSNorm curve is the geometric mean across seeds; $N_{\mathrm{eff}}$ uses the diagnostic set of the ten fastest kernels, smoothed as in Figure~\ref{fig:kernel-budget}; the GPT-4o control is included.}
\label{fig:negative-controls}
\end{figure}

\section{Diversity-Injection Details}
\label{app:diversity-details}
\begin{figure}[h]
\centering
\includegraphics[width=\linewidth]{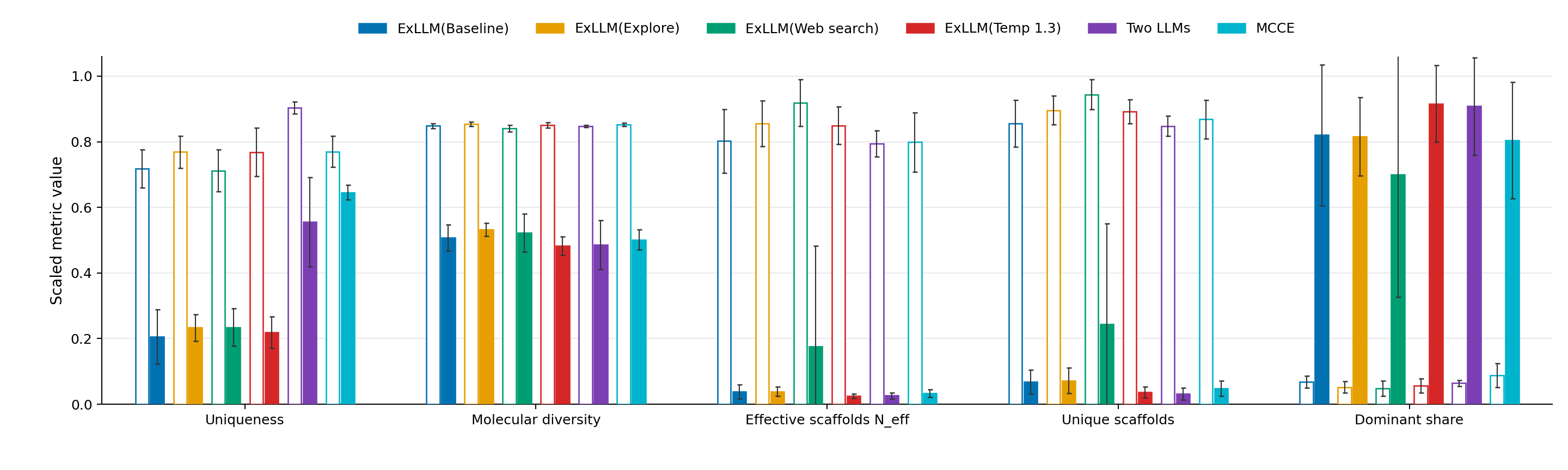}\\
\includegraphics[width=\linewidth]{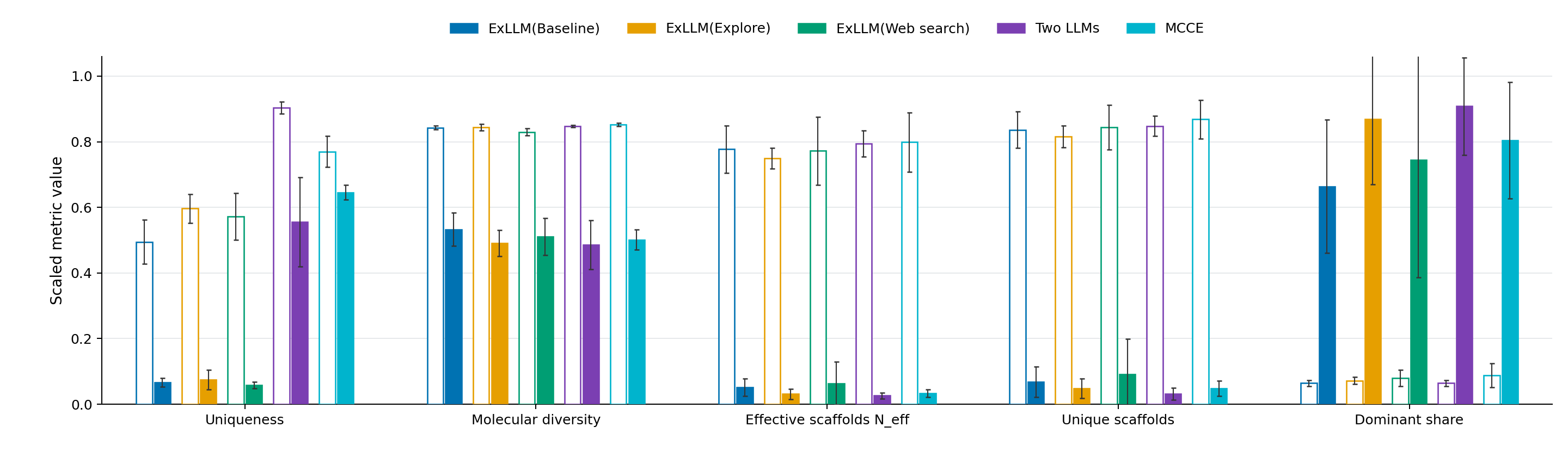}
\caption{Per-backbone diversity injection: GPT-5.4 and Sonnet. Sonnet has no temperature-1.3 arm. Two LLMs and MCCE are fixed independent comparators, not variants of the named backbone.}
\label{fig:ablation-backbones-a}
\end{figure}

\begin{figure}[h]
\centering
\includegraphics[width=\linewidth]{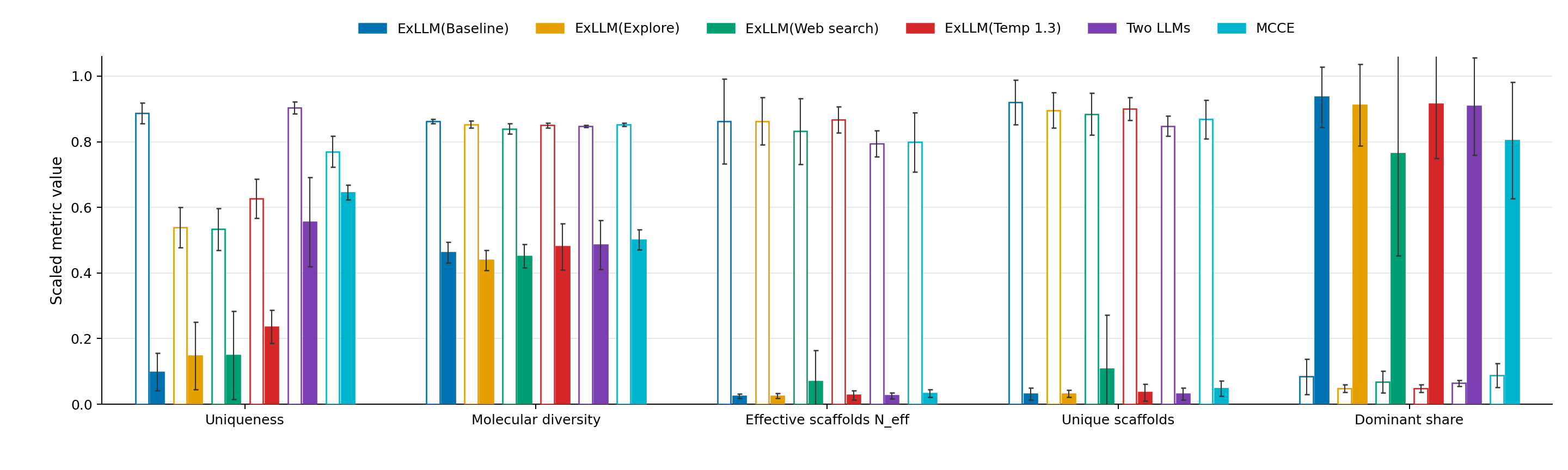}\\
\includegraphics[width=\linewidth]{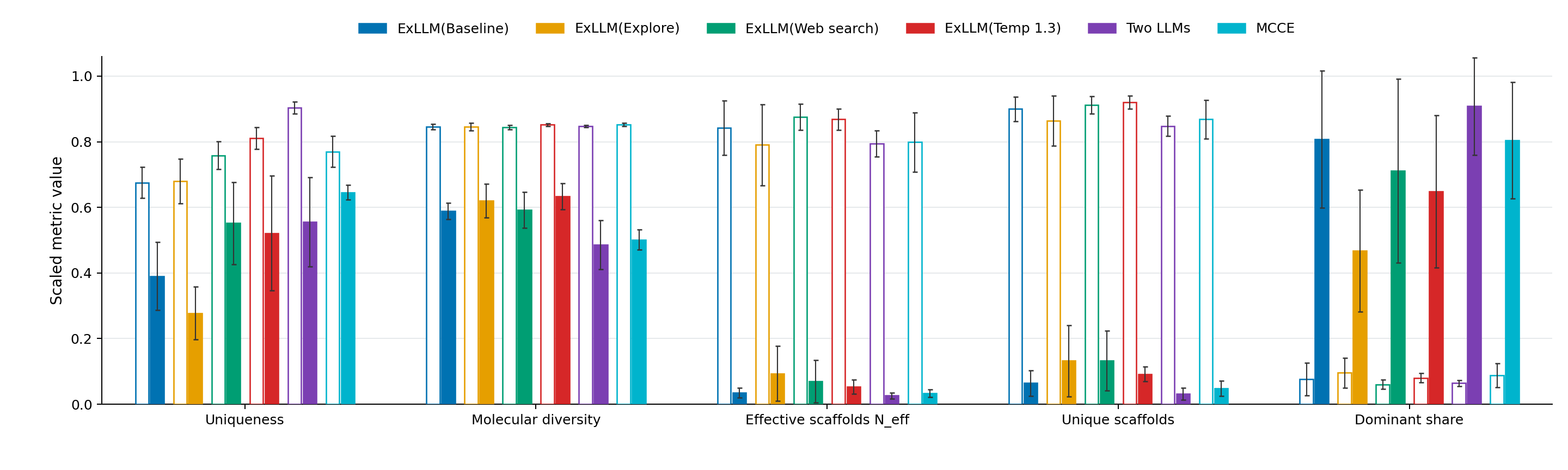}
\caption{Per-backbone diversity injection: DeepSeek and Qwen. These panels expose heterogeneity that a pooled mean can conceal; they do not establish a uniform effect size across backbones.}
\label{fig:ablation-backbones-b}
\end{figure}

Baseline, explore and web search pool four backbones; temperature pools three. This differs from the dispersion of 15 or 20 independent runs. Two LLMs uses the updated five-seed GPT-4o plus Gemini batch, not the earlier two-seed batch. MCCE is retained in both main and diversity-injection comparisons, using the same five runs rather than independent evidence. Full table is in table \ref{tab:diversity-full}, table \ref{tab:injection-concentration}, table \ref{tab:injection-backbones}. Per-backbone figures are in figure \ref{fig:ablation-backbones-a} and figure \ref{fig:ablation-backbones-b}.
\begin{table}[h]
\centering
\caption{Diversity injection, average backbones within each seed, then report mean of five seed averages. Temperature uses three backbones; baseline/explore/web use four. Two LLMs and MCCE each use five independent runs.}
\label{tab:diversity-full}
\scriptsize
\resizebox{\linewidth}{!}{%
\begin{tabular}{@{}lccccccc@{}}
\toprule
Method / condition & Top-1 & Top-10 & HV (50) & $U$ & $D$ & $b_{\mathrm{plat}}$ & Remaining (\%) \\
\midrule
ExLLM(Baseline) & $4.028\pm0.011$ & $4.005\pm0.007$ & $0.501\pm0.039$ & $0.274\pm0.044$ & $0.523\pm0.021$ & $2269.600\pm439.577$ & $54.608\pm8.792$ \\
ExLLM(Explore) & $4.048\pm0.079$ & $4.021\pm0.070$ & $0.503\pm0.115$ & $0.263\pm0.025$ & $0.520\pm0.007$ & $2262.700\pm554.565$ & $54.746\pm11.091$ \\
ExLLM(Web search) & $4.108\pm0.121$ & $4.067\pm0.091$ & $0.627\pm0.174$ & $0.301\pm0.005$ & $0.519\pm0.031$ & $2591.850\pm594.545$ & $48.163\pm11.891$ \\
ExLLM(Temp 1.3) & $4.086\pm0.116$ & $4.046\pm0.087$ & $0.577\pm0.148$ & $0.400\pm0.031$ & $0.532\pm0.029$ & $2425.200\pm1042.968$ & $51.496\pm20.859$ \\
Two LLMs & $4.165\pm0.304$ & $4.141\pm0.274$ & $0.667\pm0.356$ & $0.636\pm0.106$ & $0.485\pm0.075$ & $2032.200\pm949.502$ & $59.356\pm18.990$ \\
MCCE & $4.288\pm0.103$ & $4.220\pm0.059$ & $0.860\pm0.080$ & $0.656\pm0.020$ & $0.501\pm0.030$ & $4255.600\pm534.429$ & $14.888\pm10.689$ \\
\bottomrule
\end{tabular}}
\end{table}
\begin{table}[!h]
\centering
\caption{Diversity-injection concentration under the same aggregation as Table~\ref{tab:diversity-full}.}
\label{tab:injection-concentration}
\scriptsize
\resizebox{\linewidth}{!}{%
\begin{tabular}{@{}lcccccccccc@{}}
\toprule
Method / condition & $U_e$ & $U_l$ & $D_e$ & $D_l$ & $N_{\mathrm{eff},e}$ & $N_{\mathrm{eff},l}$ & $S_e$ & $S_l$ & $m_e$ & $m_l$ \\
\midrule
ExLLM(Baseline) & $0.693\pm0.018$ & $0.190\pm0.033$ & $0.850\pm0.006$ & $0.523\pm0.021$ & $41.040\pm4.320$ & $1.871\pm0.230$ & $43.900\pm2.626$ & $2.900\pm0.720$ & $0.073\pm0.026$ & $0.807\pm0.038$ \\
ExLLM(Explore) & $0.646\pm0.030$ & $0.183\pm0.007$ & $0.849\pm0.008$ & $0.520\pm0.007$ & $40.740\pm1.241$ & $2.366\pm1.209$ & $43.400\pm0.894$ & $3.550\pm1.866$ & $0.067\pm0.009$ & $0.766\pm0.096$ \\
ExLLM(Web search) & $0.644\pm0.032$ & $0.248\pm0.036$ & $0.839\pm0.008$ & $0.519\pm0.031$ & $42.475\pm2.613$ & $4.727\pm3.496$ & $44.800\pm1.681$ & $7.200\pm3.475$ & $0.064\pm0.016$ & $0.730\pm0.116$ \\
ExLLM(Temp 1.3) & $0.735\pm0.027$ & $0.326\pm0.055$ & $0.851\pm0.006$ & $0.532\pm0.029$ & $43.091\pm1.725$ & $1.774\pm0.442$ & $45.200\pm1.121$ & $2.733\pm0.596$ & $0.061\pm0.014$ & $0.827\pm0.096$ \\
Two LLMs & $0.903\pm0.019$ & $0.556\pm0.136$ & $0.848\pm0.003$ & $0.485\pm0.075$ & $39.731\pm1.984$ & $1.298\pm0.461$ & $42.400\pm1.517$ & $1.600\pm0.894$ & $0.064\pm0.009$ & $0.908\pm0.148$ \\
MCCE & $0.770\pm0.047$ & $0.646\pm0.022$ & $0.852\pm0.004$ & $0.501\pm0.030$ & $39.926\pm4.504$ & $1.659\pm0.547$ & $43.400\pm2.966$ & $2.400\pm1.140$ & $0.088\pm0.036$ & $0.804\pm0.177$ \\
\bottomrule
\end{tabular}}
\end{table}
\begin{table}[h]
\centering\scriptsize
\caption{Per-backbone molecular diversity-injection comparisons. Sonnet has no temperature-1.3 arm.}
\label{tab:injection-backbones}
\label{tab:ablation-0}
\label{tab:ablation-1}
\label{tab:ablation-2}
\label{tab:ablation-3}
\resizebox{\linewidth}{!}{%
\begin{tabular}{@{}lccccccc@{}}
\toprule
Method / condition & Top-1 & Top-10 & HV (50) & $U$ & $D$ & $b_{\mathrm{plat}}$ & Remaining (\%) \\
\midrule
\multicolumn{8}{l}{\textbf{ExLLM(GPT-5.4)}} \\
ExLLM(GPT-5.4) / Explore (7:2:1) & $3.993\pm0.117$ & $3.978\pm0.106$ & $0.429\pm0.216$ & $0.367\pm0.038$ & $0.532\pm0.020$ & $2094.600\pm1632.767$ & $58.108\pm32.655$ \\
ExLLM(GPT-5.4) & $4.047\pm0.118$ & $4.027\pm0.110$ & $0.551\pm0.199$ & $0.326\pm0.067$ & $0.507\pm0.040$ & $2762.800\pm829.500$ & $44.744\pm16.590$ \\
ExLLM(GPT-5.4) / Temperature 1.3 & $4.204\pm0.366$ & $4.140\pm0.295$ & $0.656\pm0.476$ & $0.342\pm0.048$ & $0.482\pm0.028$ & $2880.600\pm1771.310$ & $42.388\pm35.426$ \\
ExLLM(GPT-5.4) / Web search & $4.062\pm0.271$ & $4.018\pm0.210$ & $0.559\pm0.356$ & $0.310\pm0.061$ & $0.523\pm0.058$ & $2781.000\pm1498.193$ & $44.380\pm29.964$ \\
\midrule
\multicolumn{8}{l}{\textbf{ExLLM(Sonnet 5)}} \\
ExLLM(Sonnet 5) / Explore (7:2:1) & $4.028\pm0.062$ & $4.009\pm0.067$ & $0.468\pm0.109$ & $0.176\pm0.020$ & $0.491\pm0.040$ & $1485.800\pm345.575$ & $70.284\pm6.912$ \\
ExLLM(Sonnet 5) & $3.975\pm0.008$ & $3.962\pm0.012$ & $0.374\pm0.022$ & $0.122\pm0.011$ & $0.533\pm0.050$ & $1172.400\pm459.202$ & $76.552\pm9.184$ \\
ExLLM(Sonnet 5) / Web search & $4.048\pm0.101$ & $4.027\pm0.100$ & $0.513\pm0.182$ & $0.139\pm0.028$ & $0.511\pm0.056$ & $1722.400\pm690.889$ & $65.552\pm13.818$ \\
\midrule
\multicolumn{8}{l}{\textbf{ExLLM(DeepSeek)}} \\
ExLLM(DeepSeek) / Explore (7:2:1) & $4.116\pm0.325$ & $4.082\pm0.270$ & $0.531\pm0.435$ & $0.177\pm0.013$ & $0.439\pm0.031$ & $2168.600\pm1623.028$ & $56.628\pm32.461$ \\
ExLLM(DeepSeek) & $3.998\pm0.049$ & $3.986\pm0.043$ & $0.419\pm0.093$ & $0.312\pm0.072$ & $0.463\pm0.032$ & $2064.000\pm802.579$ & $58.720\pm16.052$ \\
ExLLM(DeepSeek) / Temperature 1.3 & $4.004\pm0.075$ & $3.995\pm0.071$ & $0.412\pm0.127$ & $0.327\pm0.078$ & $0.480\pm0.070$ & $1062.600\pm442.082$ & $78.748\pm8.842$ \\
ExLLM(DeepSeek) / Web search & $4.192\pm0.329$ & $4.138\pm0.281$ & $0.660\pm0.466$ & $0.215\pm0.053$ & $0.452\pm0.035$ & $2171.600\pm1648.882$ & $56.568\pm32.978$ \\
\midrule
\multicolumn{8}{l}{\textbf{ExLLM(Qwen)}} \\
ExLLM(Qwen) / Explore (7:2:1) & $4.057\pm0.134$ & $4.017\pm0.102$ & $0.583\pm0.269$ & $0.333\pm0.049$ & $0.620\pm0.052$ & $3301.800\pm1852.730$ & $33.964\pm37.055$ \\
ExLLM(Qwen) & $4.091\pm0.109$ & $4.047\pm0.077$ & $0.659\pm0.220$ & $0.338\pm0.094$ & $0.589\pm0.025$ & $3079.200\pm1415.091$ & $38.416\pm28.302$ \\
ExLLM(Qwen) / Temperature 1.3 & $4.050\pm0.075$ & $4.002\pm0.066$ & $0.664\pm0.087$ & $0.532\pm0.084$ & $0.634\pm0.040$ & $3332.400\pm1777.876$ & $33.352\pm35.558$ \\
ExLLM(Qwen) / Web search & $4.132\pm0.064$ & $4.085\pm0.052$ & $0.777\pm0.082$ & $0.540\pm0.060$ & $0.592\pm0.054$ & $3692.400\pm755.305$ & $26.152\pm15.106$ \\
\bottomrule
\end{tabular}}
\end{table}


\section{Older-Model and Thinking Controls}
\label{app:backbone-controls}
\begin{table}[h]
\centering
\caption{Older-model and thinking controls. Model training differences are not experimentally isolated.}
\label{tab:backbone-controls}
\scriptsize
\resizebox{\linewidth}{!}{%
\begin{tabular}{@{}lccccccc@{}}
\toprule
Method / condition & Top-1 & Top-10 & $U$ & $U_e$ & $U_l$ & $N_{\mathrm{eff},l}$ & Top-10 scaffolds \\
\midrule
ExLLM(GPT-5.4) & $4.047\pm0.118$ & $4.027\pm0.110$ & $0.326\pm0.067$ & $0.717\pm0.058$ & $0.206\pm0.082$ & $1.908\pm1.116$ & $1.800\pm0.447$ \\
ExLLM(Sonnet 5) & $3.975\pm0.008$ & $3.962\pm0.012$ & $0.122\pm0.011$ & $0.495\pm0.067$ & $0.067\pm0.013$ & $2.577\pm1.312$ & $2.400\pm1.342$ \\
ExLLM(DeepSeek) & $3.998\pm0.049$ & $3.986\pm0.043$ & $0.312\pm0.072$ & $0.886\pm0.032$ & $0.098\pm0.058$ & $1.241\pm0.332$ & $1.200\pm0.447$ \\
ExLLM(Qwen) & $4.091\pm0.109$ & $4.047\pm0.077$ & $0.338\pm0.094$ & $0.675\pm0.047$ & $0.390\pm0.103$ & $1.755\pm0.730$ & $1.800\pm1.304$ \\
ExLLM(GPT-4o) & $4.190\pm0.007$ & $4.170\pm0.006$ & $0.833\pm0.061$ & $0.988\pm0.007$ & $0.682\pm0.075$ & $1.057\pm0.083$ & $1.000\pm0.000$ \\
ExLLM(Sonnet thinking) & $3.972\pm0.004$ & $3.963\pm0.011$ & $0.203\pm0.038$ & $0.575\pm0.062$ & $0.094\pm0.018$ & $1.665\pm0.770$ & $1.600\pm0.894$ \\
\bottomrule
\end{tabular}}
\end{table}
GPT-4o's global uniqueness is $0.833$, compared with $0.326$ for GPT-5.4 and $0.122$--$0.338$ across the four main ExLLM backbones. Yet all five GPT-4o runs end with a single scaffold in their top-10, and their mean top-50 effective count is $1.057$. Distinct molecule generation therefore does not imply exploration of distinct core designs.

Sonnet thinking raises global uniqueness from $0.122$ to $0.203$, but mean Top-1 changes from $3.975$ to $3.972$ and mean final effective count from $2.577$ to $1.665$. In this comparison, more generation-level variation does not translate into better optimization. One hypothesis is that differences in post-training make some newer models more conservative in repeated proposing. These experiments do not identify that cause: model scale, pretraining, decoding and provider behavior also differ, and post-training is not independently manipulated. We therefore report an older/newer-model contrast, not a demonstrated adverse effect of additional post-training.

\section{PiEvo: Text, Candidate and Update Evidence}
\label{app:pievo-details}
Our PiEvo is adapted from official code, but evaluates all proposals rather than only the top-ranked one because we have much more budget than experiments in original PiEvo. 

\begin{table}[h]
\centering
\caption{PiEvo on molecules. Early/late are the first/last 1,000 generated attempts after initialization; updates include warm-up. Text usage and posterior summaries are attempt-weighted; events are counted at their recorded batch boundaries.}
\label{tab:pievo-molecules}
\scriptsize
\resizebox{\linewidth}{!}{%
\begin{tabular}{@{}lccccccc@{}}
\toprule
Method / condition & Updates early & Updates late & Anomalies all & Anomalies early & Anomalies late & $T_{\mathrm{eff},e}$ & $T_{\mathrm{eff},l}$ \\
\midrule
PiEvo & $13.600\pm2.966$ & $1.000\pm2.236$ & $19.600\pm13.903$ & $3.600\pm2.966$ & $1.000\pm2.236$ & $2.675\pm0.951$ & $2.229\pm1.191$ \\
PiEvo / Sigma 0.9 & $11.000\pm0.707$ & $0.400\pm0.894$ & $3.800\pm2.168$ & $1.000\pm0.707$ & $0.400\pm0.894$ & $8.213\pm0.843$ & $9.053\pm1.411$ \\
\bottomrule
\end{tabular}}
\end{table}

\begin{table}[h]
\centering
\caption{PiEvo on residence hall. Early/late are the first/last 1,000 generated attempts after initialization; updates include warm-up. Text usage and posterior summaries are attempt-weighted; events are counted at their recorded batch boundaries.}
\label{tab:pievo-residence_hall}
\scriptsize
\resizebox{\linewidth}{!}{%
\begin{tabular}{@{}lccccccc@{}}
\toprule
Method / problem & Updates early & Updates late & Anomalies all & Anomalies early & Anomalies late & $T_{\mathrm{eff},e}$ & $T_{\mathrm{eff},l}$ \\
\midrule
PiEvo /dolphin & $11.600\pm1.342$ & $0.000\pm0.000$ & $4.200\pm2.950$ & $2.600\pm1.817$ & $0.000\pm0.000$ & $5.623\pm3.048$ & $2.826\pm1.487$ \\
PiEvo /residence hall & $11.800\pm1.304$ & $0.000\pm0.000$ & $6.400\pm2.702$ & $2.200\pm1.095$ & $0.000\pm0.000$ & $7.429\pm1.589$ & $6.382\pm2.348$ \\
\bottomrule
\end{tabular}}
\end{table}

\paragraph{Counting texts is not semantic annotation.}
Effective used-text count is $\exp(-\sum_j q_j\ln q_j)$, where $q_j$ is the fraction of generated attempts using exact normalized principle text $j$. Early/late windows are the first/last 1,000 generated attempts after initialization. Usage and posterior summaries are attempt-weighted, with partial boundary batches included proportionally. Updates are counted at batch start and anomaly events at batch end; event counts are not fractionally weighted. Different wording may describe the same strategy, and one strategy can produce multiple scaffolds. 

In original molecular seed 42, the dominant late instruction emphasizes 8--15-member macrocycles, with peripheral changes intended to tune properties. The post-plateau operational-route effective count is $1.149$, with share $0.975$ and no first use of a new route after the plateau. In the structural follow-through audit, 661 of 734 newly evaluated late molecules under the explicit macrocycle instruction contain an 8--15-member ring. Its high scaffold count therefore does not refute a narrower instruction-level search.

\paragraph{Performance qualifies the interpretation.}
That seed remains $0.800$ below the attainable molecular reference, with $74.1\%$ of planned budget left at plateau onset. Other PiEvo molecular seeds show different timing: some plateau only near the budget limit, so one case does not establish universal long-term collapse. On residence hall, all five best errors are $3.42$--$8.84$ times the reference, with $75.5$--$97.8\%$ of planned budget remaining at onset. Under the current decision tree, the S/L/N counts are $4/1/0$. The last-1,000-attempt audit is summarized in Appendix~\ref{app:decision-counts}; the earlier post-plateau summaries use different windows. Dolphin's five endpoints are within $4.4\%$ of the exhaustive optimum, including one exact optimum.

\subsection{Noise sensitivity}
\label{app:pievo-sigma}
Increasing $\sigma$ from $0.01$ to $0.9$ is an additional molecular comparison, not a main baseline. It can soften posterior concentration, with mean early/late exact-text effective counts of $8.213/9.053$, compared with $2.675/2.229$ at the original setting. The higher-noise setting averages only $0.4$ late principle updates versus $1.0$ originally (Table~\ref{tab:pievo-molecules}); increasing observation noise does not supply a stagnation detector. 
\begin{table}[h]
\centering
\caption{PiEvo molecular noise sensitivity. Belief entropies are attempt-weighted over the first/last 1,000 generated attempts. They measure posterior uncertainty, not scaffold coverage.}
\label{tab:pievo-sigma}
\scriptsize
\resizebox{\linewidth}{!}{%
\begin{tabular}{@{}lcccccc@{}}
\toprule
Method / condition & Top-1 & Top-10 & $H^{\mathrm{belief}}_e$ & $H^{\mathrm{belief}}_l$ & Scaffold $N_{\mathrm{eff},l}$ & $b_{\mathrm{plat}}$ \\
\midrule
PiEvo & $4.100\pm0.200$ & $4.057\pm0.189$ & $0.010\pm0.007$ & $0.000\pm0.000$ & $30.685\pm11.115$ & $3214.800\pm1509.945$ \\
PiEvo / Sigma 0.9 & $4.068\pm0.156$ & $4.040\pm0.152$ & $0.115\pm0.026$ & $0.008\pm0.004$ & $18.972\pm9.671$ & $3595.200\pm970.716$ \\
\bottomrule
\end{tabular}}
\end{table}

\section{Principle-Intervention Details}
\label{app:intervention-results}
\label{app:injection}
These tables report performance rather than exact reference hits. Molecular L2 is supplied either from the start or after $40\%$ of budget; networks use only the latter. These are targeted interventions, not guarantees that the proposed direction is fully implemented.
\begin{table}[h]
\centering
\caption{Molecular L2 timing comparison. Performance improvements, not exact reference identity, define intervention benefit.}
\label{tab:molhint}
\scriptsize
\resizebox{\linewidth}{!}{%
\begin{tabular}{@{}lccccc@{}}
\toprule
Method / condition & Top-1 & Top-10 & $U$ & $N_{\mathrm{eff},l}$ & $m_l$ \\
\midrule
ExLLM(GPT-5.4) / Hint from 40\% & $4.021\pm0.090$ & $3.992\pm0.095$ & $0.396\pm0.038$ & $3.717\pm1.083$ & $0.544\pm0.190$ \\
ExLLM(GPT-5.4) / Hint from start & $4.264\pm0.098$ & $4.201\pm0.088$ & $0.315\pm0.040$ & $1.541\pm0.511$ & $0.856\pm0.172$ \\
ExLLM(GPT-5.4) & $4.047\pm0.118$ & $4.027\pm0.110$ & $0.326\pm0.067$ & $1.908\pm1.116$ & $0.820\pm0.215$ \\
ExLLM(Sonnet 5) / Hint from 40\% & $4.485\pm0.298$ & $4.394\pm0.254$ & $0.151\pm0.016$ & $2.440\pm1.292$ & $0.636\pm0.234$ \\
ExLLM(Sonnet 5) / Hint from start & $4.617\pm0.183$ & $4.513\pm0.163$ & $0.170\pm0.033$ & $2.162\pm0.314$ & $0.572\pm0.064$ \\
ExLLM(Sonnet 5) & $3.975\pm0.008$ & $3.962\pm0.012$ & $0.122\pm0.011$ & $2.577\pm1.312$ & $0.664\pm0.204$ \\
ExLLM(DeepSeek) / Hint from 40\% & $4.699\pm0.000$ & $4.597\pm0.015$ & $0.424\pm0.085$ & $2.392\pm0.245$ & $0.564\pm0.079$ \\
ExLLM(DeepSeek) / Hint from start & $4.699\pm0.000$ & $4.584\pm0.019$ & $0.350\pm0.025$ & $2.154\pm0.436$ & $0.656\pm0.155$ \\
ExLLM(DeepSeek) & $3.998\pm0.049$ & $3.986\pm0.043$ & $0.312\pm0.072$ & $1.241\pm0.332$ & $0.936\pm0.092$ \\
ExLLM(Qwen) / Hint from 40\% & $3.994\pm0.098$ & $3.963\pm0.079$ & $0.470\pm0.061$ & $2.859\pm1.080$ & $0.644\pm0.177$ \\
ExLLM(Qwen) / Hint from start & $4.020\pm0.051$ & $3.960\pm0.055$ & $0.550\pm0.044$ & $3.990\pm2.027$ & $0.592\pm0.217$ \\
ExLLM(Qwen) & $4.091\pm0.109$ & $4.047\pm0.077$ & $0.338\pm0.094$ & $1.755\pm0.730$ & $0.808\pm0.209$ \\
\bottomrule
\end{tabular}}
\end{table}
\begin{table}[h]
\centering
\caption{Dolphin delayed L2 intervention. This is a continuation comparison, not a test of reproducing the reference node set.}
\label{tab:dolphin-hints}
\scriptsize
\resizebox{\linewidth}{!}{%
\begin{tabular}{@{}lcccc@{}}
\toprule
Method / condition & Top-1 error & Top-10 error & $U$ & $N_{\mathrm{eff},l}$ \\
\midrule
ExLLM(GPT-5.4) / Hint from 40\% & $4.077\times10^{-3}\pm3.422\times10^{-5}$ & $4.144\times10^{-3}\pm2.249\times10^{-5}$ & $0.213\pm0.031$ & $13.947\pm1.283$ \\
ExLLM(GPT-5.4) & $4.071\times10^{-3}\pm2.406\times10^{-5}$ & $4.147\times10^{-3}\pm2.617\times10^{-5}$ & $0.228\pm0.028$ & $13.971\pm1.419$ \\
ExLLM(Sonnet 5) / Hint from 40\% & $4.115\times10^{-3}\pm7.580\times10^{-5}$ & $4.196\times10^{-3}\pm7.556\times10^{-5}$ & $0.391\pm0.061$ & $14.292\pm0.928$ \\
ExLLM(Sonnet 5) & $4.115\times10^{-3}\pm7.582\times10^{-5}$ & $4.189\times10^{-3}\pm8.557\times10^{-5}$ & $0.420\pm0.035$ & $13.990\pm0.630$ \\
ExLLM(DeepSeek) / Hint from 40\% & $4.089\times10^{-3}\pm5.906\times10^{-5}$ & $4.172\times10^{-3}\pm4.753\times10^{-5}$ & $0.336\pm0.079$ & $15.383\pm1.991$ \\
ExLLM(DeepSeek) & $4.216\times10^{-3}\pm3.604\times10^{-5}$ & $4.320\times10^{-3}\pm3.408\times10^{-5}$ & $0.271\pm0.061$ & $14.214\pm1.602$ \\
ExLLM(Qwen) / Hint from 40\% & $4.174\times10^{-3}\pm8.443\times10^{-5}$ & $4.255\times10^{-3}\pm7.479\times10^{-5}$ & $0.554\pm0.098$ & $15.200\pm2.006$ \\
ExLLM(Qwen) & $4.234\times10^{-3}\pm1.073\times10^{-4}$ & $4.373\times10^{-3}\pm1.501\times10^{-4}$ & $0.482\pm0.115$ & $21.705\pm7.702$ \\
\bottomrule
\end{tabular}}
\end{table}
\begin{table}[h]
\centering
\caption{Residence hall delayed L2 intervention. This is a continuation comparison, not a test of reproducing the reference node set.}
\label{tab:residence_hall-hints}
\scriptsize
\resizebox{\linewidth}{!}{%
\begin{tabular}{@{}lcccc@{}}
\toprule
Method / condition & Top-1 error & Top-10 error & $U$ & $N_{\mathrm{eff},l}$ \\
\midrule
ExLLM(GPT-5.4) / Hint from 40\% & $5.112\times10^{-7}\pm3.635\times10^{-7}$ & $1.152\times10^{-6}\pm2.242\times10^{-7}$ & $0.395\pm0.065$ & $27.815\pm2.855$ \\
ExLLM(GPT-5.4) & $7.005\times10^{-7}\pm5.689\times10^{-7}$ & $1.352\times10^{-6}\pm6.547\times10^{-7}$ & $0.378\pm0.109$ & $26.752\pm3.587$ \\
ExLLM(Sonnet 5) / Hint from 40\% & $1.579\times10^{-6}\pm1.759\times10^{-6}$ & $3.600\times10^{-6}\pm2.861\times10^{-6}$ & $0.341\pm0.111$ & $20.962\pm1.078$ \\
ExLLM(Sonnet 5) & $1.514\times10^{-6}\pm1.084\times10^{-6}$ & $4.434\times10^{-6}\pm2.171\times10^{-6}$ & $0.303\pm0.109$ & $18.709\pm1.172$ \\
ExLLM(DeepSeek) / Hint from 40\% & $9.447\times10^{-7}\pm5.885\times10^{-7}$ & $2.660\times10^{-6}\pm1.405\times10^{-6}$ & $0.360\pm0.087$ & $25.469\pm5.719$ \\
ExLLM(DeepSeek) & $1.402\times10^{-6}\pm8.028\times10^{-7}$ & $3.846\times10^{-6}\pm1.887\times10^{-6}$ & $0.255\pm0.049$ & $23.065\pm5.774$ \\
ExLLM(Qwen) / Hint from 40\% & $4.716\times10^{-7}\pm5.194\times10^{-8}$ & $2.481\times10^{-5}\pm2.963\times10^{-5}$ & $0.846\pm0.048$ & $91.715\pm18.594$ \\
ExLLM(Qwen) & $9.961\times10^{-7}\pm6.875\times10^{-7}$ & $4.993\times10^{-5}\pm6.598\times10^{-5}$ & $0.731\pm0.087$ & $99.446\pm26.186$ \\
\bottomrule
\end{tabular}}
\end{table}

\begin{figure}[h]
\centering
\includegraphics[width=\linewidth]{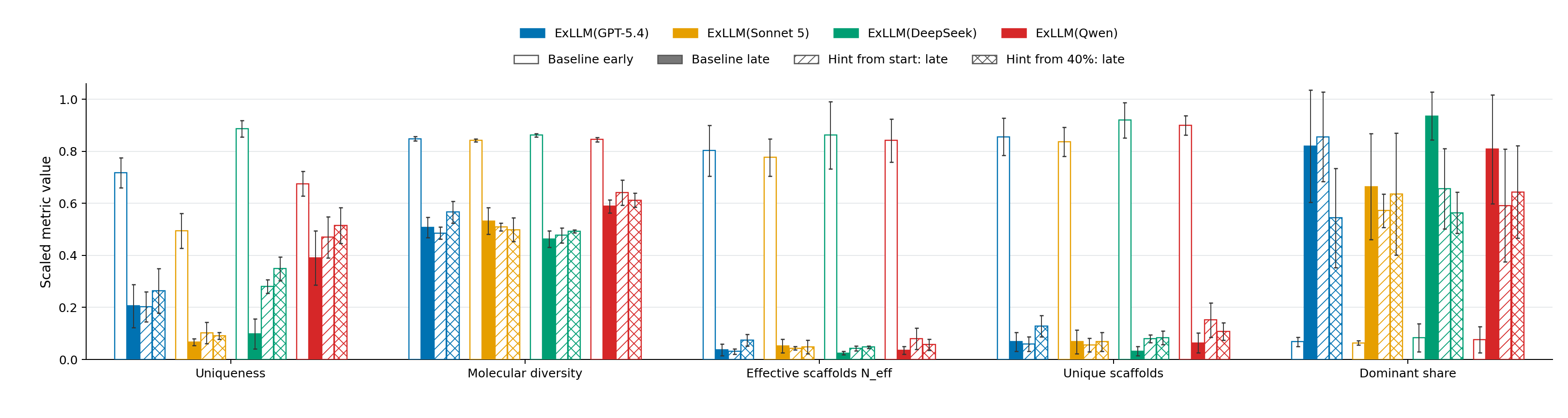}\\
\includegraphics[width=\linewidth]{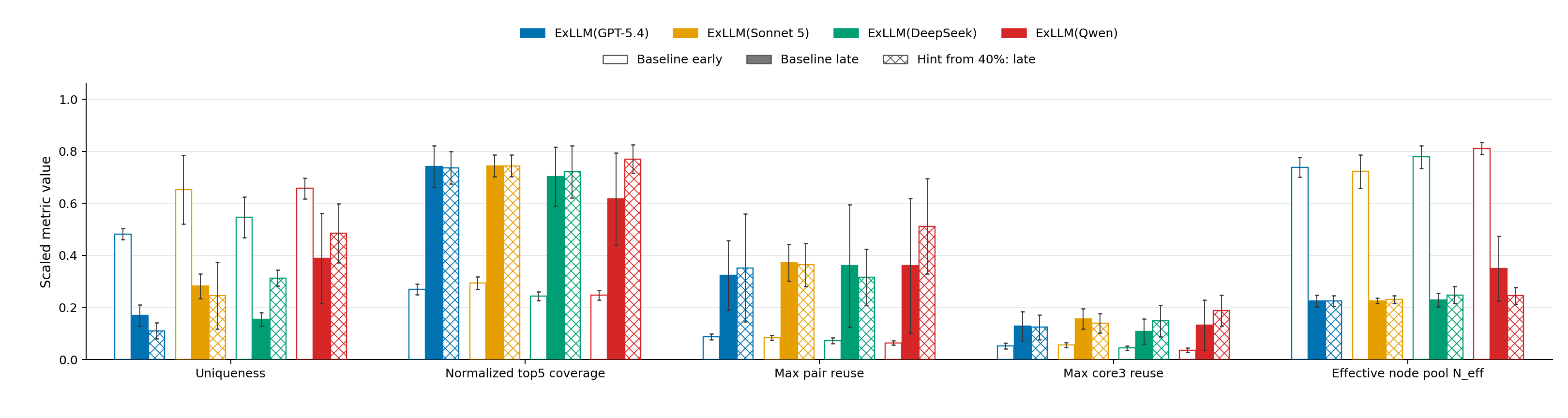}\\
\includegraphics[width=\linewidth]{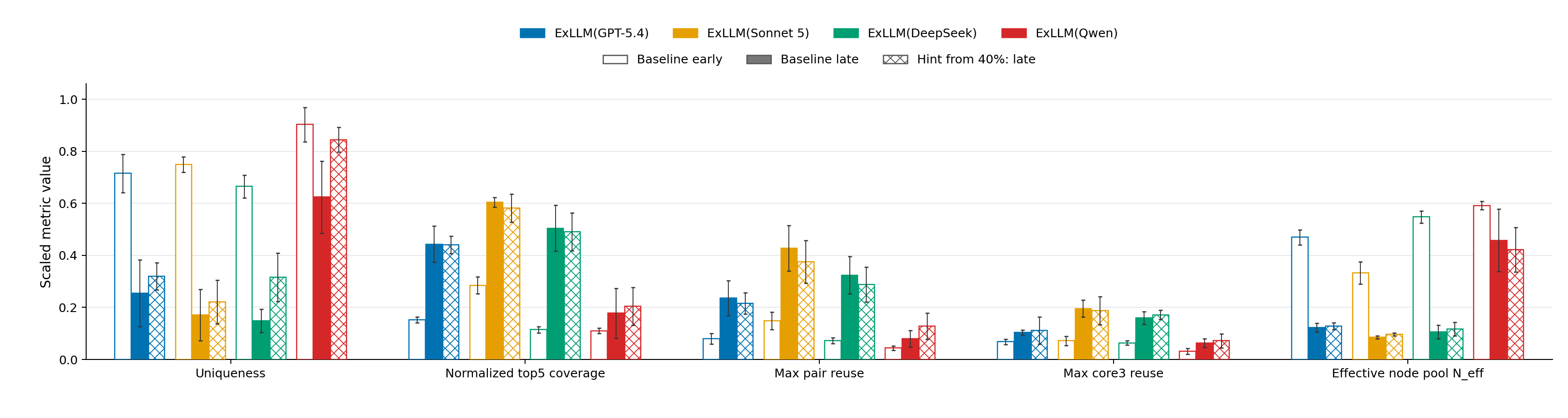}
\caption{Intervention concentration diagnostics. Within each backbone: baseline early, baseline late, from-start late when available, and delayed-hint late. Improved performance does not require an increase in principle diversity; a run may move to a better concentrated region.}
\label{fig:intervention-bars}
\end{figure}

\section{Kernel Results and Sensitivity}
\label{app:kernel-results}
\label{app:sensitivity}

\paragraph{Diagnostic set and concentration.}
At each evaluation, the kernel diagnostic set contains the ten fastest valid candidates found so far. Applying the route descriptor $\phi$ gives frequencies $\hat\pi_j^{\topten}(p)$ as in Eq.~\ref{eq:label-frequencies} and
\[
N_{\mathrm{eff},j}^{\topten}=\exp\!\left[-\sum_p \hat\pi_j^{\topten}(p)\ln\hat\pi_j^{\topten}(p)\right].
\]
Its range is $[1,10]$ for a full set; the dominant share is $\max_p\hat\pi_j^{\topten}(p)$. Concentration means final $N_{\mathrm{eff}}\le2$. We assess a run only after at least 20 valid kernels. 

\paragraph{Plateau and joint criterion.}
Let $G(t)$ be the best-so-far latency divided by expert latency,
$T_{\mathrm{obs}}$ the final observed evaluation attempt, and $B$
the planned evaluation budget. For a run ending above the reference,
define
\[
\tau_{\mathrm{ker}}
=
\min\left\{
t:
G(t)-G(T_{\mathrm{obs}})
\le
\epsilon_{\mathrm{ker}}[G(t)-1]
\right\},
\qquad \epsilon_{\mathrm{ker}}=0.10.
\]
The fraction of the planned budget remaining at plateau onset is
\[
R_{\mathrm{budget}}
=
\frac{B-\tau_{\mathrm{ker}}}{B}.
\]
The joint kernel criterion requires final $N_{\mathrm{eff}}\le 2$,
$R_{\mathrm{budget}}\ge 0.30$, and $G(T_{\mathrm{obs}})\ge 2$. All three must hold. This is a descriptive criterion, not a proof of a mathematical local optimum. For runs reaching $G\le1$, the plotting code marks the first reference-reaching point if no earlier point satisfies its plateau test; this control marker is not evidence that subsequent gains cease.

\begin{table}[h]
\centering\small
\caption{Joint kernel diagnostics for the four main ExLLM backbones, five seeds each. ``Late plateau'' means concentrated with $G\ge2$ but less than 30\% of budget remaining. The $G<2$ column does not imply matching the expert.}
\label{tab:kernel-verdicts}
\begin{tabular}{lrrrrr}
\toprule
Task & $n$ & Collapse & Late plateau & $G<2$, concentrated & Not concentrated \\
\midrule
GQA decode & 20 & 5 & 5 & 10 & 0 \\
MLA decode & 20 & 9 & 6 & 5 & 0 \\
MLA prefill & 20 & 19 & 1 & 0 & 0 \\
MoE FP8 & 20 & 15 & 4 & 0 & 1 \\
RMSNorm & 20 & 0 & 0 & 19 & 1 \\
\bottomrule
\end{tabular}
\end{table}

\begin{table}[h]
\centering\scriptsize
\caption{Kernel performance and final route concentration. ExLLM uses five seeds per task; PiEvo uses three. Late effective counts and collapse denominators include only runs with at least 20 valid kernels; latency summaries include runs with a recorded latency even when they fail that coverage requirement. $0/0$ means unassessable, not no collapse. GPT-4o is an additional control. No PiEvo RMSNorm results are supplied.}
\label{tab:kernel}
\resizebox{\linewidth}{!}{
\begin{tabular}{llrccc}
\toprule
Method & Task & $n$ & $G$ (latency / expert) & Late $N_{\mathrm{eff}}$ & Collapse / assessed \\
\midrule
ExLLM(GPT-5.4) & GQA decode & 5 & $28.015\pm15.670$ & $1.000\pm0.000$ & 3/5 \\
ExLLM(GPT-5.4) & MLA decode & 5 & $22.140\pm2.674$ & $1.000\pm0.000$ & 4/5 \\
ExLLM(GPT-5.4) & MLA prefill & 5 & $33.396\pm2.389$ & $1.000\pm0.000$ & 5/5 \\
ExLLM(GPT-5.4) & MoE FP8 & 5 & $37.548\pm1.011$ & $1.260\pm0.355$ & 5/5 \\
ExLLM(GPT-5.4) & RMSNorm & 5 & $0.467\pm0.013$ & $1.399\pm0.298$ & 0/5 \\
ExLLM(Claude Sonnet 5) & GQA decode & 5 & $0.762\pm0.062$ & $1.207\pm0.298$ & 0/5 \\
ExLLM(Claude Sonnet 5) & MLA decode & 5 & $1.151\pm0.317$ & $1.000\pm0.000$ & 0/5 \\
ExLLM(Claude Sonnet 5) & MLA prefill & 5 & $3.234\pm0.455$ & $1.000\pm0.000$ & 5/5 \\
ExLLM(Claude Sonnet 5) & MoE FP8 & 5 & $10.434\pm3.505$ & $1.168\pm0.377$ & 1/5 \\
ExLLM(Claude Sonnet 5) & RMSNorm & 5 & $0.449\pm0.009$ & $1.360\pm0.805$ & 0/5 \\
ExLLM(DeepSeek V4 Flash) & GQA decode & 5 & $1.528\pm0.529$ & $1.168\pm0.377$ & 0/5 \\
ExLLM(DeepSeek V4 Flash) & MLA decode & 5 & $7.054\pm2.786$ & $1.077\pm0.172$ & 1/5 \\
ExLLM(DeepSeek V4 Flash) & MLA prefill & 5 & $34.779\pm22.097$ & $1.000\pm0.000$ & 5/5 \\
ExLLM(DeepSeek V4 Flash) & MoE FP8 & 5 & $29.543\pm8.427$ & $1.317\pm0.709$ & 4/5 \\
ExLLM(DeepSeek V4 Flash) & RMSNorm & 5 & $0.457\pm0.010$ & $1.269\pm0.421$ & 0/5 \\
ExLLM(Qwen3.5-35B) & GQA decode & 5 & $44.126\pm14.952$ & $1.130\pm0.290$ & 2/5 \\
ExLLM(Qwen3.5-35B) & MLA decode & 5 & $73.652\pm2.103$ & $1.000\pm0.000$ & 4/5 \\
ExLLM(Qwen3.5-35B) & MLA prefill & 5 & $61.453\pm12.202$ & $1.000\pm0.000$ & 4/5 \\
ExLLM(Qwen3.5-35B) & MoE FP8 & 5 & $46.419\pm12.678$ & $1.000\pm0.000$ & 5/5 \\
ExLLM(Qwen3.5-35B) & RMSNorm & 5 & $0.461\pm0.007$ & $1.000\pm0.000$ & 0/5 \\
ExLLM(GPT-4o) & GQA decode & 5 & $2758.283\pm44.046$ & -- & 0/0 \\
ExLLM(GPT-4o) & MLA decode & 5 & $152.824\pm37.781$ & $1.000\pm0.000$ & 1/2 \\
ExLLM(GPT-4o) & MLA prefill & 5 & -- & -- & 0/0 \\
ExLLM(GPT-4o) & MoE FP8 & 5 & $70.345\pm0.936$ & -- & 0/0 \\
ExLLM(GPT-4o) & RMSNorm & 5 & $0.465\pm0.002$ & $1.000\pm0.000$ & 0/5 \\
PiEvo(Claude Sonnet 5) & GQA decode & 3 & $1.065\pm0.112$ & $1.678\pm0.309$ & 0/3 \\
PiEvo(Claude Sonnet 5) & MLA decode & 3 & $1.534\pm0.674$ & $1.128\pm0.222$ & 1/3 \\
PiEvo(Claude Sonnet 5) & MLA prefill & 3 & $2.303\pm1.044$ & $1.433\pm0.375$ & 1/3 \\
PiEvo(Claude Sonnet 5) & MoE FP8 & 3 & $22.785\pm5.584$ & $1.816\pm0.912$ & 1/3 \\
PiEvo(Gemini 2.5 Flash) & GQA decode & 3 & $1887.464\pm1634.618$ & -- & 0/0 \\
PiEvo(Gemini 2.5 Flash) & MLA decode & 3 & $173.297\pm0.958$ & -- & 0/0 \\
PiEvo(Gemini 2.5 Flash) & MLA prefill & 3 & $63.976\pm0.000$ & -- & 0/0 \\
PiEvo(Gemini 2.5 Flash) & MoE FP8 & 3 & $57.502\pm22.985$ & $1.976\pm0.838$ & 1/2 \\
\bottomrule
\end{tabular}}
\end{table}

\paragraph{Threshold sensitivity.}
The supplied sensitivity grid includes GPT-4o: 82 of 100 non-control baseline runs are assessable, versus 80/80 for the four main backbones. At the default settings it gives 49/82 joint cases (48/80 without GPT-4o). Holding the other settings fixed, changing the effective-count ceiling from 1.5 to 2 to 3 gives 47, 49 and 50 cases; changing the remaining-fraction threshold to 0.2/0.3/0.4/0.5 gives 55/49/43/36. Plateau tolerances 0.05/0.10/0.20 give 43/49/53 cases. The recent-20-valid-candidate diagnostic gives 46 cases at the default thresholds. Thus the exact count is threshold-dependent, but concentration with substantial performance headroom is not confined to one setting.

\begin{figure}[h]
\centering
\includegraphics[width=\linewidth]{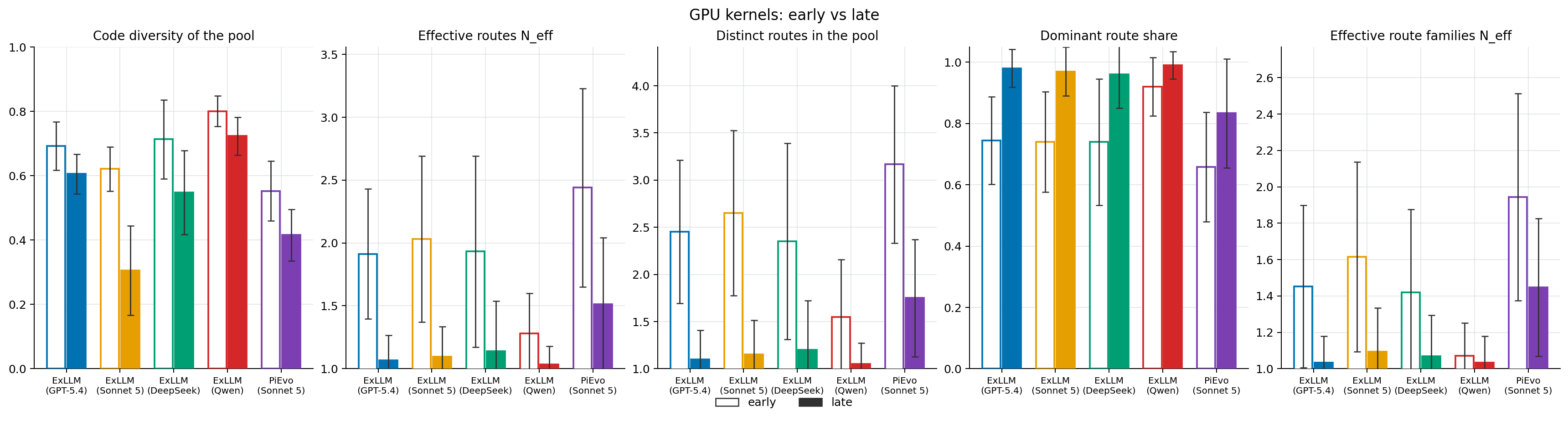}
\caption{Kernel early/late diagnostics in the fastest-ten sets. Empty/filled bars denote initial/final sets. Code diversity is mean pairwise token 5-gram Jaccard distance, not the fraction of distinct candidates. Fine-grained routes and their broader families are both shown.}
\label{fig:kernel-bars}
\end{figure}

\begin{figure}[h]
\centering
\includegraphics[width=\linewidth]{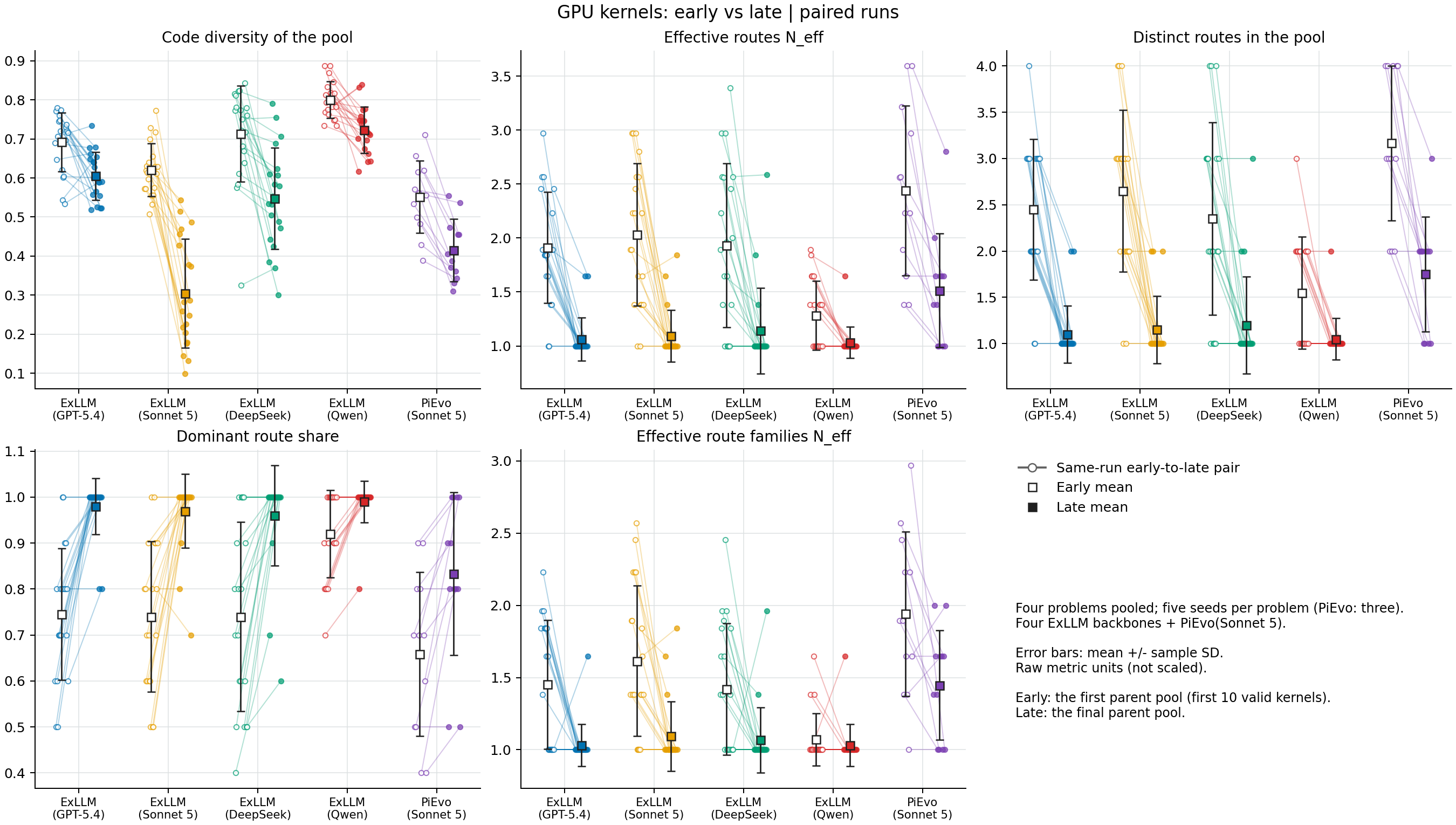}
\caption{The same kernel early/late diagnostics with paired individual runs. Differences across problems and seeds remain visible rather than being inferred solely from pooled means.}
\end{figure}

\paragraph{Intervention coverage and comparison.}
Table~\ref{tab:kernel-hint-results} uses only the four main backbones. Each start condition has 48 planned runs (four tasks, four backbones, three seeds); L2 and L3 each lack a usable final latency in two runs. Each onset condition has 41 matched runs, selected by the experiment's onset detector, not retrospectively by the final joint criterion above. Gain is baseline latency divided by hinted latency. Better means gain $>1.05$, worse means gain $<0.95$, and the remaining cases fall within that band. These cutoffs are on gain, not exactly a symmetric 5\% change in latency.

\paragraph{Onset branch point.} Onset arms branch from the baseline run once its search has narrowed. The branch point was fixed online, before any hinted run, as the first time the run's recent proposals stayed concentrated on one route for two consecutive windows of 20 valid kernels; it is found in 41 of the 48 baseline runs of the four main backbones with seeds 42--44, and the other seven have no onset arm. Checked against the conditions of Section~\ref{sec:diagnostic-rule}, every branch point already satisfies $\Narrow$: the diagnostic set of the ten fastest kernels has $N_{\mathrm{eff}}\le2$ at that moment (median $1.0$). In 36 of the 41 runs $\Headroom$ also holds, with best latency at least twice the expert's; the other five are Sonnet and DeepSeek runs already within twice the expert latency. About half of the branch points fall after the run's plateau onset in Figure~\ref{fig:kernel-budget} and half shortly before it. By the end of the budget, 28 of the 41 baseline runs meet the joint criterion, 7 are concentrated but still improving, and 6 converge near the expert.

\begin{table}[h]
\centering\small
\caption{Paired kernel intervention outcomes across the four main backbones. Geometric-mean gain above one favors the hint. Missing final latencies are excluded, not assigned neutral gains.}
\label{tab:kernel-hint-results}
\begin{tabular}{llrrrrr}
\toprule
Timing & Hint & Pairs / planned & Better & Worse & Within band & Geom. gain \\
\midrule
Start & L1 & 48/48 & 28 & 12 & 8 & 1.169 \\
Start & L2 & 46/48 & 14 & 21 & 11 & 0.833 \\
Start & L3 & 46/48 & 20 & 15 & 11 & 1.067 \\
Onset & L1 & 41/41 & 12 & 10 & 19 & 1.110 \\
Onset & L2 & 41/41 & 13 & 9 & 19 & 1.010 \\
Onset & L3 & 41/41 & 16 & 8 & 17 & 0.986 \\
\bottomrule
\end{tabular}
\end{table}

\begin{figure}[h]
\centering
\includegraphics[width=\linewidth]{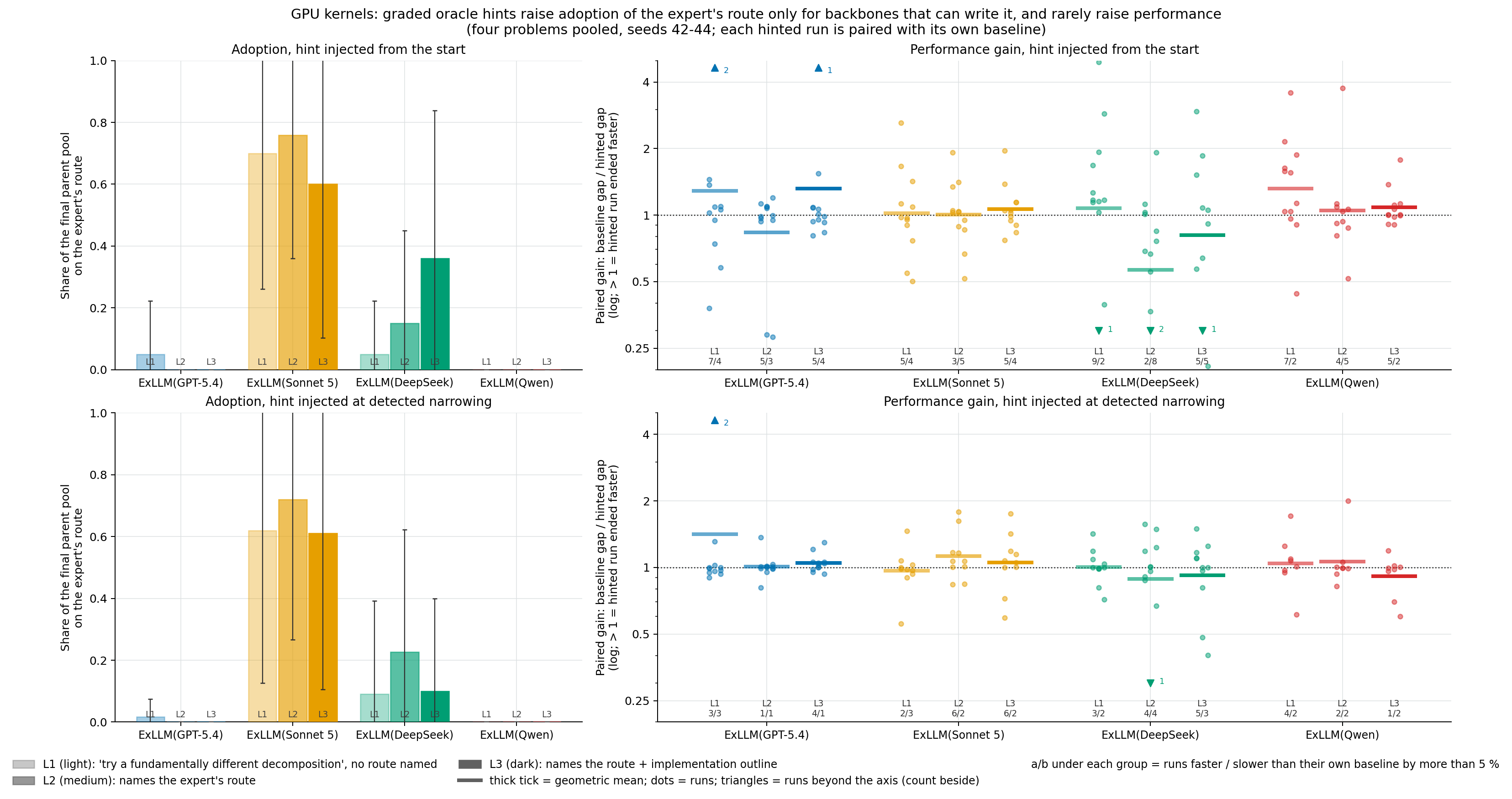}
\caption{Kernel principle interventions, paired with each run's baseline on seeds 42--44. Left: reference-route share in the final fastest-ten set (runs with at least 20 valid kernels). Right: baseline/hinted latency, with geometric means and individual runs; values above one indicate benefit. Top/bottom rows inject from the start/at detected narrowing. Adoption and performance need not improve together. Exact coverage and gain-band definitions are in Table~\ref{tab:kernel-hint-results}.}
\label{fig:kernel-hints}
\end{figure}

\begin{figure}[h]
\centering
\includegraphics[width=\linewidth]{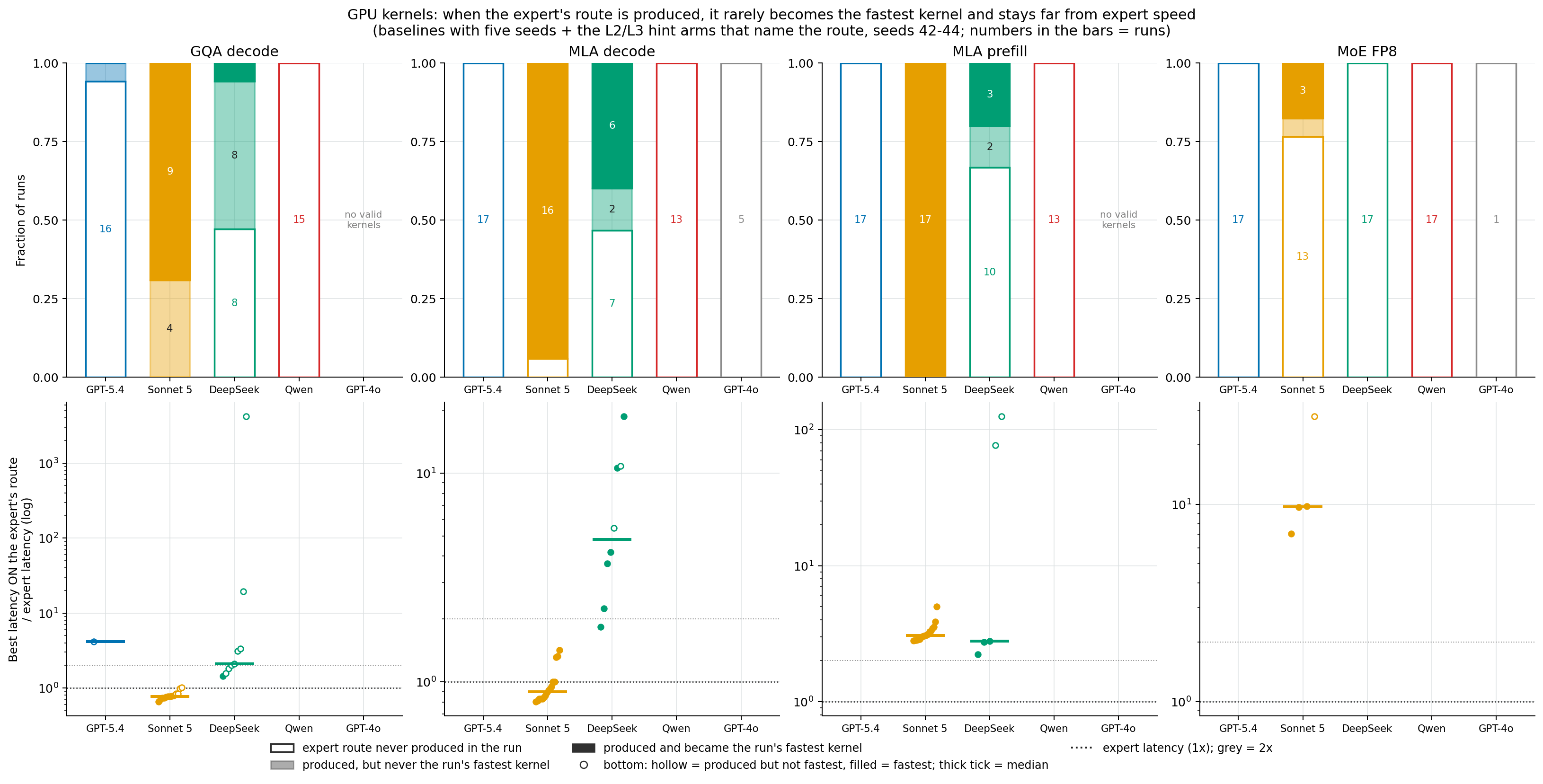}
\caption{Best achieved latency by executed route in baselines and L2/L3 interventions. Route membership does not ensure an expert-quality implementation.}
\label{fig:kernel-route-performance}
\end{figure}

\begin{figure}[h]
\centering
\includegraphics[width=\linewidth]{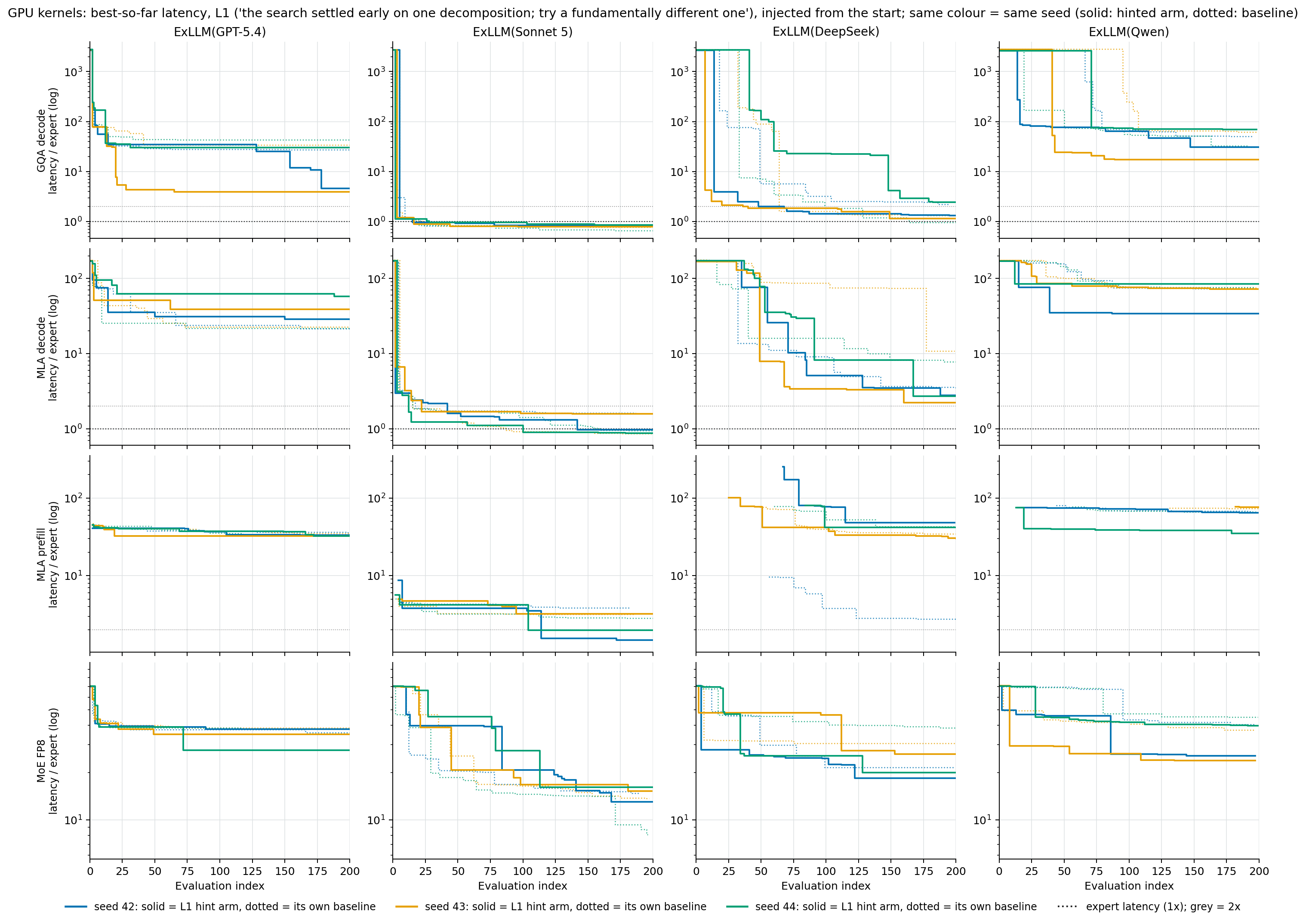}
\caption{L1 from the start: hinted runs (solid) and their baselines (dotted), matched by seed and color. Latency is divided by expert latency; lower is better.}
\end{figure}
\begin{figure}[h]
\centering
\includegraphics[width=\linewidth]{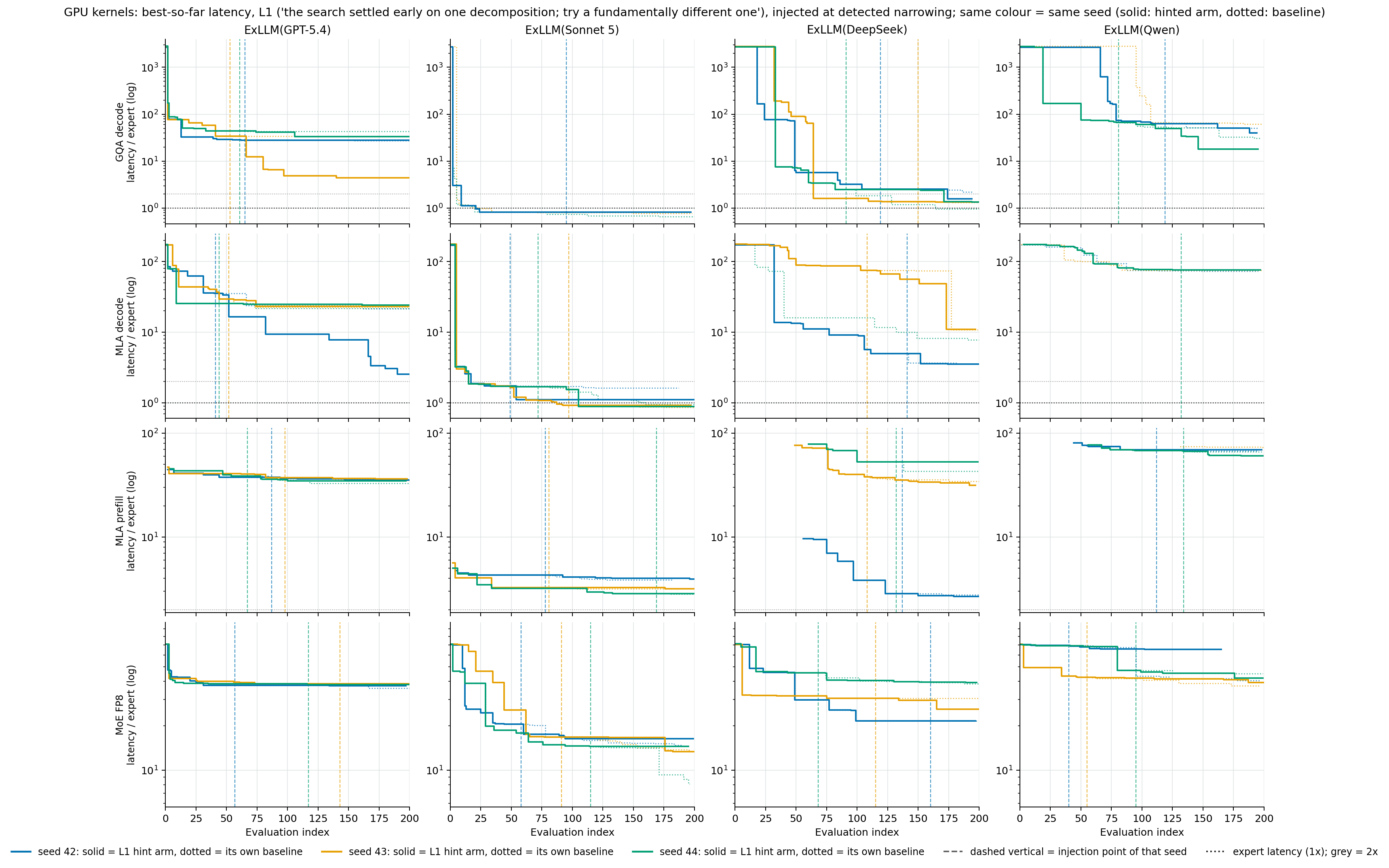}
\caption{L1 at detected narrowing. Dashed vertical lines mark injection; other conventions match the from-start comparison.}
\end{figure}
\begin{figure}[h]
\centering
\includegraphics[width=\linewidth]{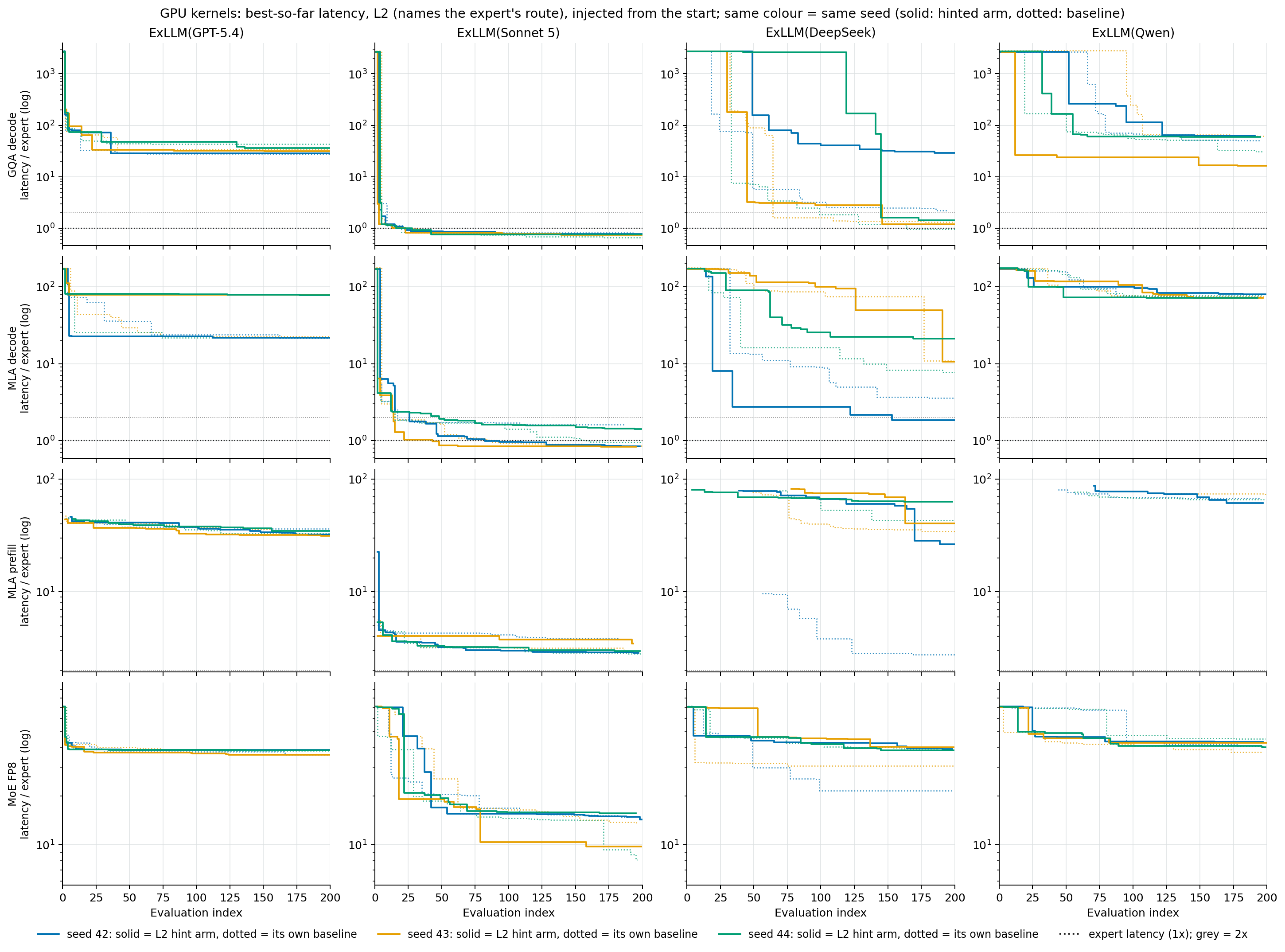}
\caption{L2 from the start, paired with baseline trajectories.}
\end{figure}
\begin{figure}[h]
\centering
\includegraphics[width=\linewidth]{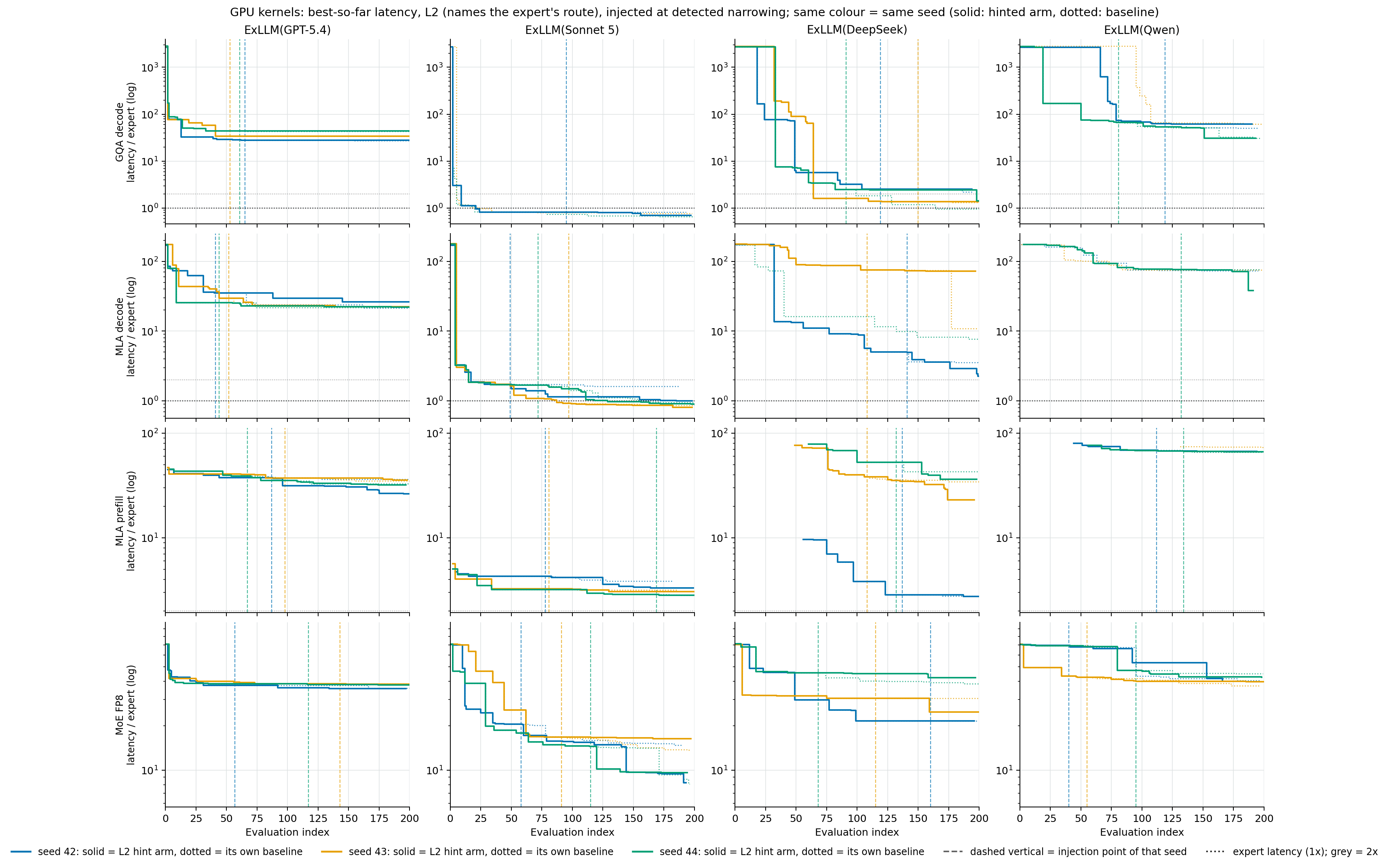}
\caption{L2 at detected narrowing, paired with baseline trajectories.}
\end{figure}
\begin{figure}[h]
\centering
\includegraphics[width=\linewidth]{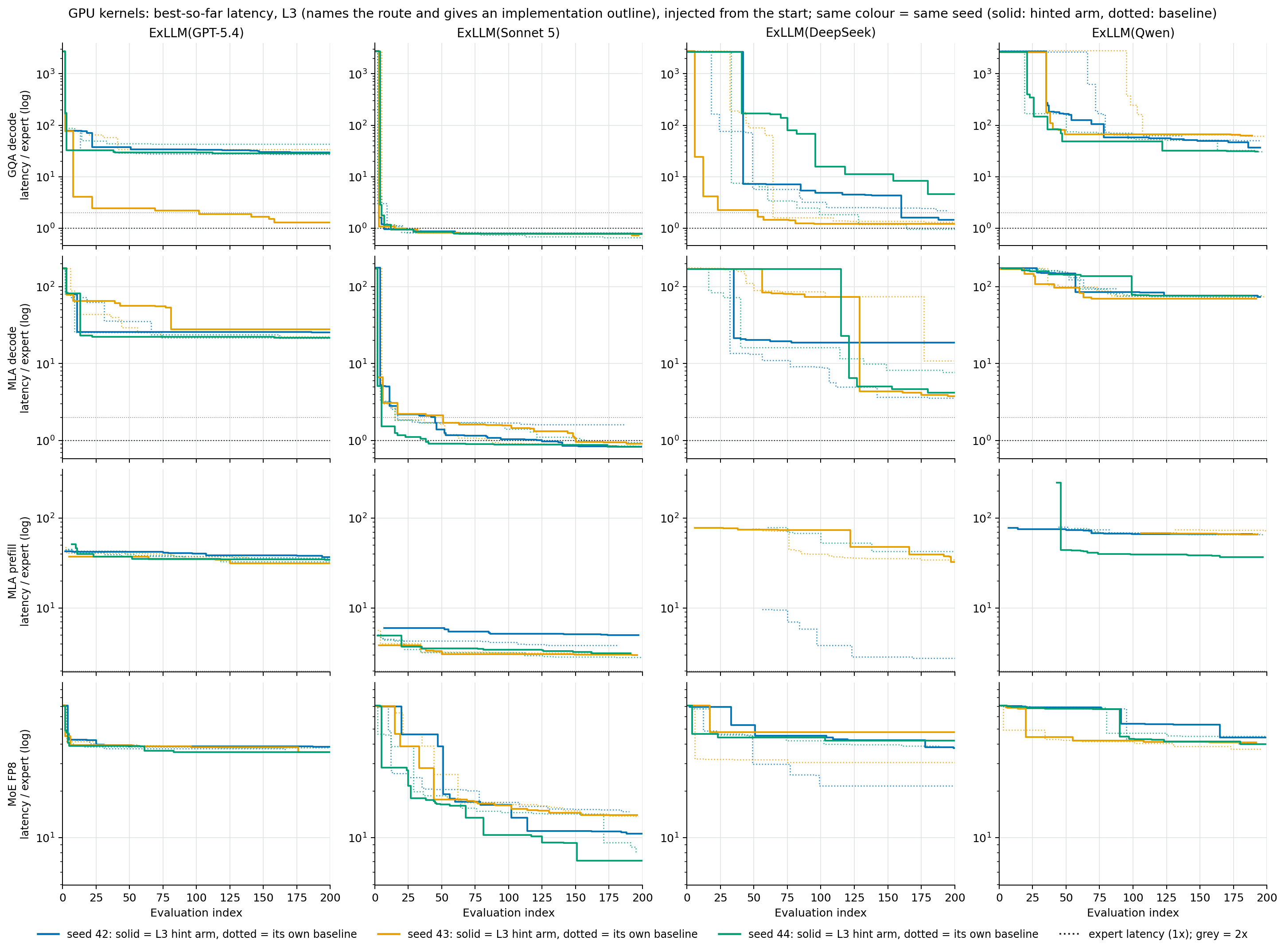}
\caption{L3 from the start, paired with baseline trajectories.}
\end{figure}
\begin{figure}[h]
\centering
\includegraphics[width=\linewidth]{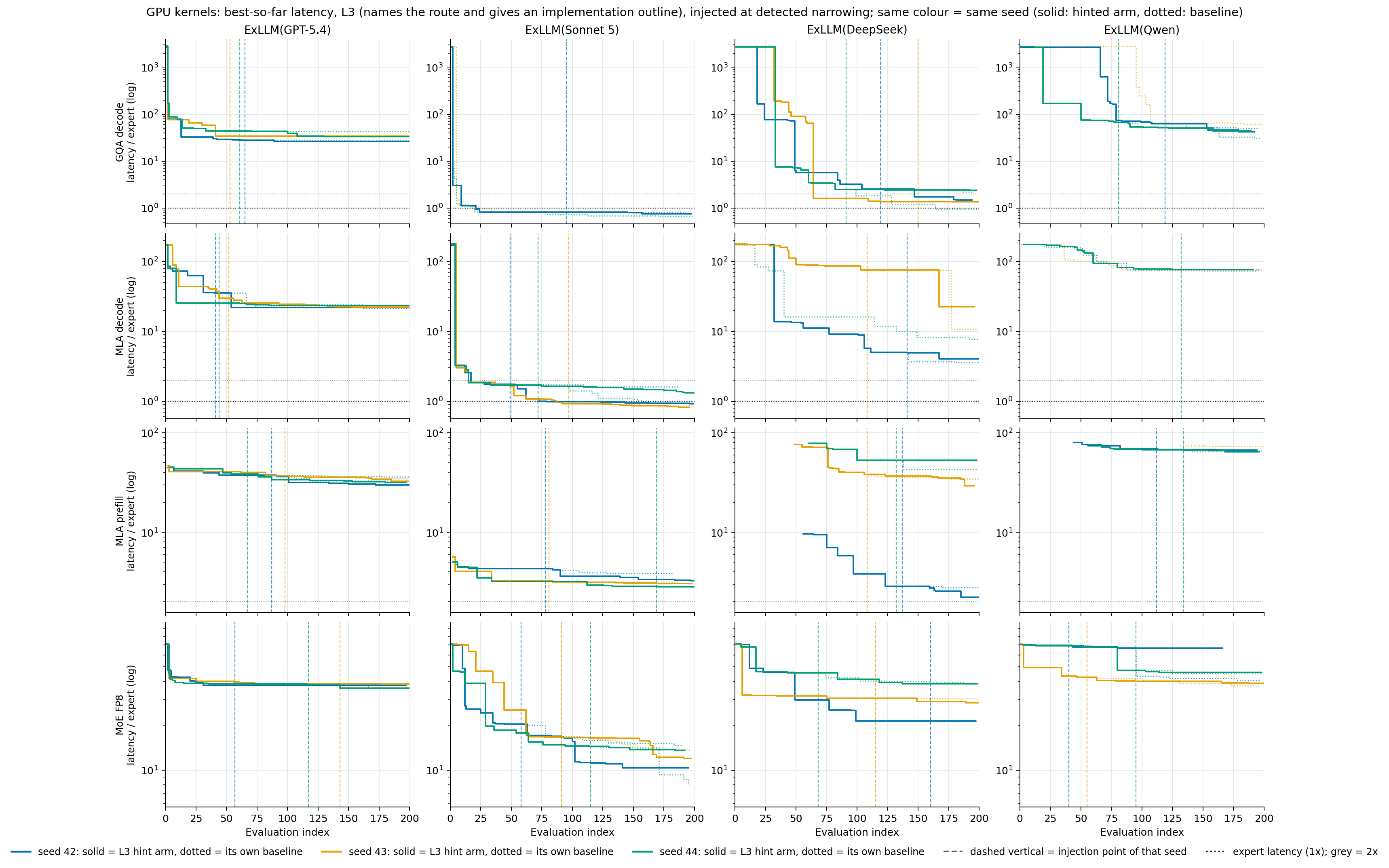}
\caption{L3 at detected narrowing, paired with baseline trajectories.}
\end{figure}

\section{First-Attempt Quality and Subsequent Development}
\label{app:mech}
Events are the first appearances of a route after the run's first route in the 80 ExLLM baseline runs of the four main backbones (four operators, five seeds): 186 events in 70 runs; the other ten runs used a single route throughout. $r$ and $N_{\mathrm{after}}$ follow Eq.~(\ref{eq:firstshot}); the control keeps the events with at least 50 attempts left when the route appeared. The expert's route accounts for 18 events, 16 from Sonnet and 2 from DeepSeek; 16 start slower than the incumbent and 11 later produce the run's fastest kernel (Table~\ref{tab:firstshot-expert}).
\begin{figure}[h]
\centering
\includegraphics[width=\linewidth]{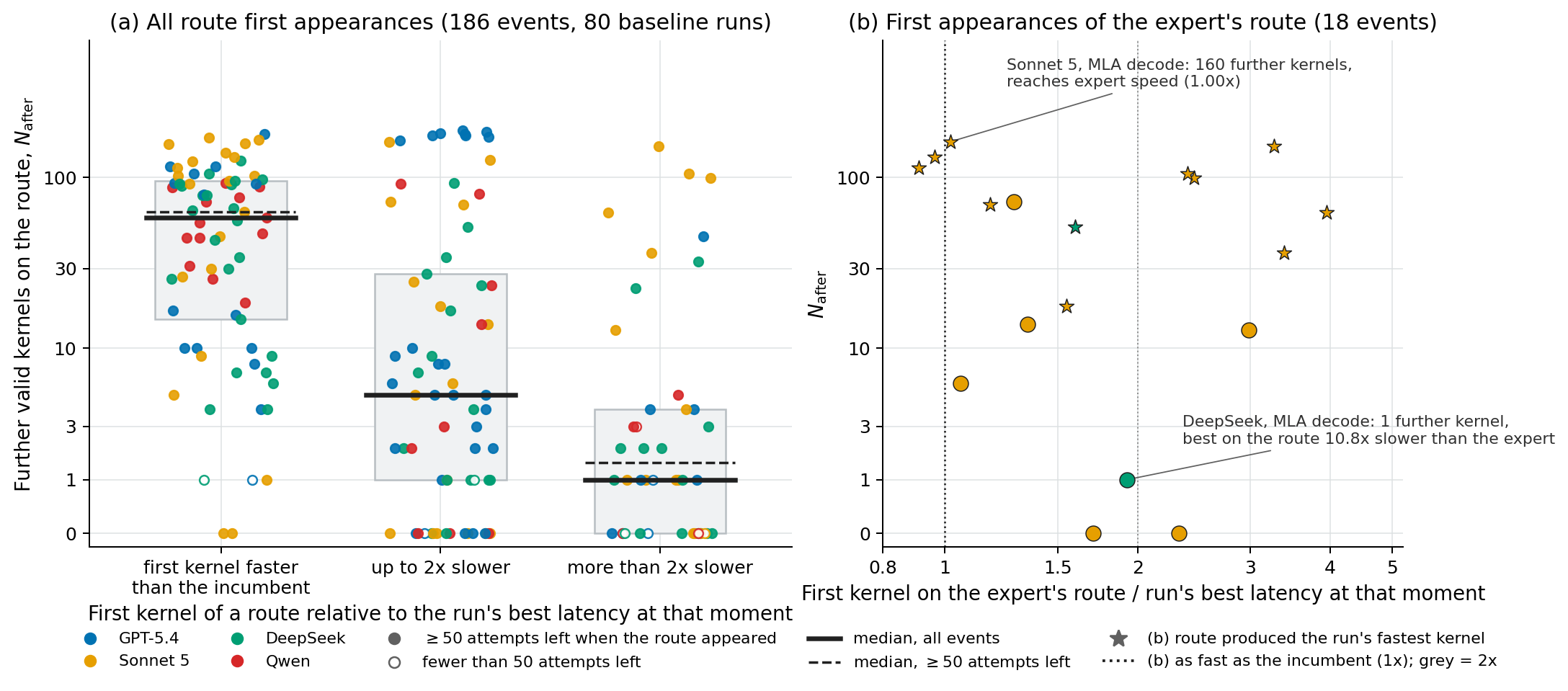}
\caption{Kernel first-attempt quality versus subsequent development. (a) All 186 events, grouped by the ratio of the route's first kernel to the run's best latency at that moment. Dots are events coloured by backbone (filled: at least 50 attempts left, hollow: fewer); boxes span the interquartile range; the solid line is the median and the dashed line the median of the control. The vertical axis is logarithmic in $1+N_{\mathrm{after}}$ so that zero can be shown. (b) The 18 first appearances of the expert's route; stars mark routes that later produced the run's fastest kernel.}
\label{fig:mechanism}
\end{figure}
\begin{table}[h]
\centering
\caption{Further development of a route by the quality of its first kernel (186 events, 80 baseline runs, five seeds). IQR is the interquartile range of $N_{\mathrm{after}}$; the last two columns restrict to events with at least 50 attempts left.}
\label{tab:firstshot}
\resizebox{\linewidth}{!}{
\begin{tabular}{lrrrrrr}
\toprule
First kernel of the route vs.\ the run's best so far & events & never developed again & median $N_{\mathrm{after}}$ & IQR & events with $\ge 50$ attempts left & their median $N_{\mathrm{after}}$ \\
\midrule
First kernel faster than the incumbent & 73 & 3\% & 59 & 15--96 & 71 & 64 \\
Up to 2x slower & 69 & 23\% & 5 & 1--28 & 67 & 5 \\
More than 2x slower & 44 & 36\% & 1 & 0--4 & 36 & 1.5 \\
\midrule
All events & 186 & 18\% & 9.5 & 1--78.5 & 174 & 14 \\
\bottomrule
\end{tabular}}
\end{table}
\begin{table}[h]
\centering
\caption{The 18 first appearances of the expert's route in the baseline runs: evaluation attempt of the first kernel, its ratio $r$ to the run's best latency at that moment, further kernels on the route, and the best latency reached on the route relative to the expert ($\star$: the route produced the run's fastest kernel).}
\label{tab:firstshot-expert}
\small
\begin{tabular}{llrrrrl}
\toprule
Problem & Backbone & seed & attempt & $r$ & $N_{\mathrm{after}}$ & best on the route / expert \\
\midrule
GQA decode & Sonnet 5 & 42 & 22 & 1.35 & 14 & 0.98 \\
GQA decode & Sonnet 5 & 43 & 15 & 3.39 & 37 & 0.77\,$\star$ \\
GQA decode & Sonnet 5 & 44 & 14 & 3.95 & 63 & 0.66\,$\star$ \\
GQA decode & Sonnet 5 & 45 & 19 & 2.99 & 13 & 1.01 \\
GQA decode & Sonnet 5 & 46 & 13 & 1.55 & 18 & 0.79\,$\star$ \\
MLA decode & Sonnet 5 & 43 & 16 & 3.27 & 151 & 0.86\,$\star$ \\
MLA decode & Sonnet 5 & 44 & 12 & 2.45 & 99 & 0.95\,$\star$ \\
MLA decode & Sonnet 5 & 45 & 12 & 1.02 & 160 & 1.00\,$\star$ \\
MLA decode & Sonnet 5 & 46 & 5 & 2.39 & 105 & 1.33\,$\star$ \\
MLA decode & DeepSeek & 42 & 42 & 1.93 & 1 & 10.79 \\
MLA prefill & Sonnet 5 & 43 & 5 & 1.06 & 6 & 4.16 \\
MLA prefill & Sonnet 5 & 43 & 127 & 2.32 & 0 & 7.35 \\
MLA prefill & Sonnet 5 & 44 & 5 & 1.28 & 73 & 3.20 \\
MLA prefill & Sonnet 5 & 44 & 76 & 1.18 & 70 & 2.83\,$\star$ \\
MLA prefill & Sonnet 5 & 45 & 7 & 0.96 & 131 & 2.85\,$\star$ \\
MLA prefill & Sonnet 5 & 45 & 58 & 1.70 & 0 & 5.46 \\
MLA prefill & Sonnet 5 & 46 & 28 & 0.91 & 113 & 3.55\,$\star$ \\
MLA prefill & DeepSeek & 42 & 74 & 1.60 & 52 & 2.78\,$\star$ \\
\bottomrule
\end{tabular}
\end{table}

\section{Saved experience summaries}
\label{app:experience}
The following excerpts illustrate restrictive rules in the final Residence Hall experience; they are examples rather than an exhaustive coding of all runs. Claude records \emph{Trusted core pool (zero error): \{205, 184, 113, 25, 60, 169, 104, 30, 97\}} and a list of banned nodes. The phrase \emph{zero error} is the model's wording, not an evaluator-verified claim. DeepSeek states that candidates must be five distinct integers from \emph{whitelist \{42, 87, 118, 139, 143, 168, 196, 211\}}. GPT-5.4 recommends \emph{keep 143 fixed} and \emph{conservative one-node, role-preserving edits around stable cores}. Qwen records \emph{strict arithmetic progressions with uniform spacing}, a rule with no explicit network-role justification.

Hinted runs need not faithfully restate the supplied principle to benefit from it. The saved-summary analysis reports no clear reference-principle alignment in Qwen's five Residence Hall summaries, and memorized narrow rules in four, despite lower mean error. GPT-5.4 also retains restrictive core-editing language. These observations distinguish the content of the stored summary from the actual proposals, scores, and potential influence of a separately prepended hint.

\subsection{Performance and budget}
\paragraph{Molecular fitness: sum of normalized objectives.}\mbox{}\\
Let $a_d(x)$ be raw objective $d$, with fixed bounds $L_d,U_d$, and $\widetilde a_d(x)=(a_d(x)-L_d)/(U_d-L_d)$. Orient each objective toward maximization:
\begin{equation}
v_d(x)=
\begin{cases}
\widetilde a_d(x),&d\text{ is maximized},\\
1-\widetilde a_d(x),&d\text{ is minimized},
\end{cases}
\qquad
f(x)=\sum_{d=1}^{M}v_d(x),\qquad z_d(x)=1-v_d(x).
\label{eq:metric-fitness}
\end{equation}
The evaluator stores $\mathbf z(x)$ as a minimization vector. For the five molecular objectives,
\begin{equation}
\begin{split}
f(x)={}&1-\frac{\mathrm{SA}(x)-1}{9}
 +(1-\mathrm{DRD2}(x))+\mathrm{QED}(x)\\
&+(1-\mathrm{GSK3}\beta(x))+\mathrm{JNK3}(x).
\end{split}
\end{equation}
SA has bounds $[1,10]$ and the other objectives $[0,1]$, so $f\in[0,5]$, with larger better. Normalization follows previous work \citep{ran2025exllm}.

\paragraph{Top-1 and top-10 fitness.}\mbox{}\\
Let $\mathcal X_t$ be the archive of distinct valid candidates at budget $t$, ordered as $x_{(1)},x_{(2)},\ldots$ by decreasing $f$, and $k_t=\min(K,|\mathcal X_t|)$. Then
\begin{equation}
F_K(t)=\frac{1}{k_t}\sum_{i=1}^{k_t}f(x_{(i)}),
\qquad F_1(t)=\max_{x\in\mathcal X_t}f(x).
\label{eq:metric-topk}
\end{equation}
Top-10 $F$ means $F_{10}$, the average over ten distinct candidates, not ten generations. At the final budget, $F_1(T)=F_T$ of Eq.~(\ref{eq:headroom}). Molecular $F_K\in[0,5]$, with larger better.

\paragraph{Sentinel approximation error and top-10 error.}\mbox{}\\
For network state $y_{cv}$ at condition $c$ and node $v$, write $\bar y_c=N^{-1}\sum_v y_{cv}$ and $\widehat y_c(x)=r^{-1}\sum_{v\in x}y_{cv}$ for a subset of size $r$. The unweighted evaluator returns
\begin{equation}
e(x)=\frac{\sum_c(\widehat y_c(x)-\bar y_c)^2}{\sum_c\bar y_c},
\qquad
\mathrm{Err}_K(t)=\frac{1}{k_t}\sum_{i=1}^{k_t}e(x_{[i]}),
\label{eq:metric-error}
\end{equation}
where $x_{[i]}$ are ordered by increasing error. The denominator is positive on both datasets; error is nonnegative with no universal upper bound, and smaller is better. This is not variance-normalized MSE. The theoretical utility $f(x)=-\ln e(x)$ in Table~\ref{tab:instantiation} gives the same ranking as the implementation's score $-e(x)$; tables average raw errors, not log errors.

\paragraph{Kernel latency and reference gap.}\mbox{}\\
For workload latencies $\ell_w(x)>0$, $w=1,\ldots,W$, geometric-mean latency and the reference ratio are
\begin{equation}
\ell(x)=\exp\!\left(\frac{1}{W}\sum_{w=1}^{W}\ln\ell_w(x)\right),
\qquad
G(t)=\frac{\min_{x\in\mathcal X_t}\ell(x)}{\ell(x_{\mathrm{ref}})}.
\label{eq:metric-gap}
\end{equation}
Smaller is better: $G=1$ matches the expert and $G<1$ beats it. With $f=-\ln\ell$, $\Delta_T=\ln G(T)$. Replacing latency by error gives the analogous network error ratio and log gap. 

\paragraph{Hypervolume (HV).}\mbox{}\\
Using the normalized minimization objectives $\mathbf z(x)$ from Eq.~\ref{eq:metric-fitness}, let $\mathcal Z_{50}^{\mathrm{ND}}(b)$ be the nondominated objective vectors in $\mathcal A_{50}(b)$. With reference vector $\mathbf z_{\mathrm{ref}}=(1.1,\ldots,1.1)$, we compute
\begin{equation}
\mathrm{HV}_{50}(b)=\lambda_M\!\left(
\bigcup_{\mathbf z\in\mathcal Z_{50}^{\mathrm{ND}}(b)}
\prod_{d=1}^{M}[z_d,1.1]\right),
\end{equation}
where $\lambda_M$ denotes $M$-dimensional volume; a vector outside the reference box contributes no volume. Higher HV indicates better joint objective coverage. For normalized objectives in $[0,1]$, $0\le\mathrm{HV}_{50}\le1.1^M$; with five objectives the upper bound is approximately $1.611$, not $1$. The reported values are not rescaled to $[0,1]$.

\paragraph{Reference-region hits.}\mbox{}\\
For run $s$, $h_s=\mathbf 1[F_1^{(s)}(T)>4.6]$. The reported count is $\sum_{s=1}^{n}h_s$ out of $n$, e.g., $3/5$. It measures whether any evaluated molecule enters the high-score region, not whether it exactly reproduces the reference structure. Larger is better.
\section{Metric definitions}
\label{app:metrics}

\subsection{Selected-population diagnostics}
\label{app:population-metrics}
The following metrics use the same selected-50 reconstruction as the molecular and network budget plots.

\paragraph{Reconstructed sets and timing.}\mbox{}\\
Let $\mathcal X_b$ be the distinct valid candidates observed by evaluation budget $b$, and $\mathcal A_{50}(b)$ the diagnostic selection of 50 candidates. Duplicate molecules are canonicalized; network candidates are deduplicated by unordered node set. The top-10 comparison selects the ten best scalar fitness values separately.

Early molecular snapshots follow the first complete offspring generation. PiEvo networks have no initial population; their early snapshot is the first complete round containing at least 50 distinct candidates. Late snapshots use the current final checkpoint. 

Budget curves sample every 50 evaluations, from 0 to 5,000 for molecules, 3,100 for Dolphin and 10,850 for residence hall. Stars mark mean per-seed plateau onset on the mean Top-1 curve. Residence-hall error is displayed on a log axis, but means and SD are calculated in raw error units.

\paragraph{Scaffold effective count, richness and dominant share.}\mbox{}\\
For canonical Bemis--Murcko scaffold frequency $p_s(b)=|\{x\in\mathcal A_{50}(b):\phi(x)=s\}|/50$,
\begin{equation}
\begin{aligned}
N_{\mathrm{eff}}^{\mathrm{scaf}}(b)&=\exp\!\left[-\sum_s p_s(b)\ln p_s(b)\right],\\
S_{50}(b)&=|\{s:p_s(b)>0\}|,\qquad m_{50}(b)=\max_s p_s(b).
\end{aligned}
\end{equation}
The ranges are $1\le N_{\mathrm{eff}}^{\mathrm{scaf}}\le S_{50}\le50$ and $1/50\le m_{50}\le1$. Smaller effective count and richness, or larger share, indicate concentration, not necessarily failure. Similar but nonidentical scaffolds can still implement the same strategy. Molecular diversity uses Eq.~\ref{eq:metric-tanimoto} on the same $\mathcal A_{50}(b)$. Bars divide effective and unique counts by 50; curves retain raw counts.

\paragraph{Node effective count and reuse.}\mbox{}\\
For $q=50$ distinct sets of $r$ nodes, let $c_v=\sum_{x\in\mathcal A_{50}}\mathbf1[v\in x]$ and $p_v=c_v/(qr)$. Then
\begin{equation}
\begin{aligned}
N_{\mathrm{eff}}^{\mathrm{node}}&=\exp\!\left(-\sum_v p_v\ln p_v\right),\\
R_2&=\max_{u<v}\frac{\sum_{x\in\mathcal A_{50}}\mathbf1[\{u,v\}\subseteq x]}q,\\
R_3&=\max_{u<v<w}\frac{\sum_{x\in\mathcal A_{50}}\mathbf1[\{u,v,w\}\subseteq x]}q.
\end{aligned}
\end{equation}
$R_2,R_3\in[1/50,1]$ for 50 nonempty sets of size at least three; high values indicate a repeatedly selected core. If $V_5$ contains the five most-used nodes, normalized top-five coverage is
\begin{equation}
C_5=\frac{\sum_{v\in V_5}c_v/(qr)}{C_5^{\max}(N,r,q)}.
\end{equation}
For 50 distinct sets, $C_5^{\max}=0.775$ on Dolphin and $0.804$ on residence hall, so $C_5\in[0,1]$. These are graph-independent combinatorial maxima given $N,r,q$, not fitted to a run. The corresponding minimum effective counts are $7.837$ and $7.981$, respectively: the minimum is not simply $r$, because all 50 sets must differ. Upper bounds are $\min(N,qr)$. Bars divide node effective count by $N$; trajectories retain raw counts. These metrics describe node identities and shared subsets; low effective node count alone does not imply poor performance.

\paragraph{Early and late uniqueness.}\mbox{}\\
Let $G$ count generated candidates and $R(G)$ count proposals repeated relative to prior history. For a generation-count window $[a,b]$,
\begin{equation}
U_{[a,b]}=1-\frac{R(b)-R(a)}{b-a}.
\end{equation}
We use the first and last 1,000 generated candidates after initialization, not 1,000 evaluations. Counters at unsaved boundaries are linearly interpolated within the boundary batches, except for ReEvo: its generated count is incremented for a whole batch before duplicate counts are updated candidate by candidate. ReEvo's first/last-1,000 windows are therefore reconstructed exactly from ordered saved proposals, with canonical-SMILES deduplication and all final counts checked against the logs. Global uniqueness and cumulative-uniqueness curves retain the logged values. Thus a high unique-candidate rate and a concentrated scaffold distribution are compatible.

\paragraph{Retrospective plateau onset and reference headroom.}\mbox{}\\
For observed endpoint $T_{\mathrm{obs}}$ and best molecular fitness $F_1(b)$, define
\begin{equation}
b_{\mathrm{plat}}=\min\{b:F_1(T_{\mathrm{obs}})-F_1(b)\le\epsilon_F\},\qquad
\epsilon_F=0.01.
\end{equation}
For best network error $E_1(b)$ use
\begin{equation}
b_{\mathrm{plat}}=\min\left\{b:
\frac{E_1(b)-E_1(T_{\mathrm{obs}})}{E_1(b)}\le0.01\right\},\qquad
R_{\mathrm{budget}}=\frac{B-b_{\mathrm{plat}}}{B},
\end{equation}
where $B$ is the planned budget. A working long-plateau condition is $R_{\mathrm{budget}}\ge0.3$. Molecular headroom uses $F_1<4.3$, a gap greater than $0.3985$ to the reference. Both networks use $E_1/E_{\mathrm{ref}}\ge2$, with the exhaustive Dolphin reference and $E_{\mathrm{ref}}=2.3817544421\times10^{-7}$ for residence hall. A gap does not establish restriction, and a plateau near a good reference is not classified as harmful simply because it is long. The graded verdict is defined in Section~\ref{sec:diagnostic-rule}, with structural concentration thresholds in Appendix~\ref{app:diagnostic-details}. 

\paragraph{Diagnostic support rates.}\mbox{}\\
Let $n_S,n_L,n_N$ count strong support, limited support, and collapse not established in the specified comparison, and let $n_{\mathrm{assess}}=n_S+n_L+n_N$. Then
\begin{equation}
P_{\mathrm{strong}}=\frac{n_S}{n_{\mathrm{assess}}},\qquad
P_{\mathrm{support}}=\frac{n_S+n_L}{n_{\mathrm{assess}}}.
\end{equation}
Both lie in $[0,1]$ and are descriptive diagnostic frequencies, not probabilities that collapse is true. Higher values indicate more evidence of the failure mode, not better optimization. Runs with insufficient valid candidates to assess the kernel structural diagnostic are excluded from this denominator and reported separately. Aggregate non-control rates exclude Dolphin and RMSNorm. Limited support is not confirmed collapse, and not established is not proof of its absence. Table entries headed Remaining (\%) are $100R_{\mathrm{budget}}$.

\subsection{First-shot and intervention metrics}
\paragraph{First-shot quality and subsequent investment.}\mbox{}\\
\label{app:mech-def}
For a kernel route $p$, let $t_{\mathrm{first}}(p)=\min\{i:\phi(x_i)=p\}$. Its first-shot ratio and subsequent investment are
\begin{equation}\label{eq:firstshot}
r(p)=\frac{\ell(x_{t_{\mathrm{first}}(p)})}{\min_{i<t_{\mathrm{first}}(p)}\ell(x_i)},
\qquad
N_{\mathrm{after}}(p)=\sum_{i>t_{\mathrm{first}}(p)}\mathbf1[\phi(x_i)=p].
\end{equation}
The first route is excluded because it has no prior comparator. Smaller $r$ means a stronger first attempt: $r<1$ beats the prior best, whereas $r>1$ is slower. The displayed groups are $r<1$, $1\le r<2$, and $r\ge2$. Investment is a nonnegative count bounded by remaining attempts; it measures development effort, not performance. The analysis contains 186 route-first-appearance events. The budget control retains events with at least 50 evaluation attempts left. For a group of events $\mathcal E$, the fraction never developed again is
\begin{equation}
P_{\mathrm{no\ development}}=\frac{1}{|\mathcal E|}\sum_{p\in\mathcal E}\mathbf1[N_{\mathrm{after}}(p)=0].
\end{equation}
This lies in $[0,1]$; larger values indicate less follow-up, not necessarily a worse final solution.

\paragraph{Relative benefit and paired kernel gain.}\mbox{}\\
Relative reduction in network group-mean error and paired kernel gain are
\begin{equation}
I_{\mathrm{error}}=1-\frac{\overline{\mathrm{Err}}_{1,\mathrm{hint}}}
{\overline{\mathrm{Err}}_{1,\mathrm{base}}},
\qquad
g_s=\frac{\ell_{\mathrm{base},s}^{\star}}{\ell_{\mathrm{hint},s}^{\star}}.
\end{equation}
Positive $I_{\mathrm{error}}$ means lower mean error; it is not a mean of per-seed percentage reductions. Kernel comparisons match task, backbone, and seed, using whole runs for from-start arms and post-onset candidates over the remaining budget for onset arms. The summary counts $g_s>1.05$ as better, $g_s<0.95$ as worse, and the rest as unchanged. $g_s=1$ means no gain; larger is better.

\paragraph{Kernel route concentration and reference-route adoption.}\mbox{}\\
For the diagnostic set $\mathcal A_{10}(b)$ of up to ten fastest valid kernels, let $n_b=|\mathcal A_{10}(b)|$ and use the label frequencies in Eq.~\ref{eq:label-frequencies}. We report
\begin{equation}
S_{\mathrm{route}}=|\{p:\hat\pi_b(p)>0\}|,\quad
N_{\mathrm{eff}}^{\mathrm{route}}=\exp\!\left[-\sum_p\hat\pi_b(p)\ln\hat\pi_b(p)\right],
\quad m_{\mathrm{route}}=\max_p\hat\pi_b(p).
\end{equation}
For $n_b>0$, $1\le N_{\mathrm{eff}}^{\mathrm{route}}\le S_{\mathrm{route}}\le n_b$ and $1/n_b\le m_{\mathrm{route}}\le1$. Route-family effective count uses the same formula after merging leaf routes by the annotation rubric. Lower effective count or higher dominant share indicates concentration, not necessarily failure. Reference-route adoption is
\begin{equation}
q_{\mathrm{ref}}(b)=\frac{1}{n_b}\sum_{x\in\mathcal A_{10}(b)}\mathbf1[\phi(x)=p_{\mathrm{ref}}],
\end{equation}
where $p_{\mathrm{ref}}$ is the expert's route. It lies in $[0,1]$; higher values mean greater adoption, not necessarily a faster implementation. Intervention figures use this final fastest-ten share, not a last-20-proposal retention threshold. Route-conditioned performance is the best observed latency on a route divided by expert latency; smaller is better and it is undefined if that route was never evaluated. The kernel performance curves show $1/G$ (Eq.~\ref{eq:metric-gap}), so larger is better and one matches the expert.

\subsection{PiEvo and statistical summaries}
\paragraph{Belief-posterior entropy.}\mbox{}\\
Let $b_t(p)$ be PiEvo's recorded posterior over $J_t$ active textual principles, distinct from the structural descriptor distribution $\hat\pi_b$. Then
\begin{equation}
H_t^{\mathrm{belief}}=-\sum_{p=1}^{J_t}b_t(p)\ln b_t(p),
\qquad
\overline H^{\mathrm{belief}}=
\frac{\sum_{t\in\mathcal B}w_t H_t^{\mathrm{belief}}}{\sum_{t\in\mathcal B}w_t},
\end{equation}
where $\mathcal B$ indexes recorded belief updates in the first or last 1,000 generated attempts, and $w_t$ is the number of attempts from that round inside the window. Entropy lies in $[0,\ln J_t]$ nats. Zero means all posterior mass is on one principle, or only one exists. Higher means less concentrated belief, not necessarily more diverse molecules or better search. Tables report attempt-weighted averages within runs, then mean$\pm$std across seeds; missing beliefs are omitted rather than assigned zero entropy.

\paragraph{Effective used-text count and semantic strategy share.}\mbox{}\\
In a generated-attempt window $\mathcal W$, let $w_t$ be the number of attempts from round $t$ inside the window and $j_t$ its actually used principle text, normalized only for whitespace. With $W=\sum_t w_t$,
\begin{equation}
q_j=\frac{\sum_t w_t\mathbf1[j_t=j]}{W},\qquad
T_{\mathrm{eff}}=\exp\!\left[-\sum_jq_j\ln q_j\right].
\end{equation}
This is the $T_{\mathrm{eff},e/l}$ reported in the PiEvo tables. Its range is $[1,K_{\mathcal W}]$, where $K_{\mathcal W}$ is the number of distinct used texts. Larger means more varied text usage, not necessarily distinct strategies. For semantic-route statistics, merge wording variants using the recorded operational labels and apply the same effective-count formula to the resulting frequencies. Dominant strategy share is their maximum, in $[1/K_{\mathrm{route}},1]$; larger means more concentrated use. The annotations and their limitations are described in Appendix~\ref{app:pievo-details}.

\paragraph{Anomaly counts and principle-update counts.}\mbox{}\\
Let $\mathcal R_{\mathcal W}$ contain rounds whose batch ends fall within $\mathcal W$, and $\mathcal Q_{\mathcal W}$ contain principle decisions at batch starts within $\mathcal W$. From the complete saved records,
\begin{equation}
A_{\mathrm{count}}(\mathcal W)=\sum_{t\in\mathcal R_{\mathcal W}}\mathbf1[\mathrm{triggered}_t],
\quad
A_{\mathrm{rate}}(\mathcal W)=\frac{A_{\mathrm{count}}(\mathcal W)}{|\mathcal R_{\mathcal W}|},
\quad
U_{\mathrm{principle}}(\mathcal W)=\sum_{t\in\mathcal Q_{\mathcal W}}\mathbf1[g_t\ne\texttt{PAUSE\_PRINCIPLES}].
\end{equation}
Counts are integers bounded by the number of eligible records; the rate lies in $[0,1]$ and is undefined for an empty window. Updates count generation/refinement calls, not proven new semantic directions; higher counts do not necessarily mean better search. Events are not fractionally weighted at window boundaries. 
Unless stated otherwise, early/late PiEvo tables use the first/last 1,000 attempts after initial-candidate loading and include forced principle warm-up in the early window. The main-text update comparison instead uses the 1,000 attempts immediately after the ten warm-up rounds: $3.800\pm2.775$ updates versus $1.000\pm2.236$ late for original PiEvo, and $1.000\pm0.707$ versus $0.400\pm0.894$ for $\sigma=0.9$. These shifted update windows do not change the early posterior or text-usage windows in the tables. Trigger scarcity alone does not show that small improvements suppress anomalies.

\paragraph{Statistical summaries.}\mbox{}\\
For a run-level metric $z_s$ on $n$ observed seeds,
\begin{equation}
\bar z=\frac{1}{n}\sum_{s=1}^{n}z_s,\qquad
s_z=\sqrt{\frac{1}{n-1}\sum_{s=1}^{n}(z_s-\bar z)^2}.
\end{equation}
Tables and convergence bands use mean$\pm$sample standard deviation, not standard error. Nonlinear metrics are calculated within runs before aggregation. First-attempt plots summarize events by the median and interquartile range: the table reports $[Q_{0.25},Q_{0.75}]$, while the IQR width is $Q_{0.75}-Q_{0.25}$. These describe event variation, not confidence intervals. For positive paired kernel gains, the geometric mean is $\exp(n^{-1}\sum_s\ln g_s)$; pooled kernel speed curves likewise average the log ratios before exponentiating. Diversity-injection summaries first average backbones within each seed, then compute over the five seed averages.

\paragraph{Token usage and estimated price.}\mbox{}\\
For token counts $T_{\mathrm{in}},T_{\mathrm{out}}$ and USD prices per million tokens $c_{\mathrm{in}},c_{\mathrm{out}}$,
\begin{equation}
T_{\mathrm{in}}^{(\mathrm M)}=\frac{T_{\mathrm{in}}}{10^6},\quad
T_{\mathrm{out}}^{(\mathrm M)}=\frac{T_{\mathrm{out}}}{10^6},\quad
C_{\mathrm{USD}}=c_{\mathrm{in}}T_{\mathrm{in}}^{(\mathrm M)}
+c_{\mathrm{out}}T_{\mathrm{out}}^{(\mathrm M)}.
\end{equation}
Cost is computed per run before aggregation. It is a list-price estimate, excluding caching discounts and infrastructure costs.

\subsection{Candidate-level metrics}
\paragraph{Validity and uniqueness.}\mbox{}\\
Let $N_{\mathrm{gen}}$, $N_{\mathrm{fail}}$, and $N_{\mathrm{rep}}$ be the logged generated, invalid, and duplicate counts. 
\begin{equation}
V=1-\frac{N_{\mathrm{fail}}}{N_{\mathrm{gen}}},
\qquad
U=1-\frac{N_{\mathrm{rep}}}{N_{\mathrm{gen}}}.
\label{eq:metric-valid-unique}
\end{equation}
Both lie in $[0,1]$: larger means fewer invalid proposals or repetitions, respectively. Molecular duplicates use canonical SMILES. Network uniqueness retains the runner's logged duplicate definition; it is not recomputed by the unordered-set deduplication used for selected-population metrics. Importantly, $U$ is a non-repetition fraction over \emph{all generated candidates}, not the fraction of valid candidates that are distinct. Invalid proposals contribute to $N_{\mathrm{gen}}$ but not $N_{\mathrm{rep}}$, so $U$ should be read alongside $V$. Kernel validity is $N_{\mathrm{pass}}/N_{\mathrm{eval}}$, where passing requires compilation and harness correctness.

\paragraph{Molecular Tanimoto diversity.}\mbox{}\\
Let $B(x)$ be the active bits of a 2,048-bit Morgan fingerprint (radius 2, chirality disabled). For the $n_t$ distinct molecules in $\mathcal A_{50}(t)$,
\begin{equation}
\begin{split}
\operatorname{Tan}(x,x')&=\frac{|B(x)\cap B(x')|}{|B(x)\cup B(x')|},\\
\mathrm{Div}_{\mathrm{mol}}(t)&=
\frac{2}{n_t(n_t-1)}\sum_{i<k}
[1-\operatorname{Tan}(x_i,x_k)].
\end{split}
\label{eq:metric-tanimoto}
\end{equation}
Self-pairs are excluded. The statistic is undefined for fewer than two molecules and lies in $[0,1]$ otherwise. Higher means more fingerprint variation among the selected 50 molecules, not necessarily more scaffold-level principles.

\paragraph{Code diversity.}\mbox{}\\
Tokenize source into identifiers, digit sequences, and individual non-whitespace symbols. Let $B_5(x)$ be its set of consecutive five-token sequences. Within a window $\mathcal W$ of $n_{\mathcal W}\ge2$ candidates,
\begin{equation}
\mathrm{Div}_{\mathrm{code}}(\mathcal W)=
\frac{2}{n_{\mathcal W}(n_{\mathcal W}-1)}
\sum_{i<k}\left[1-\frac{|B_5(x_i)\cap B_5(x_k)|}{|B_5(x_i)\cup B_5(x_k)|}\right].
\end{equation}
A shorter source contributes its whole token sequence; two empty sets have distance zero. Range $[0,1]$: larger means more surface variation, not necessarily a different executed route.

\section{Oracle-principle hints}
\label{app:hints}
The hints of Table~\ref{tab:oracle-hints} are written as principle-level search directions. They name a region or role pattern but no candidate, scaffold, node set or ranked list.

\begin{table}[h]
\centering
\caption{Oracle-principle hints used in the interventions.}
\label{tab:oracle-hints}
\scriptsize
\begin{tabular}{p{0.19\linewidth}p{0.72\linewidth}}
\toprule
\textbf{Task} & \textbf{Hint} \\
\midrule
Molecular (L2) & Search around compact heteroaryl-amide chemotypes rather than only refining generic benzamide or para-methoxy aryl-amide families. Favor low-synthetic-complexity molecules with a small sulfur-containing or heteroaromatic ring, a simple aryl/benzyl partner, and light substituent tuning to keep QED high while driving the unwanted target probabilities near zero. \\
Dolphin (L2) & A direction is a heterogeneous degree profile: include one low-degree or leaf-like sentinel and combine it with low-to-mid and moderate/high-role nodes whose errors complement each other as a set. \\
Residence hall (L2) & Look for five sentinels that jointly cover different network roles across degree and neighborhood structure. Try heterogeneous complementarity, try a mixed low/mid/moderately-high profile and test replacements that preserve that role coverage. \\
\midrule
Kernel, all problems (L1) & Optimization trajectories on this problem typically settle early on one overall design approach whose latency stays far above what the hardware allows, and further refinement within that approach does not close the gap. Whenever the implementations shown share the same overall algorithmic decomposition, do not keep tuning it; propose a candidate whose decomposition of the computation is fundamentally different. \\
GQA / MLA paged decode (L2) & Instead of one program per (sequence, head) that walks the whole KV sequence serially, split each sequence's KV range into several chunks processed by different programs in parallel, each producing partial attention statistics (running max, normalizer, accumulated output), and merge the chunks afterwards with log-sum-exp rescaling. Pack the heads that share the same KV as the row dimension of a matrix product so that scores and outputs run on tensor cores and each KV block is read once. \\
MLA paged prefill (L2) & Treat the computation as a two-dimensional tiling---a block of query tokens (with all heads, which share the same compressed KV) against a block of KV tokens---so that score computation and output accumulation are matrix products on tensor cores, with an online softmax carried across KV blocks and the causal mask applied only where a block crosses the diagonal. Long sequences can additionally be split along the KV dimension across programs and merged with log-sum-exp rescaling. Avoid host-side loops over sequences. \\
FP8 MoE (L2) & Do not dequantize the FP8 weights to fp32 or bf16 before the GEMMs. Feed the FP8 activations and weights directly to tensor-core matrix multiplies and apply the 128-block scales to the accumulator as the K dimension is traversed. Group the tokens by expert once and run the per-expert GEMMs as one grouped launch rather than a Python loop over experts, fusing the SwiGLU between the two GEMMs and the top-k weighted reduction after the second. \\
MLA paged decode (L3, example) & The L2 direction, followed by an outline: (1) choose a chunk length and a grid over (sequence, chunk); programs past the sequence length exit early. (2) Chunk kernel: load q\_nope as a [16, 512] tile and q\_pe as [16, 64]; iterate the chunk's KV positions in blocks, gather compressed-KV rows [block, 512] and rope rows [block, 64] through kv\_indices; scores = dot(q\_nope, ckv$^\top$) + dot(q\_pe, kpe$^\top$) scaled by sm\_scale; keep a per-head running max and normalizer; accumulate dot(P, ckv) into a [16, 512] tile; write partial max, normalizer and output to scratch indexed by (sequence, chunk). (3) Merge kernel per sequence and head: global max over chunks, rescale each chunk, sum normalizers and outputs, divide, emit the log-sum-exp. (4) Pick the chunk length so that sequences $\times$ chunks is a small multiple of the SM count. \\
\bottomrule
\end{tabular}
\end{table}

\end{document}

%% file: math_commands.tex
\usepackage{amsmath,amsfonts,bm}

\def\eqref#1{equation~\ref{#1}}

\def\1{\bm{1}}

\DeclareMathAlphabet{\mathsfit}{\encodingdefault}{\sfdefault}{m}{sl}
\SetMathAlphabet{\mathsfit}{bold}{\encodingdefault}{\sfdefault}{bx}{n}

